\documentclass[runningheads]{llncs}

\usepackage{eccv}

\usepackage{eccvabbrv}

\usepackage{graphicx}
\usepackage{booktabs}
\usepackage{float}
\usepackage{xcolor}
\usepackage{multirow}

\usepackage[ruled,linesnumbered,vlined]{algorithm2e}
\SetKwInput{KwInput}{Input}
\SetKwInput{KwOutput}{Output}
\SetKwInput{KwHyper}{Hyperparameters}
\SetKwComment{Comment}{$\triangleright$~}{}
\SetKwProg{Fn}{Function}{:}{}
\SetAlgoNlRelativeSize{-1}
\SetNlSkip{0.5em}

\definecolor{fluxblue}{RGB}{31,78,161}
\definecolor{sdxlred}{RGB}{178,34,52}
\newcommand{\gimg}[1]{\includegraphics[width=0.165\linewidth,height=0.165\linewidth]{#1}}
\newlength{\gpw}\newlength{\gph}
\newcommand{\gpimg}[1]{%
  \setlength{\gpw}{0.178\linewidth}%
  \setlength{\gph}{\dimexpr 3\gpw/2\relax}%
  \parbox[c][\gph][c]{\gpw}{%
    \centering\includegraphics[width=\gpw,height=\gph]{#1}}}
\newcommand{\grimg}[2]{%
  \setlength{\gpw}{0.178\linewidth}%
  \setlength{\gph}{\dimexpr 3\gpw/2\relax}%
  \setlength{\gpw}{\dimexpr(\gph-0.5mm)/2\relax}%
  \parbox[c][\gph][c]{0.178\linewidth}{%
    \centering
    \includegraphics[width=\gpw,height=\gpw]{#1}\\[0.5mm]%
    \includegraphics[width=\gpw,height=\gpw]{#2}}}

\usepackage[accsupp]{axessibility}  

\usepackage[hypertexnames=false]{hyperref}
\hypersetup{
  pdftitle={The Quadratic Geometry of Flow Matching: Semantic Granularity Alignment for Text-to-Image Synthesis},
  pdfauthor={Zhinan Xiong, Shunqi Yuan}
}

\usepackage{orcidlink}

\begin{document}

\title{The Quadratic Geometry of Flow Matching: Semantic Granularity Alignment for Text-to-Image Synthesis}

\titlerunning{Quadratic Geometry of Flow Matching: SGA for T2I Synthesis}

\author{Zhinan Xiong\inst{1}\thanks{Corresponding author.} \and
Shunqi Yuan\inst{2}}

\authorrunning{Z.~Xiong and S.~Yuan}

\institute{Conservatoire National des Arts et M\'{e}tiers, Paris, France\\
\email{zhinan.xiong.auditeur@lecnam.net} \and
Sun Yat-sen University, Guangzhou, China}

\maketitle
\enlargethispage{10mm}
\begin{abstract}
  In this work, we analyze the optimization dynamics of generative fine-tuning. We observe that under the Flow Matching framework, the standard MSE objective can be formulated as a Quadratic Form governed by a dynamically evolving Neural Tangent Kernel (NTK). This geometric perspective reveals a latent Data Interaction Matrix, where diagonal terms represent independent sample learning and off-diagonal terms encode residual correlation between heterogeneous features.

  Although standard training implicitly optimizes these cross-term interferences, it does so without explicit control; moreover, the prevailing data-homogeneity assumption may constrain the model's effective capacity. Motivated by this insight, we propose Semantic Granularity Alignment (SGA), using Text-to-Image synthesis as a testbed. SGA engineers targeted interventions in the vector residual field to mitigate gradient conflicts. Evaluations across DiT and U-Net architectures confirm that SGA advances the efficiency--quality trade-off by accelerating convergence and improving structural integrity.
  \keywords{Parameter-Efficient Fine-Tuning \and Large Vision Models \and Generative Domain Adaptation \and Data Efficiency \and Flow Matching \and Conditional Flow Matching}
\end{abstract}
\vspace{-2mm}
\begin{figure}[H]
\centering
\setlength{\tabcolsep}{0.5mm}
\footnotesize
\setlength{\gpw}{0.3\linewidth}%
\setlength{\gph}{\dimexpr 3\gpw/2\relax}%
\begin{tabular}{@{}ccc@{}}
\textbf{\color{sdxlred}Ref.} & \textbf{\color{gray}Baseline} & \textbf{\color{fluxblue}SGA (Ours)} \\[1mm]
\parbox[c][\gph][c]{\gpw}{%
  \centering
  \includegraphics[width=\dimexpr(\gph-1mm)/2\relax,height=\dimexpr(\gph-1mm)/2\relax]{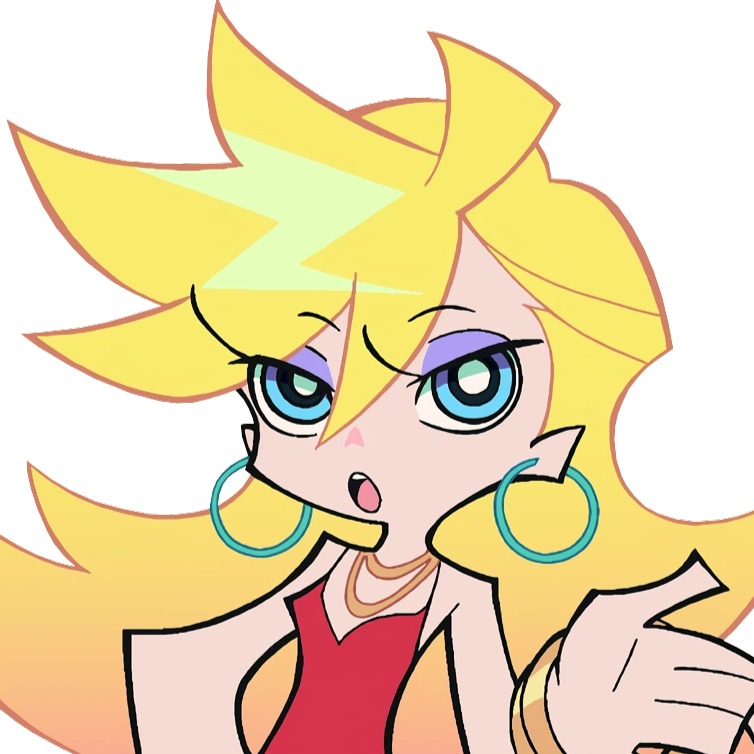}\\[1mm]%
  \includegraphics[width=\dimexpr(\gph-1mm)/2\relax,height=\dimexpr(\gph-1mm)/2\relax]{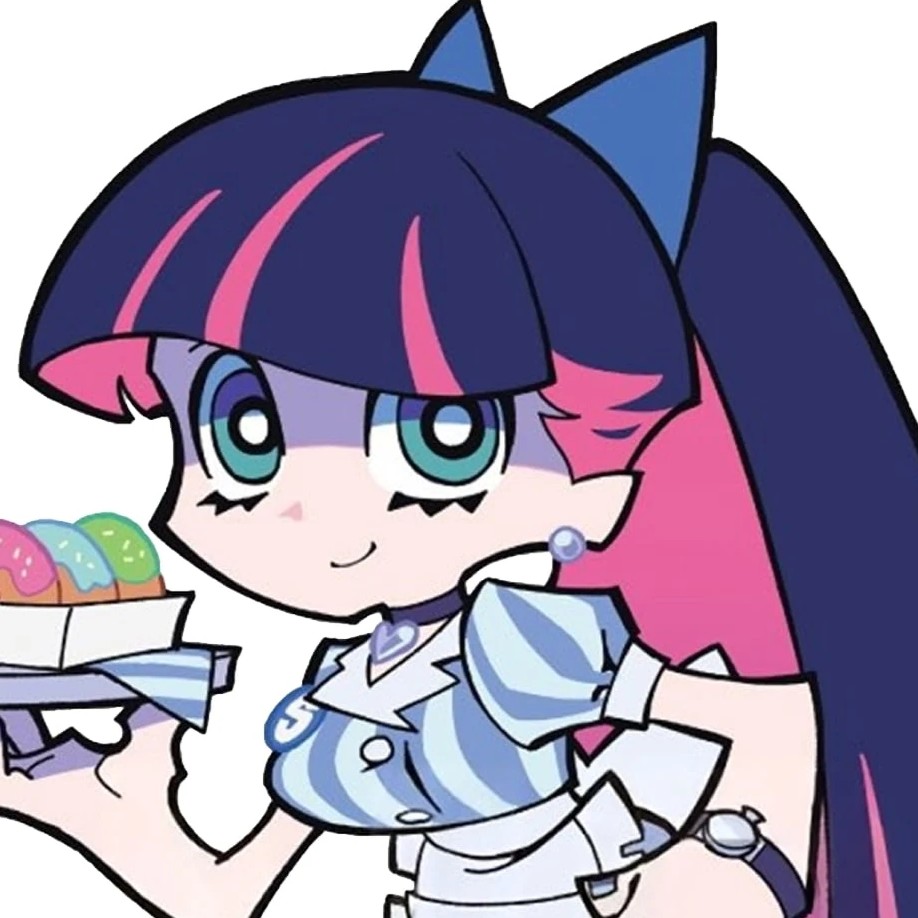}} &
\parbox[c][\gph][c]{\gpw}{%
  \centering\includegraphics[width=\gpw,height=\gph]{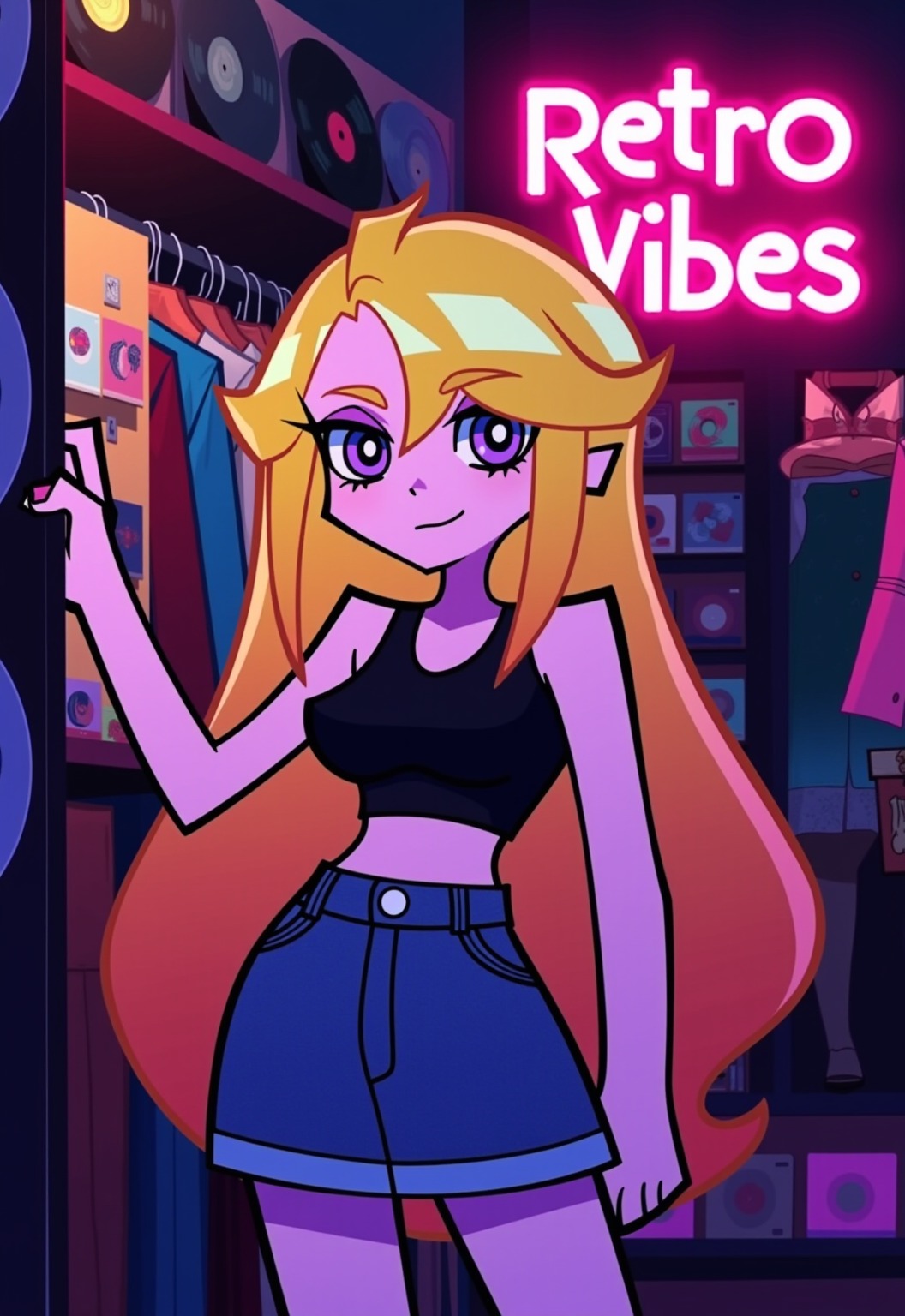}} &
\parbox[c][\gph][c]{\gpw}{%
  \centering\includegraphics[width=\gpw,height=\gph]{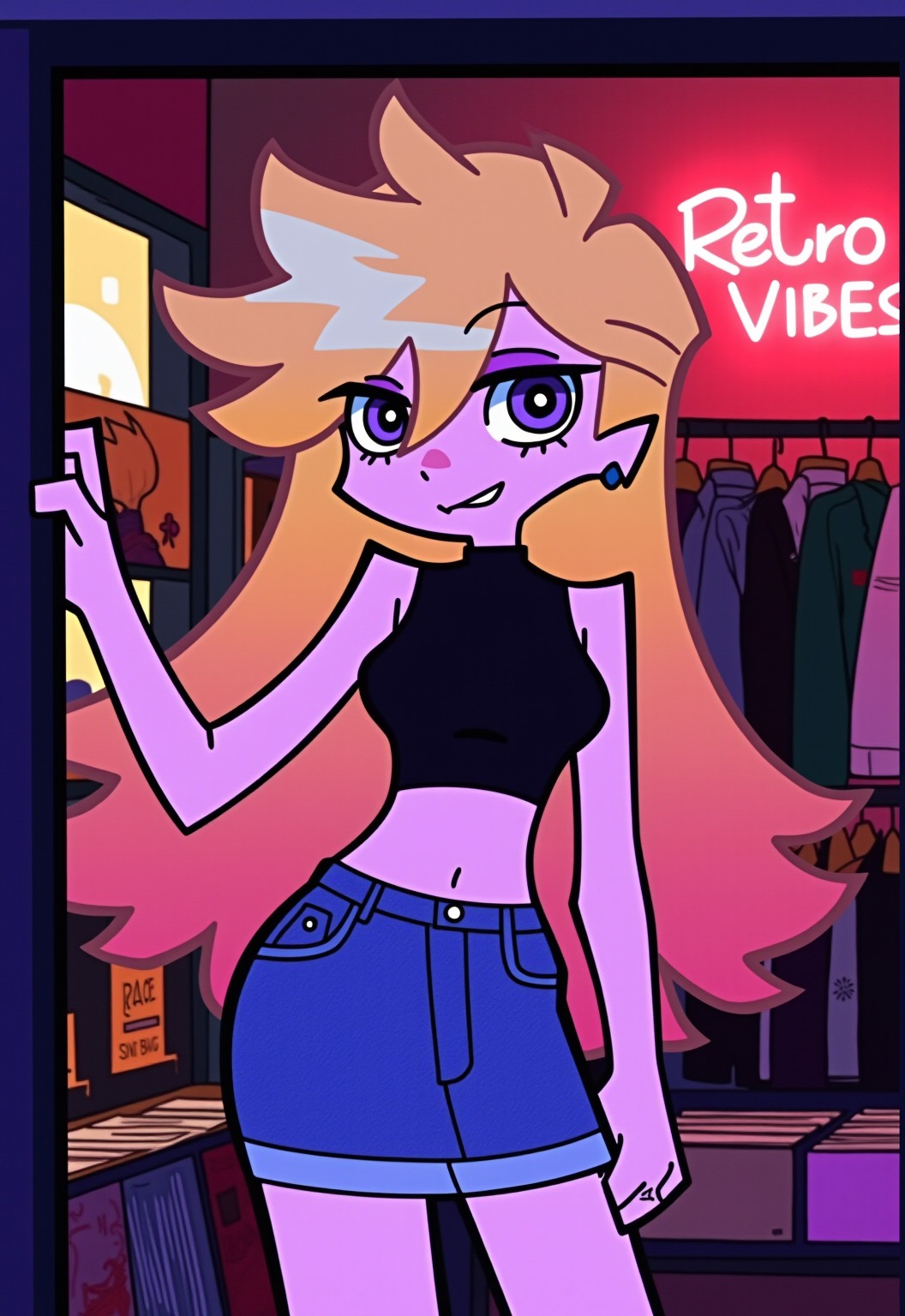}} \\
\end{tabular}
\vspace{-2mm}
\end{figure}
\vspace{-2mm}
\begin{figure}[H]
\centering
\setlength{\tabcolsep}{0.5mm}
\footnotesize
\setlength{\gpw}{0.3\linewidth}%
\setlength{\gph}{\dimexpr 3\gpw/2\relax}%
\begin{tabular}{@{}ccc@{}}
\parbox[c][\gph][c]{\gpw}{%
  \centering
  \includegraphics[width=\dimexpr(\gph-1mm)/2\relax,height=\dimexpr(\gph-1mm)/2\relax]{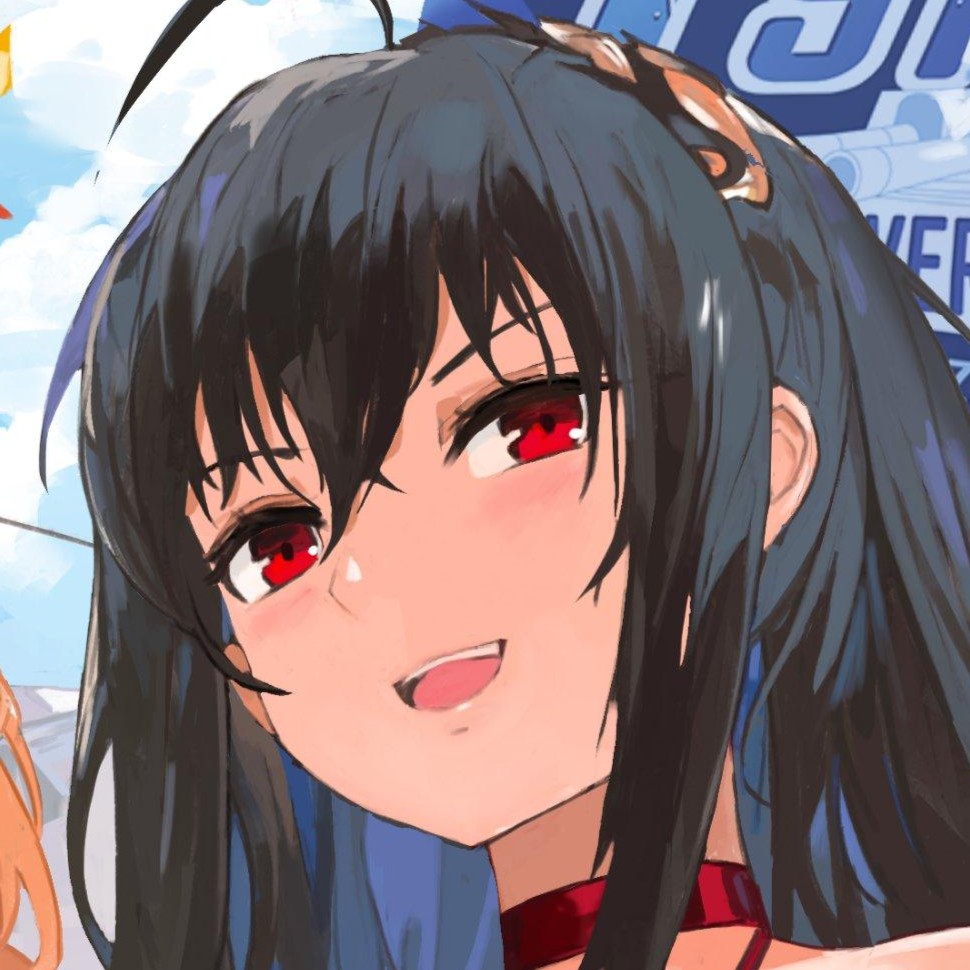}\\[1mm]%
  \includegraphics[width=\dimexpr(\gph-1mm)/2\relax,height=\dimexpr(\gph-1mm)/2\relax]{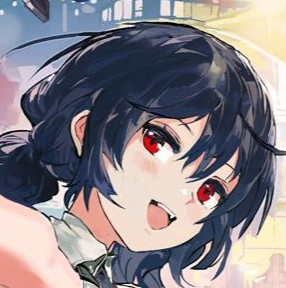}} &
\parbox[c][\gph][c]{\gpw}{%
  \centering\includegraphics[width=\gpw,height=\gph]{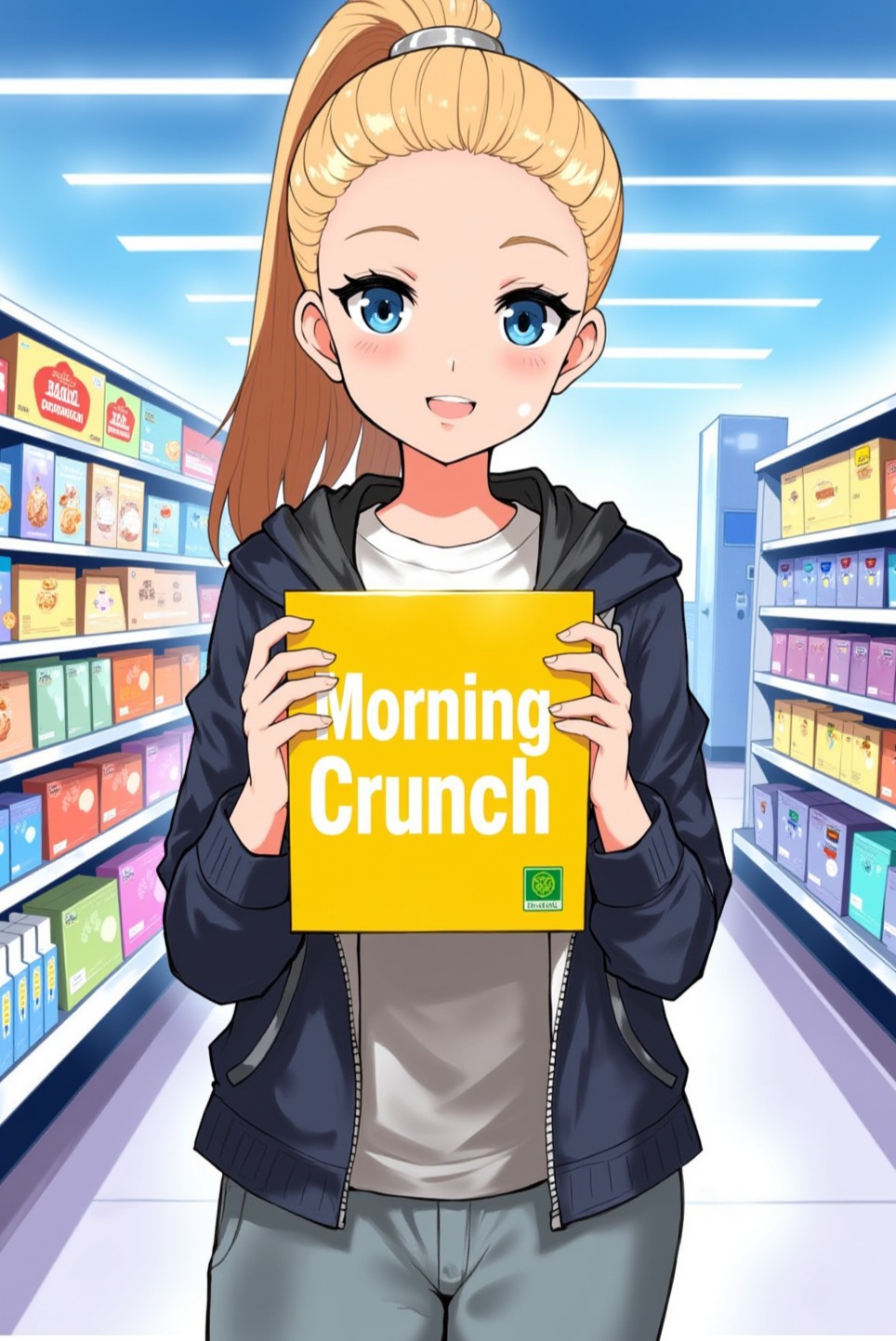}} &
\parbox[c][\gph][c]{\gpw}{%
  \centering\includegraphics[width=\gpw,height=\gph]{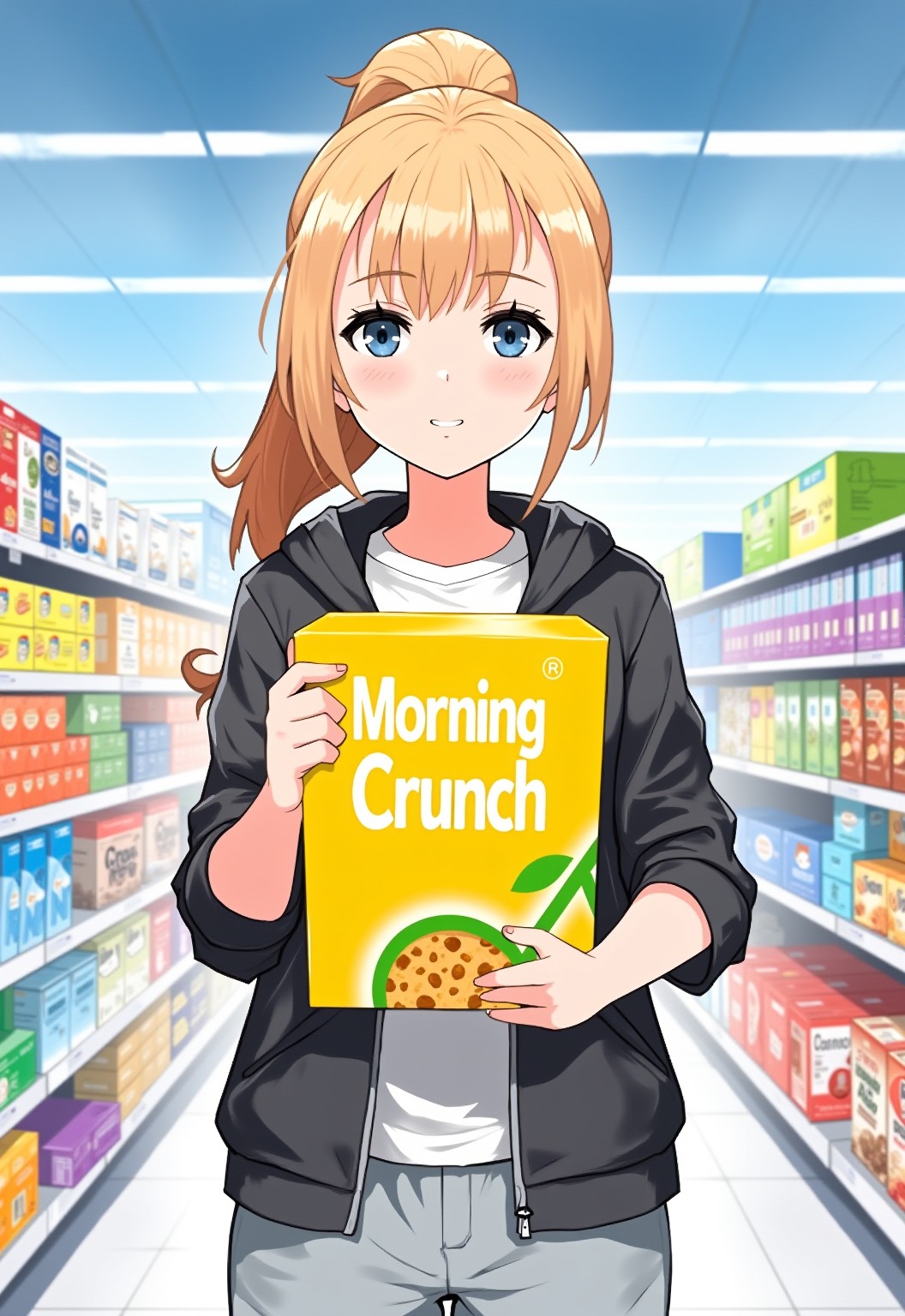}} \\
\noalign{\vspace{2mm}}
\parbox[c][\gph][c]{\gpw}{%
  \centering
  \includegraphics[width=\dimexpr(\gph-1mm)/2\relax,height=\dimexpr(\gph-1mm)/2\relax]{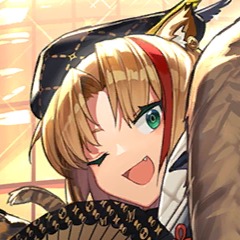}\\[1mm]%
  \includegraphics[width=\dimexpr(\gph-1mm)/2\relax,height=\dimexpr(\gph-1mm)/2\relax]{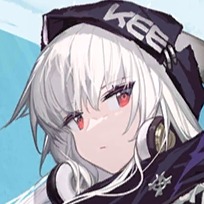}} &
\parbox[c][\gph][c]{\gpw}{%
  \centering\includegraphics[width=\gpw,height=\gph]{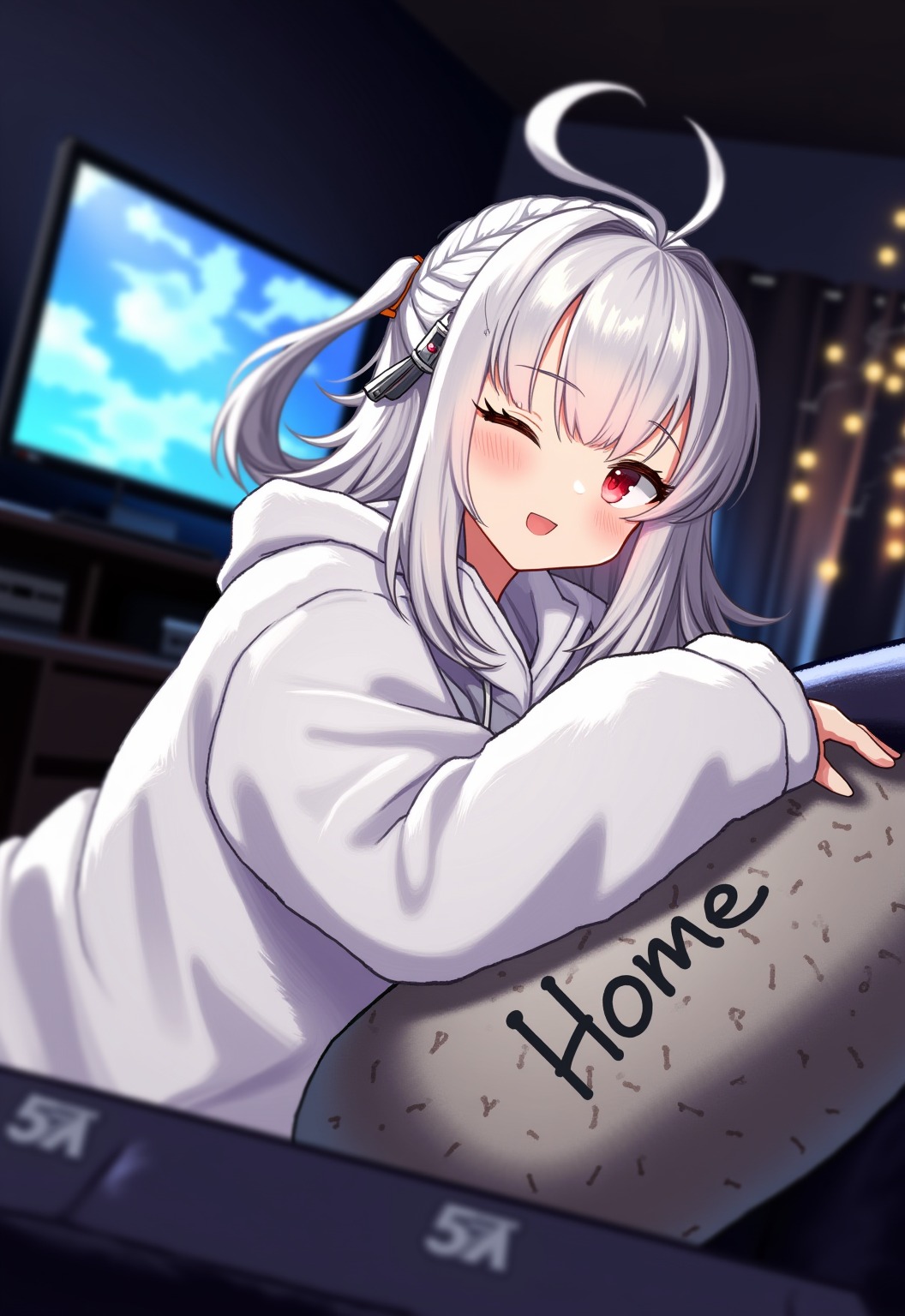}} &
\parbox[c][\gph][c]{\gpw}{%
  \centering\includegraphics[width=\gpw,height=\gph]{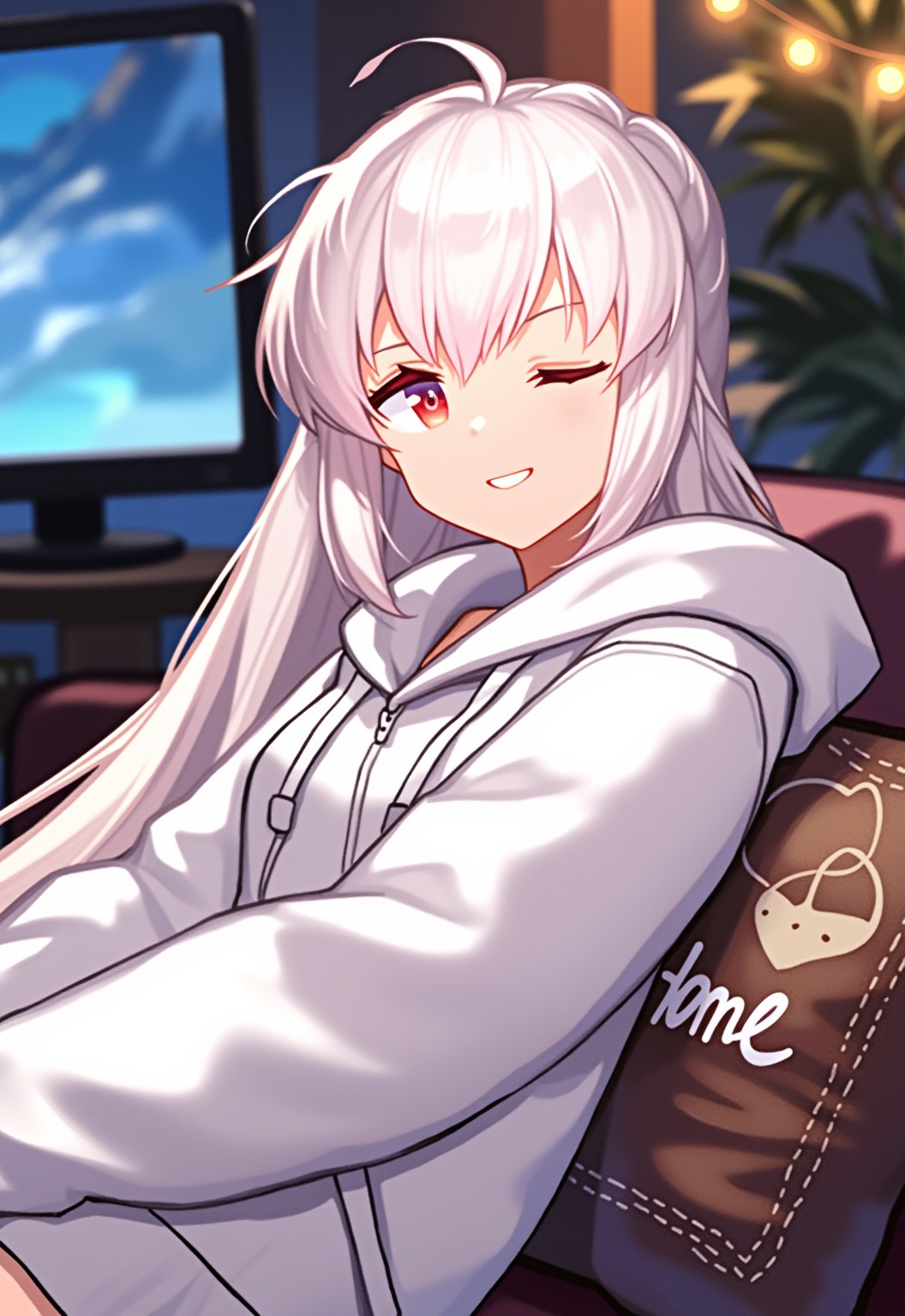}} \\
\end{tabular}
\vspace{-1mm}
\caption{Qualitative comparison between the Baseline and our SGA method across three GDA domains. For each row, the reference images (left) define the target domain; the Baseline (middle) fails to preserve domain-specific attributes, while SGA (right) faithfully captures the target domain characteristics.}
\label{fig:teaser-rows}
\vspace{-2mm}
\end{figure}

\section{Introduction}
\label{sec:intro}

Generative post-training has firmly established itself as a standard paradigm. While the community has achieved remarkable progress in architectural efficiency (\eg, LoRA~\cite{hu2022lora}) and sampling acceleration (\eg, ODE solvers), the principles governing data composition remain largely empirical. Unlike mathematically well-defined optimization mechanisms, current practices still lack systematic theoretical guidance on how data mixtures dictate convergence, often treating datasets as homogeneous signals. This ``data-oblivious'' approach neglects the complex, often conflicting gradients arising from multi-granular features, creating a hidden bottleneck for efficient adaptation.

\noindent\textbf{Key Observation: The Quadratic Geometry of Flow Matching.}
We revisit the mathematical foundations of Flow Matching (FM). We demonstrate that under standard assumptions, for architectures adhering to the FM framework, optimizing the standard Mean Squared Error (MSE) objective is mathematically equivalent to optimizing a latent Quadratic Form, which in turn corresponds to shaping a dynamically evolving Neural Tangent Kernel (NTK). This geometric perspective suggests that the learning dynamics are governed by a Data Interaction Matrix, whose diagonal terms represent the independent learning of specific data samples, while off-diagonal terms encode the residual correlation---constructive or destructive---between heterogeneous data features.

\noindent\textbf{Methodology.}
Consequently, fine-tuning is not merely fitting a distribution but seeking a geometric equilibrium within this interaction field. To operationalize this geometric insight within Text-to-Image (T2I) synthesis, we propose Semantic Granularity Alignment (SGA). By engineering targeted interventions in the output-vector residual space, SGA aligns the data structure with the underlying optimization geometry, thereby accelerating convergence while improving generative integrity. Concretely, SGA comprises two mechanisms: (i)~\emph{Tuple-wise Optimization}, which constructs multi-granularity semantic tuples so that cross-scale features co-occur within each update step, dampening gradient oscillation; and (ii)~\emph{Scale-Adaptive Modulation}, which decouples and re-aligns structural and textural learning across distinct denoising frequency bands. Empirically, this alignment improves the efficiency-quality trade-off without significant additional overhead.

\section{Related Work}
\label{sec:related}

\noindent\textbf{Conditional Flow Matching.}
Diffusion models~\cite{ho2020denoising,song2021scorebased} and their continuous-time generalization through Flow Matching~\cite{lipman2023flow,albergo2023building} and Rectified Flow~\cite{liu2023flow} have established a unified framework for generative modeling, scaled effectively by Rectified Flow Transformers~\cite{esser2024scaling} and extended to discrete state-spaces~\cite{campbell2024generative} and variational formulations~\cite{guo2025variational}. Conditional Flow Matching (CFM)~\cite{tong2024improving} further generalizes these into a tractable regression objective. Latent Diffusion Models~\cite{rombach2022high} project generation into a compressed latent space, while Classifier-Free Guidance~\cite{ho2022classifier} has become the de facto conditional sampling strategy.

\noindent\textbf{Parameter-Efficient Fine-Tuning.}
Low-Rank Adaptation (LoRA)~\cite{hu2022lora} and its variant DoRA~\cite{liu2024dora} enable competitive fine-tuning by constraining weight updates to a low-rank subspace, providing the architectural backbone upon which data-level strategies can be built. LyCORIS~\cite{yeh2024navigating} further extends this family with alternative decomposition strategies for T2I customization.

\noindent\textbf{Data-Centric Generative Adaptation.}
Engineering solutions such as Aspect Ratio Bucketing~\cite{novelai2022bucketing}, Zero Terminal SNR~\cite{lin2024common}, and community-driven data toolchains~\cite{deepghs2024} have become standard in Generative Domain Adaptation (GDA). Subject-driven methods such as DreamBooth~\cite{ruiz2023dreambooth} and Textual Inversion~\cite{gal2023image} pioneered few-shot personalization, while ZipLoRA~\cite{shah2024ziplora} and subsequent efforts~\cite{frenkel2025blora,ouyang2025klora,ye2023ipadapter,sohn2024styledrop,wang2024instantstyle} advanced few-shot stylized generation. Yet these methods operate at the preprocessing or embedding level; a unified framework connecting data structure design with the optimization objective is less frequently addressed.

\section{Methodology}
\label{sec:method}

\subsection{Theoretical Formulation: The Geometry of Interaction}
\label{sec:method-theory}

We analyze the optimization objective of generative fine-tuning through a geometric lens. We demonstrate that under the Flow Matching~\cite{lipman2023flow} framework, the minimization of the standard Mean Squared Error (MSE) loss can be reformulated as the optimization of a Quadratic Interaction Matrix, which governs the interference between heterogeneous data manifolds.

\subsubsection{The General Objective: Flow Matching (FM).}
Consider a probability path $p_t(x)$ transforming a noise distribution $p_0$ to a data distribution $p_1$~\cite{lipman2023flow,liu2023flow}. The objective of Flow Matching is to regress a time-dependent vector field $v_t(x; \theta)$ that approximates the ground-truth marginal vector field $u_t(x)$:
\begin{equation}
\mathcal{L}_{FM}(\theta) = \mathbb{E}_{t, p_t(x)} \left[ \| v_t(x; \theta) - u_t(x) \|^2 \right]
\label{eq:fm}
\end{equation}

\noindent\textbf{Proposition 1} (Linear Superposition of Vector Fields).
We model the data distribution empirically as a mixture of Dirac measures over $N$ latent sub-manifolds indexed by $\xi$~\cite{bishop2006pattern,goodfellow2016deep}. Under the CFM framework~\cite{tong2024improving}, the marginal probability density $p_t(x)$ decomposes linearly over conditionals, and with the standard Gaussian conditional path assumption, the optimal marginal vector field $u_t(x)$ can be expressed as a probability-weighted superposition of the sub-fields $u^\xi_t(x)$:
\begin{equation}
u_t(x) = \sum_{\xi=1}^{N} \alpha_\xi(x) \cdot u^\xi_t(x)
\label{eq:decomposition}
\end{equation}
where $\alpha_\xi(x)$ represents the local density (or weight) of the $\xi$-th sub-manifold.

\noindent\textbf{Proposition 2} (The Quadratic Form of Optimization).
Substituting \cref{eq:decomposition} into \cref{eq:fm} and applying the Partition of Unity ($\sum_{\xi}\alpha_{\xi}(x)=1$), we define the per-granularity residual $\Delta_\xi(x,t) \triangleq v_\theta(x,t) - u_t^\xi(x)$ and expand the squared norm to obtain a Quadratic Form (see Suppl.~B for derivation):
\begin{equation}
\mathcal{L}_{FM}^{local}(x) = \boldsymbol{\alpha}^\top \mathbf{\Omega} \, \boldsymbol{\alpha}
= \underbrace{\textstyle\sum_{\xi} \alpha_\xi^2 \|\Delta_\xi\|^2}_{\text{Independent Learning}}
+ \underbrace{\textstyle\sum_{\xi \neq \eta} \alpha_\xi \alpha_\eta \langle \Delta_\xi, \Delta_\eta \rangle}_{\text{Residual Correlation}}
\label{eq:quadratic}
\end{equation}
where $\mathbf{\Omega}(x) \in \mathbb{R}^{N \times N}$ is the symmetric \emph{Data Interference Matrix} with entries $\Omega_{\xi\eta} = \langle \Delta_\xi, \Delta_\eta \rangle$.

\noindent\textbf{Proposition 3} (NTK Emergence in Gradient Space).
Let $J_\theta(x,t) = \partial v_\theta / \partial \theta$ denote the network Jacobian. By the chain rule, the total parameter gradient decomposes as $g_{\text{total}} = J_\theta^\top e(x,t) = \sum_\xi g_\xi$, with $e(x,t) \triangleq v_\theta(x,t) - u_t(x)$ denoting the prediction residual, where $g_\xi \triangleq \alpha_\xi \, J_\theta^\top \Delta_\xi$ (see Suppl.~E). The pairwise gradient interaction:
\begin{equation}
\langle g_\xi, g_\eta \rangle = \alpha_\xi \alpha_\eta \, \Delta_\xi^\top \underbrace{J_\theta J_\theta^\top}_{\Theta_\theta(x,t)} \Delta_\eta
\label{eq:ntk-inner}
\end{equation}
reveals the Neural Tangent Kernel (NTK)~\cite{jacot2018neural} $\Theta_\theta(x,t)$: the output-space geometry of $\mathbf{\Omega}$ directly governs the parameter-space gradient dynamics through the NTK.

\begin{figure}[H]
\centering
\includegraphics[width=\linewidth]{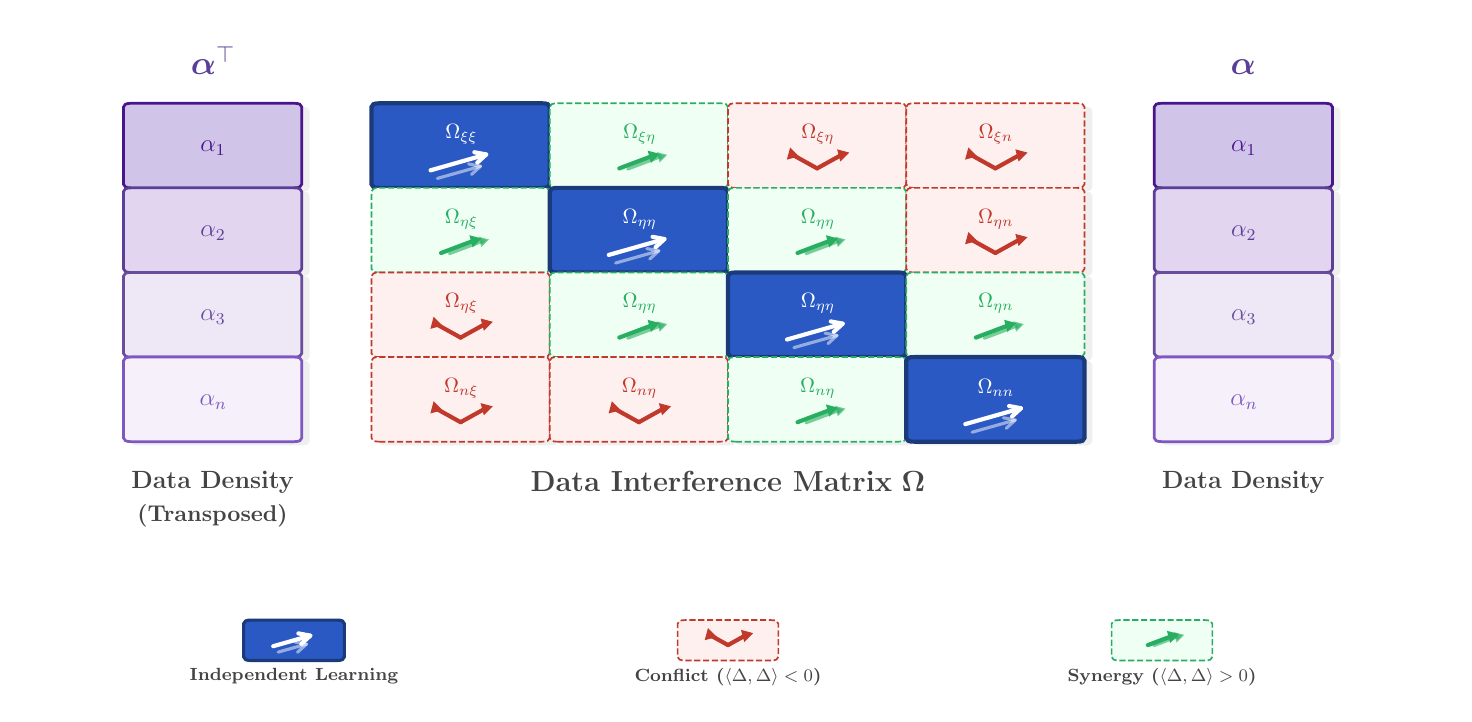}
\caption{Structure of the Data Interference Matrix $\mathbf{\Omega}$. Diagonal entries (blue) represent independent learning within each sub-manifold. Off-diagonal entries encode cross-scale interactions: constructive (synergy, green) or destructive (conflict, red).}
\label{fig:interaction-matrix}
\end{figure}

\noindent\textbf{Corollary 1} (Optimization Equivalence).
Since the marginal vector field $u_t(x)$ is generally intractable, Conditional Flow Matching (CFM)~\cite{tong2024improving} is employed as a proxy, typically instantiated as a Mean Squared Error (MSE) loss. Given that $\nabla_\theta \mathcal{L}_{FM} = \nabla_\theta \mathcal{L}_{CFM}$, we establish the following relationship:
\begingroup\allowdisplaybreaks
\begin{align}
\text{Optimizing MSE} &\iff \text{Optimizing CFM} \iff \text{Optimizing FM} \notag\\
&\iff \min_{\boldsymbol{\theta}} \mathbb{E}_{x,t} \!\left[ \boldsymbol{\alpha}^\top \mathbf{\Omega}(\theta)\, \boldsymbol{\alpha} \right] \iff \text{Shaping } \Theta_\theta(x,t)
\label{eq:equivalence}
\end{align}
\endgroup

This derivation highlights the intrinsic geometry of fine-tuning: the MSE objective is, in effect, training a vector residual field governed by a dynamically evolving NTK. However, the intrinsic complexity of the NTK renders direct analytical intervention intractable. We therefore adopt an engineering approach: rather than manipulating the kernel itself, we intervene in the vector residual field by restructuring the data that shapes it. Specifically, we first apply Hierarchical Semantic Decomposition (H-SD) to partition the original dataset into three semantically distinct sub-manifolds---Macro (global structure), Meso (mid-level layout), and Micro (fine-grained texture)---so that the training set is reorganized as $Y = \bigcup_{\xi \in \mathcal{T}} Y_\xi$ with $\mathcal{T}=\{\text{Macro}, \text{Meso}, \text{Micro}\}$. By the linearity of the conditional vector field (Eq.~\ref{eq:decomposition}), each entry of the resulting matrix captures the vector-field residual associated with a particular sub-manifold rather than a mere granularity label; we therefore term the instantiated $\mathbf{\Omega}$ the \emph{Semantic Interference Matrix}. The diagonal entries $\Omega_{\xi\xi}$ capture the self-alignment error within each granularity, while the off-diagonal entries $\Omega_{\xi\eta}$ quantify cross-scale semantic interference. When opposing residual directions arise in output space (\eg, $\langle \Delta_{\text{Macro}}, \Delta_{\text{Micro}} \rangle < 0$), this divergence propagates through the network Jacobian into parameter space, inducing gradient conflict and learning oscillation across update steps. Standard fine-tuning handles this off-diagonal interference only implicitly through stochastic sampling; our SGA framework instead provides an explicit, structured mechanism to regulate it.

\subsection{The Fine-Tuning Dilemma}
\label{sec:dilemma}

\begin{figure}[H]
\centering
\includegraphics[width=0.78\linewidth]{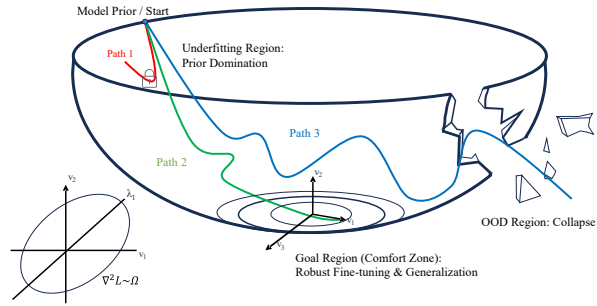}
\vspace{-2mm}
\caption{The optimization landscape $\mathcal{L}_{FM} = \boldsymbol{\alpha}^\top \mathbf{\Omega}\,\boldsymbol{\alpha}$ of CFM fine-tuning. \textbf{Path~1} (red): Underfitting Region. \textbf{Path~3} (blue): OOD Region. Only \textbf{Path~2} (green) navigates the narrow Goal Region.}
\label{fig:loss-landscape}
\vspace{-2mm}
\end{figure}

\begin{figure}[H]
\centering
\includegraphics[width=\linewidth]{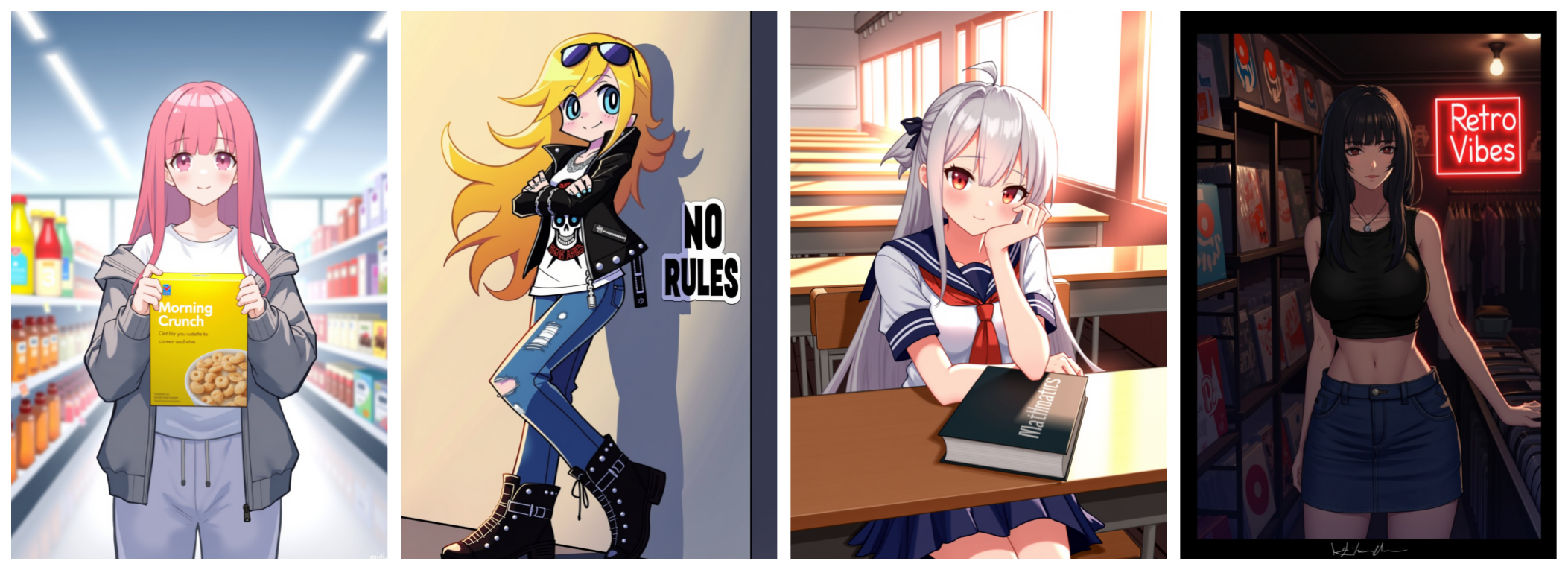}
\vspace{-2mm}
\caption{Failure analysis of standard fine-tuning. All outputs exhibit dominant prior characteristics regardless of target domain---a \emph{prior capture} phenomenon that traps the model in the Underfitting Region of \cref{fig:loss-landscape}.}
\label{fig:failure-cases}
\vspace{-2mm}
\end{figure}

As illustrated in \cref{fig:loss-landscape}, drawing on well-established perspectives from optimization geometry and generalization theory~\cite{li2018visualizing,belkin2019reconciling,kirkpatrick2017overcoming}, the optimization landscape exhibits three regimes: an \emph{Underfitting Region} where the pre-trained prior suppresses downstream updates, an \emph{OOD Region} where aggressive training causes catastrophic forgetting, and a narrow \emph{Sweet Spot} between them. Standard fine-tuning frequently falls into the underfitting regime (\cref{fig:failure-cases}), while aggressive hyperparameter tuning often overshoots into the OOD region. This directly motivates our framework: H-SD (\cref{sec:method-dataset}) amplifies the downstream signal to escape underfitting, while Tuple-wise Optimization (\cref{sec:method-tuple}) and Scale-Adaptive Modulation (\cref{sec:method-modulation}) prevent OOD drift.

\subsection{Construction of the Hierarchical Semantic Dataset}
\label{sec:method-dataset}

\begin{figure}[H]
\centering
\includegraphics[width=\linewidth]{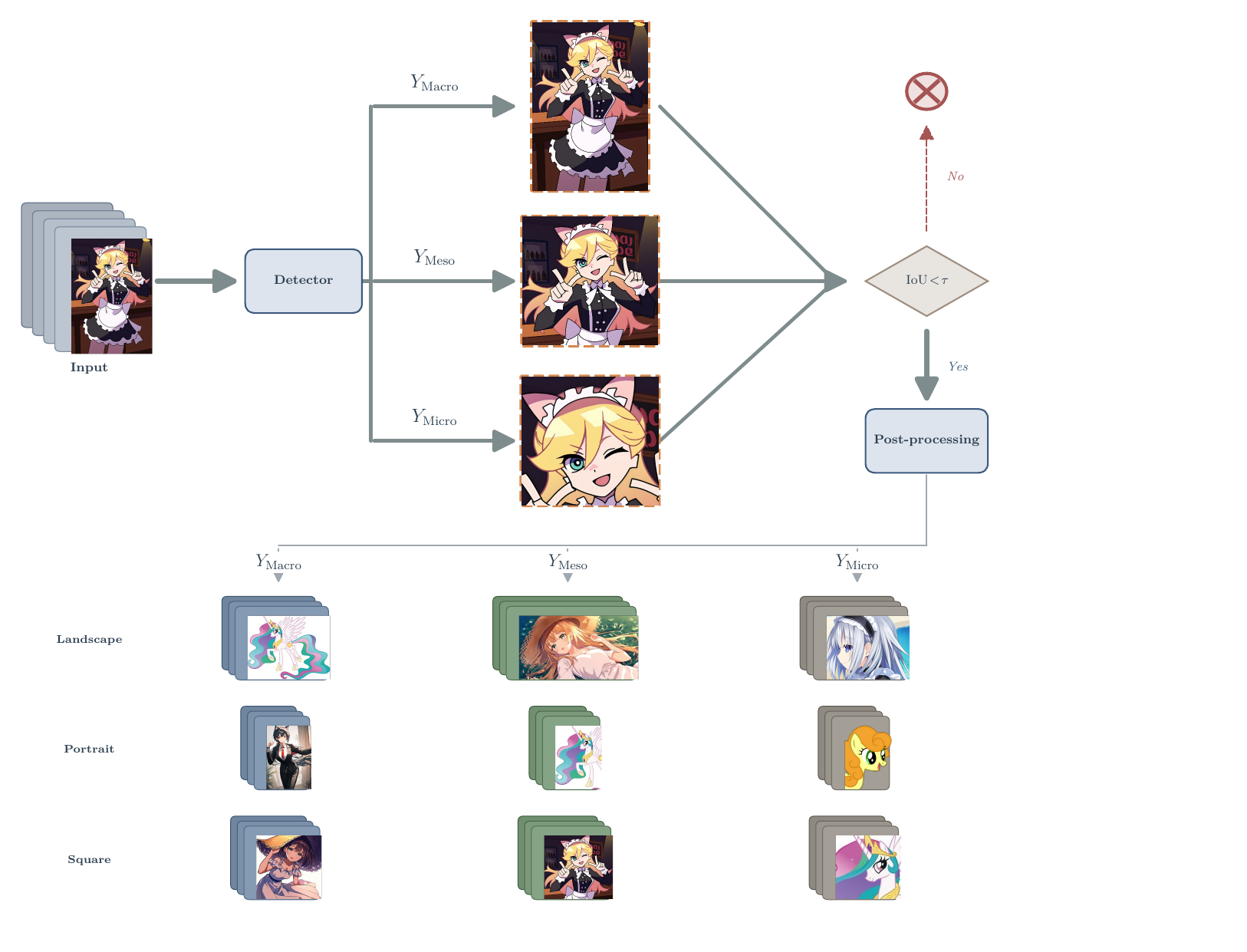}
\caption{Overview of the Hierarchical Semantic Decomposition (H-SD) pipeline. Raw images are parsed by a general-purpose detector (\eg, multimodal models, YOLO, or Grounding DINO) into three granularities ($Y_{\text{Macro}}$, $Y_{\text{Meso}}$, $Y_{\text{Micro}}$), filtered via IoU-based redundancy elimination, and post-processed into aspect-ratio-preserving resolution buckets.}
\label{fig:hsd-pipeline}
\end{figure}

To instantiate the theoretical sub-manifolds from \cref{sec:method-theory}, we implement the H-SD pipeline (\cref{fig:hsd-pipeline}). A general-purpose object detector (\eg, multimodal models, YOLO, or Grounding DINO) first parses each image into three semantic tiers---Subject (Macro), Sub-structures (Meso), and Details (Micro)---while preserving Part-Whole traceability. We then apply IoU-based filtering across hierarchies to discard spatially redundant slices, ensuring that each sub-manifold contributes distinct information to $\mathbf{\Omega}$. The retained slices are further refined through a post-processing pipeline that includes aspect-ratio-preserving downsampling and resolution bucketing~\cite{novelai2022bucketing}, amplifying local features that are typically diluted in global objectives.

\subsection{Tuple-wise Optimization Protocol}
\label{sec:method-tuple}

Although the standard MSE objective implicitly optimizes the full quadratic form $\boldsymbol{\alpha}^\top\mathbf{\Omega}\,\boldsymbol{\alpha}$, when semantic slices (\eg, Macro and Micro) are sampled into separate batches, each gradient step is usually dominated by a single scale's contribution. Alternating between such scale-biased updates produces oscillating gradient directions, a phenomenon well-documented in multi-task optimization~\cite{yu2020gradient}. To address this, we construct \emph{semantic tuples} that enforce the co-occurrence of hierarchically related slices within the same batch. For a set of $K$ semantic scales indexed by $k$, the joint loss is:
\begin{equation}
\mathcal{L}_{Tuple}(\theta) = \frac{1}{K}\sum_{k=1}^{K} \lambda_k \cdot \mathcal{L}_{CFM}(\theta;\, \mathcal{B}_k)
\label{eq:tuple-obj}
\end{equation}
where $\mathcal{B}_k$ is the subset of slices at granularity $k$ within the tuple and $\lambda_k$ is a balancing coefficient (set to $1.0$). By co-sampling all scales in a single optimization step, we attempt to engineer an intervention on the output of the vector residual field, encouraging the per-step gradient to balance contributions from both the diagonal (self-alignment) and off-diagonal (cross-scale residual correlation) terms of $\mathbf{\Omega}$.

\subsection{Scale-Adaptive Modulation}
\label{sec:method-modulation}

The Tuple-wise protocol enforces spatial consistency but does not address the spectral discrepancy between granularities: Macro slices are dominated by low-frequency geometry, while Micro slices encode high-frequency texture. A uniform optimization schedule conflates these distinct signal bands, introducing cross-frequency noise. Prior work has shown that the noise-level sampling strategy significantly affects both convergence speed and final generation quality in diffusion and flow-based models~\cite{hang2023efficient,karras2022elucidating}. We resolve this via two architecture-specific mechanisms (see Suppl.~D for derivations):

\subsubsection{Frequency-Time Alignment (For DiT Architectures).}
In MM-DiT~\cite{peebles2023scalable} models (\eg, FLUX.1), where time-step sampling follows a Logit-Normal distribution~\cite{esser2024scaling}, global structure emerges at high noise levels ($t \to 1$) and fine details at low levels ($t \to 0$). We condition the time-step sampling $\pi(t)$ on scale $S$, shifting probability mass toward $t \to 1$ for Macro (structural focus) and toward $t \to 0$ for Micro (textural refinement).

\subsubsection{SNR-Aware Reweighting (For U-Net Architectures).}
For U-Net models (\eg, SDXL~\cite{podell2024sdxl}), we modulate the loss weight $\omega(t, S)$ by granularity, building on insights from Zero Terminal SNR~\cite{lin2024common}: increasing the weight for Micro slices in high-SNR regimes to sustain gradient supervision on fine details, and reducing it for Macro slices to prevent overfitting to compression artifacts.

\begin{figure}[H]
\centering
\includegraphics[width=\linewidth]{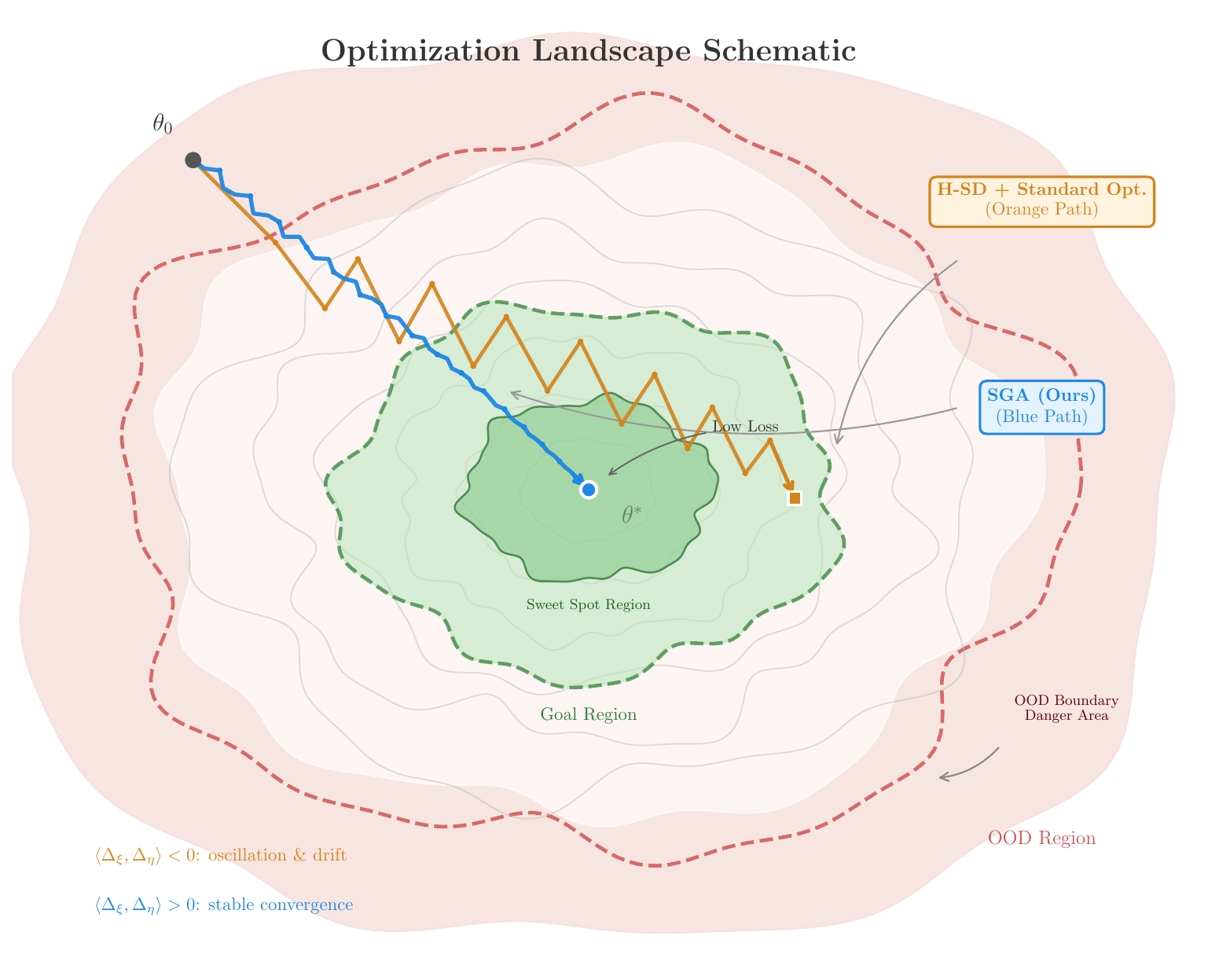}
\caption{Optimization trajectories on the H-SD-augmented loss landscape. \textbf{H-SD + Standard Opt.} (orange): after signal amplification via H-SD, vanilla CFM training still suffers from scale-biased gradient updates, causing persistent oscillation and drift toward the OOD boundary. \textbf{SGA} (blue): by adding Tuple-wise co-sampling and Scale-Adaptive Modulation, SGA stabilizes the descent and converges to the Goal Region. The residual correlation sign $\langle \Delta_\xi, \Delta_\eta \rangle$ determines whether cross-scale interactions are constructive ($>0$, stable) or destructive ($<0$, oscillation).}
\label{fig:opt-trajectory}
\end{figure}

\section{Experiments}
\label{sec:experiments}
\vspace{-1mm}
\subsection{Experimental Setup}
\label{sec:exp-setup}

\noindent\textbf{Datasets \& Scope.}\;
We curated diverse GDA datasets with 100 to several hundred images per domain. For the MM-DiT architecture (FLUX), we evaluate on 6 datasets; for the U-Net architecture (Animagine XL 3.1), we select 3 datasets to validate cross-architecture generalizability (full experimental setup in Suppl.~F).

\noindent\textbf{Architectures \& Baselines.}\;
We evaluate SGA on two frameworks: (1)~\textbf{DiT:} FLUX~\cite{flux2024} with DoRA~\cite{liu2024dora} and AdamW~\cite{kingma2015adam,loshchilov2019decoupled}; (2)~\textbf{U-Net:} Animagine XL~3.1 (SDXL-based~\cite{podell2024sdxl}) with LoCon~\cite{yeh2024navigating} and Lion~\cite{chen2024symbolic}. We compare against standard fine-tuning with Aspect Ratio Bucketing (ARB)~\cite{novelai2022bucketing} and horizontal flipping; SGA is orthogonal to ARB and fully compatible with its resolution logic. All hyperparameters are identical between baseline and SGA.

\noindent\textbf{Training \& Implementation.}\;
Learning rates follow community-established defaults. Following the critical batch size principle~\cite{mccandlish2018empirical}, we set batch sizes according to model scale: FLUX uses 8, Animagine XL~3.1 uses 2. Training budgets are reported as multiples of a reference GPU-time $N_1$ calibrated per dataset. All experiments run on NVIDIA RTX PRO 6000 Blackwell GPUs (96\,GB GDDR7).

\vspace{-1mm}
\subsection{Qualitative Results}
\label{sec:exp-qualitative}
\vspace{-1mm}

\begin{figure}[H]
\centering
\setlength{\tabcolsep}{0.5mm}
\footnotesize
\begin{tabular}{@{}c ccccc@{}}
& \textbf{Ref.} & \textbf{BL ($1.0\,N_1$)} & \textbf{BL ($1.5\,N_1$)} & \textbf{SGA ($1.0\,N_1$)} & \textbf{SGA ($1.5\,N_1$)} \\[1mm]
\rotatebox[origin=c]{90}{\small\color{fluxblue}\textbf{FLUX}\;\color{black}$\boldsymbol{\delta}$} &
\grimg{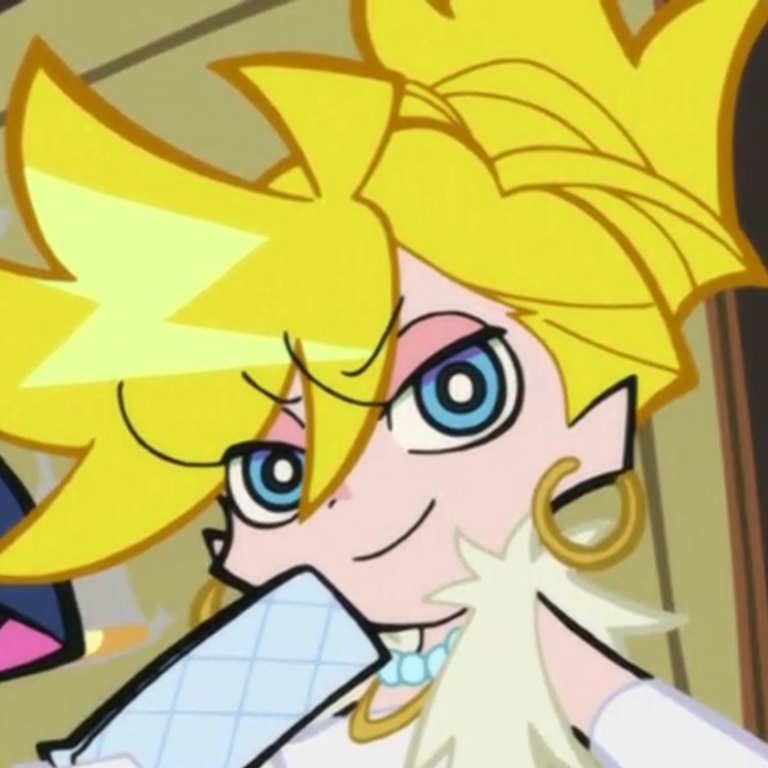}{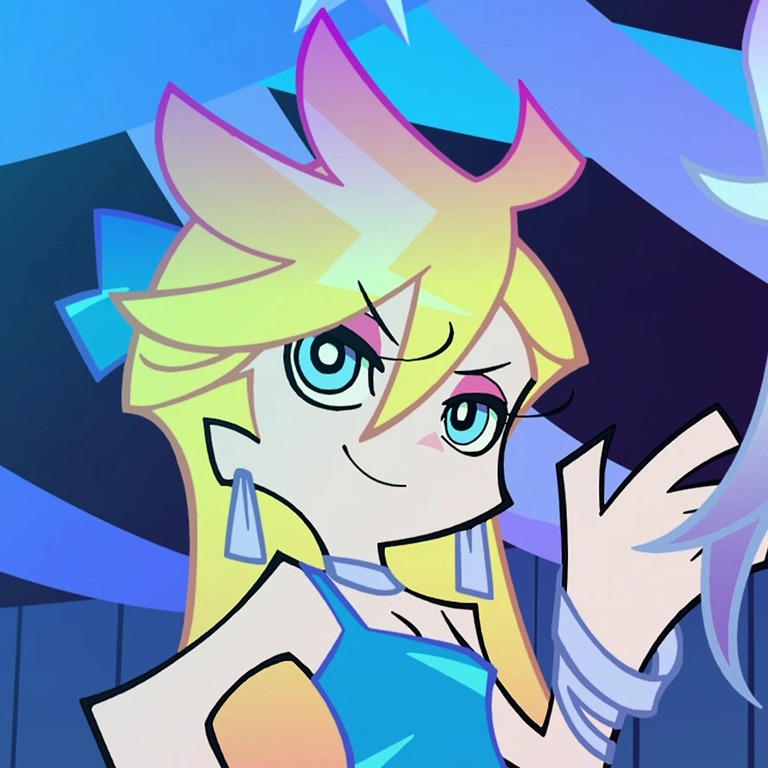} &
\gpimg{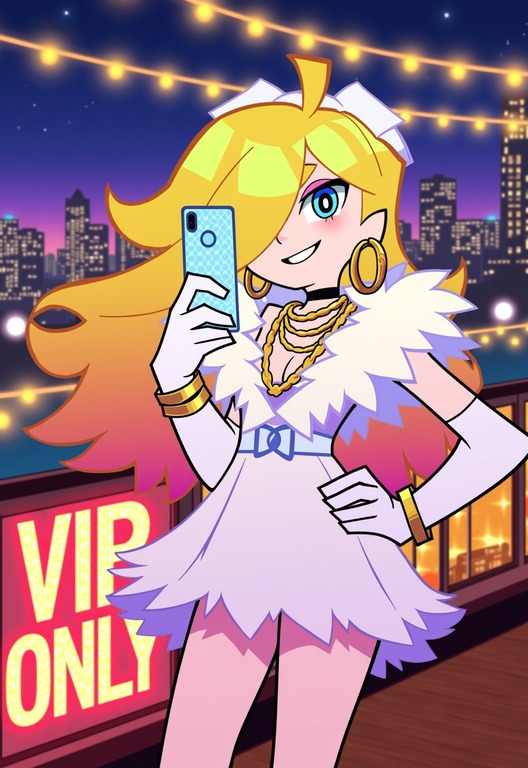} &
\gpimg{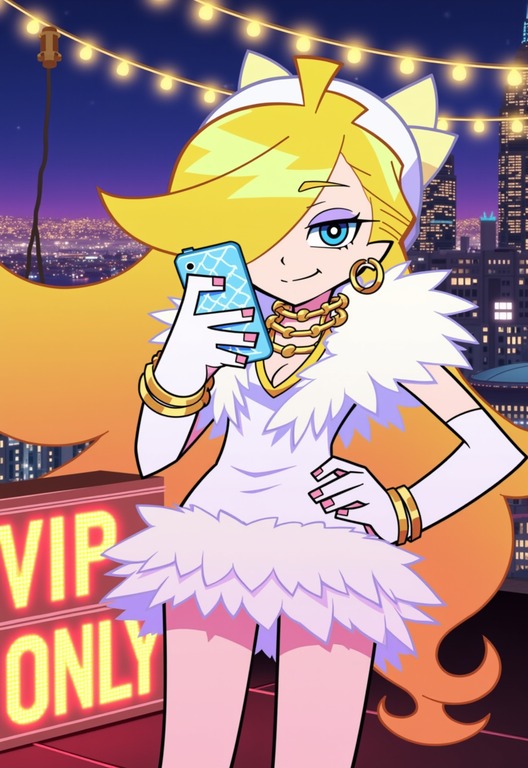} &
\gpimg{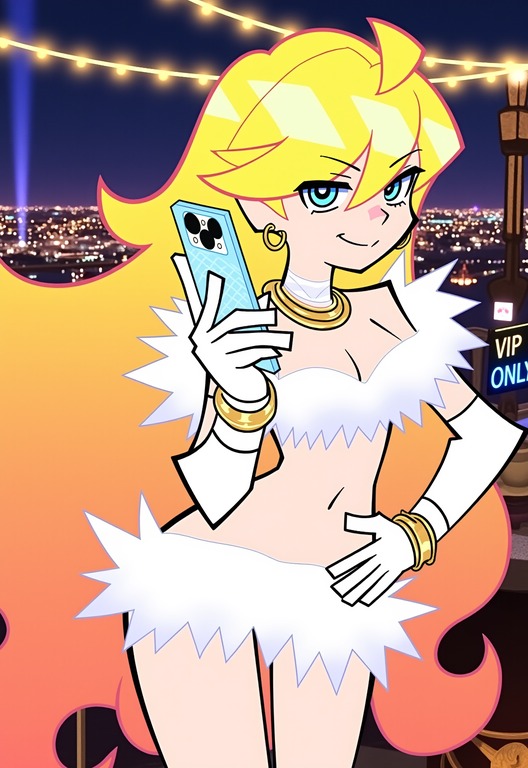} &
\gpimg{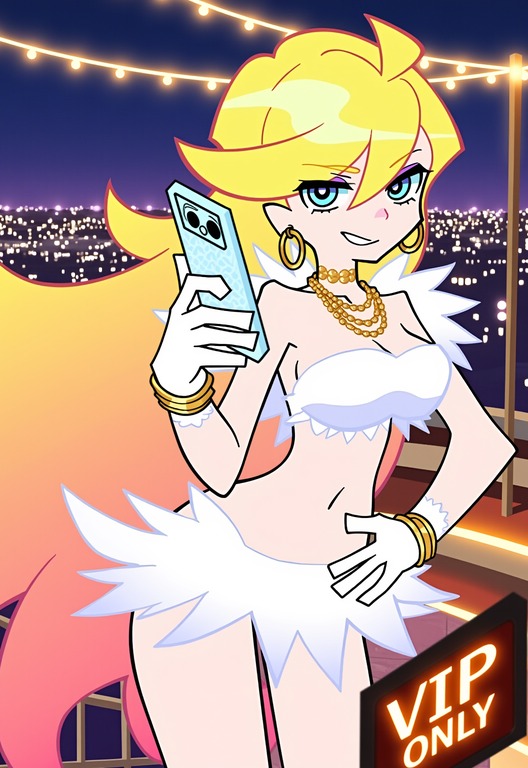} \\
\noalign{\vspace{1mm}}
\rotatebox[origin=c]{90}{\small\color{fluxblue}\textbf{FLUX}\;\color{black}$\boldsymbol{\varepsilon}$} &
\grimg{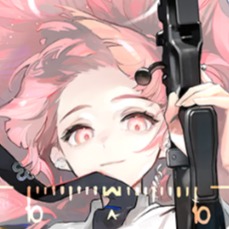}{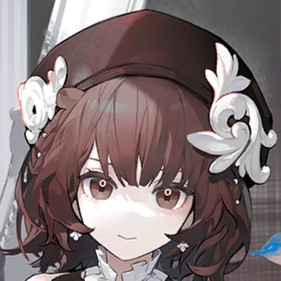} &
\gpimg{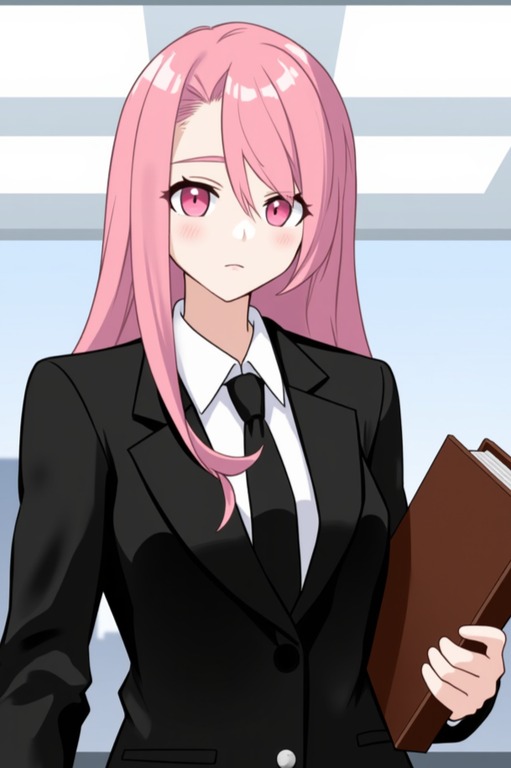} &
\gpimg{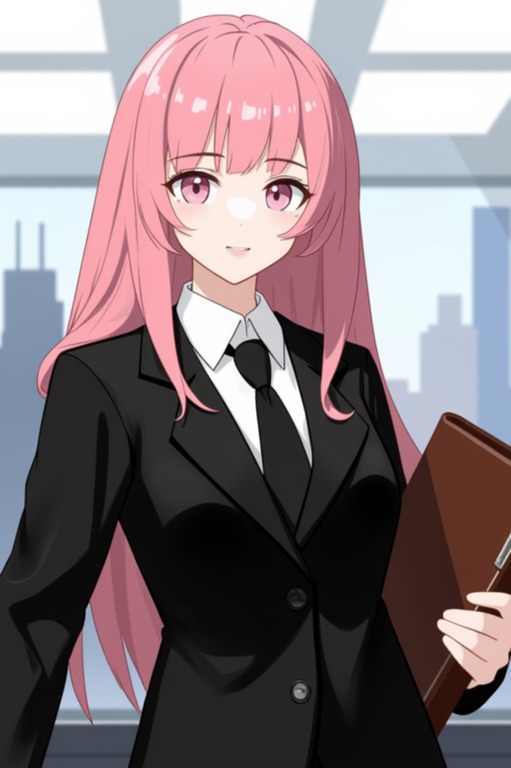} &
\gpimg{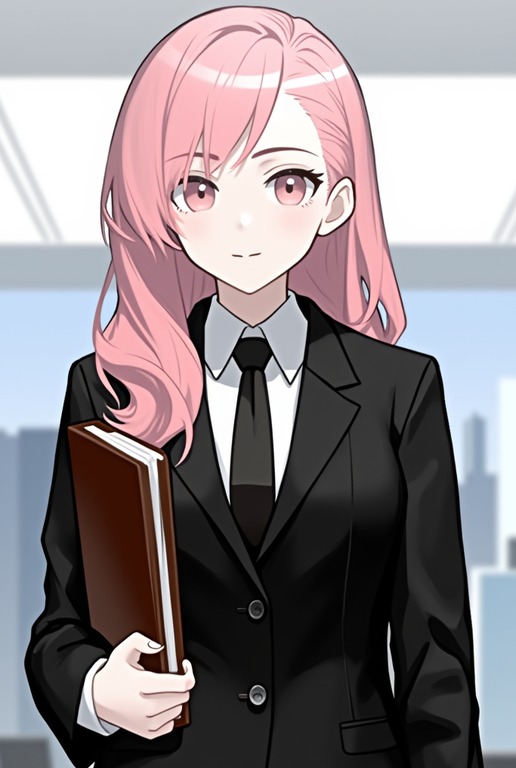} &
\gpimg{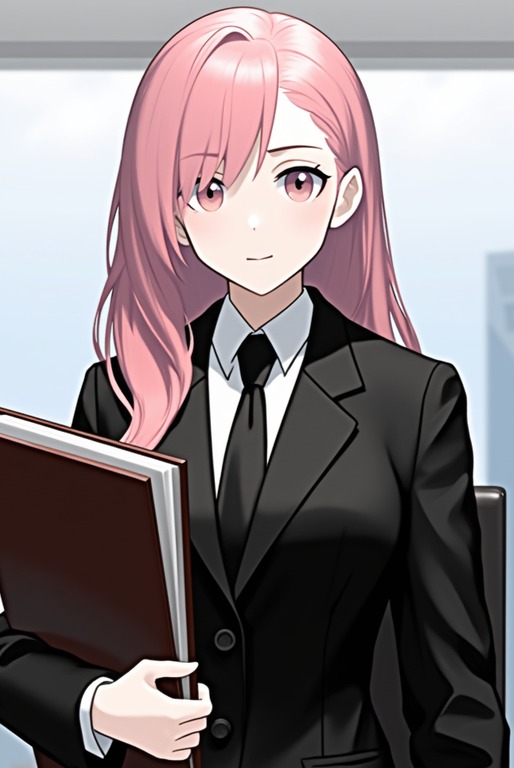} \\
\noalign{\vspace{1mm}}
\rotatebox[origin=c]{90}{\small\color{fluxblue}\textbf{FLUX}\;\color{black}$\boldsymbol{\zeta}$} &
\grimg{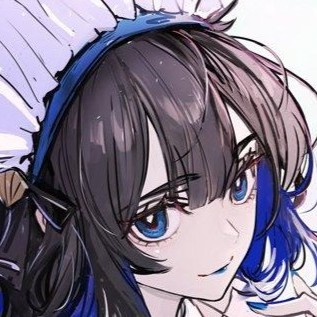}{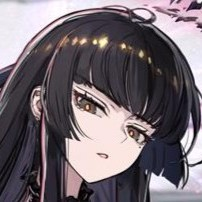} &
\gpimg{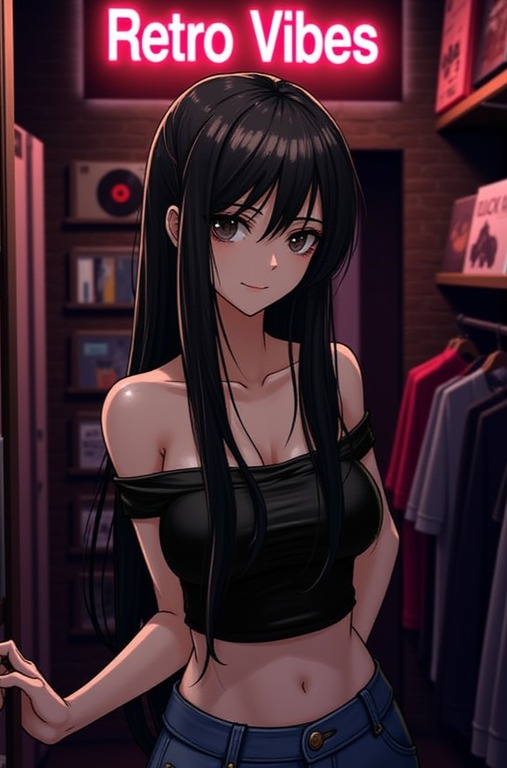} &
\gpimg{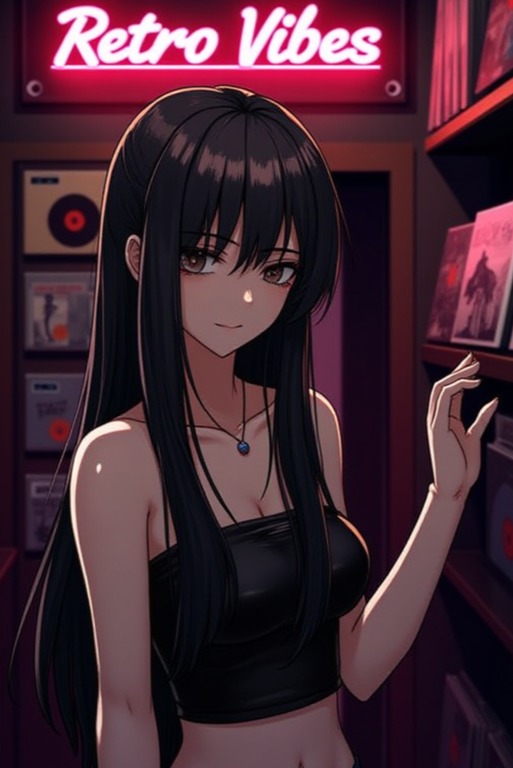} &
\gpimg{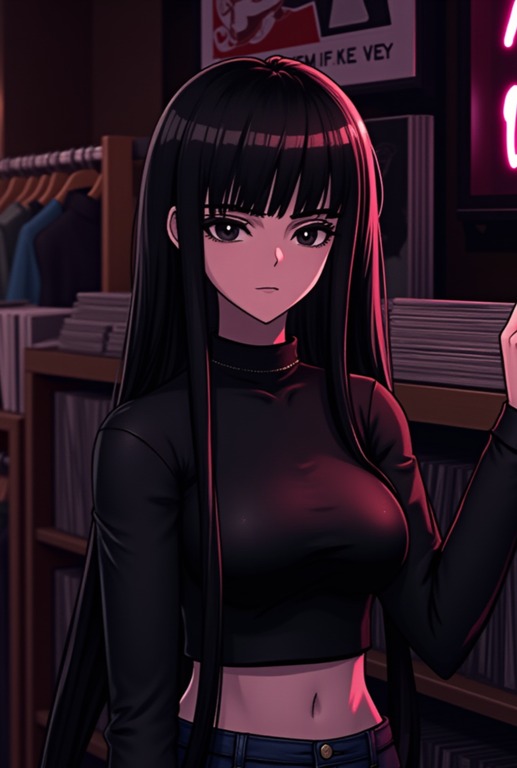} &
\gpimg{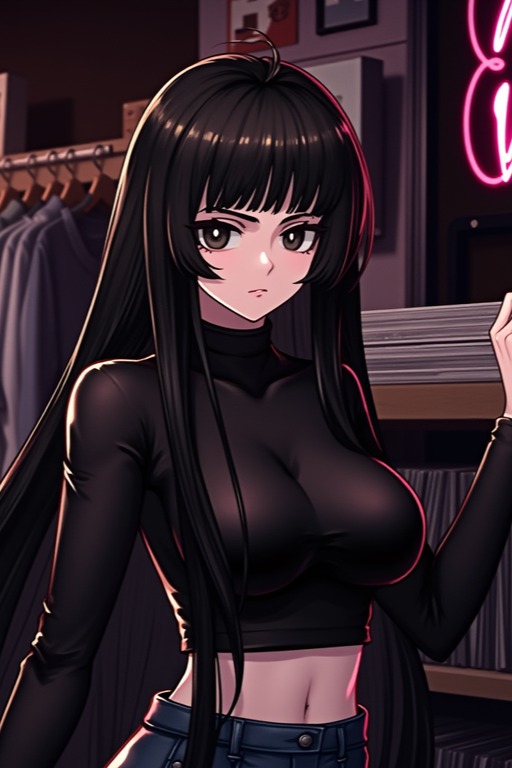} \\
\end{tabular}
\caption{\textbf{Qualitative Comparison (FLUX-only).} Domains $\delta$, $\varepsilon$, and $\zeta$, evaluated exclusively on FLUX (portrait layout).}
\label{fig:qualitative-flux-only}
\end{figure}

\begin{figure}[!t]
\centering
\setlength{\tabcolsep}{0.5mm}
\footnotesize
\begin{tabular}{@{}c ccccc@{}}
& \textbf{Ref.} & \textbf{BL ($1.0\,N_1$)} & \textbf{BL ($1.5\,N_1$)} & \textbf{SGA ($1.0\,N_1$)} & \textbf{SGA ($1.5\,N_1$)} \\[1mm]
\multirow{2}{*}{\smash{\rotatebox[origin=c]{90}{\scriptsize\color{sdxlred}\textbf{SDXL}\;\color{fluxblue}\textbf{FLUX}\;\color{black}$\boldsymbol{\alpha}$}}} &
\gimg{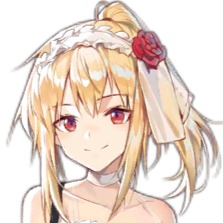} &
\gimg{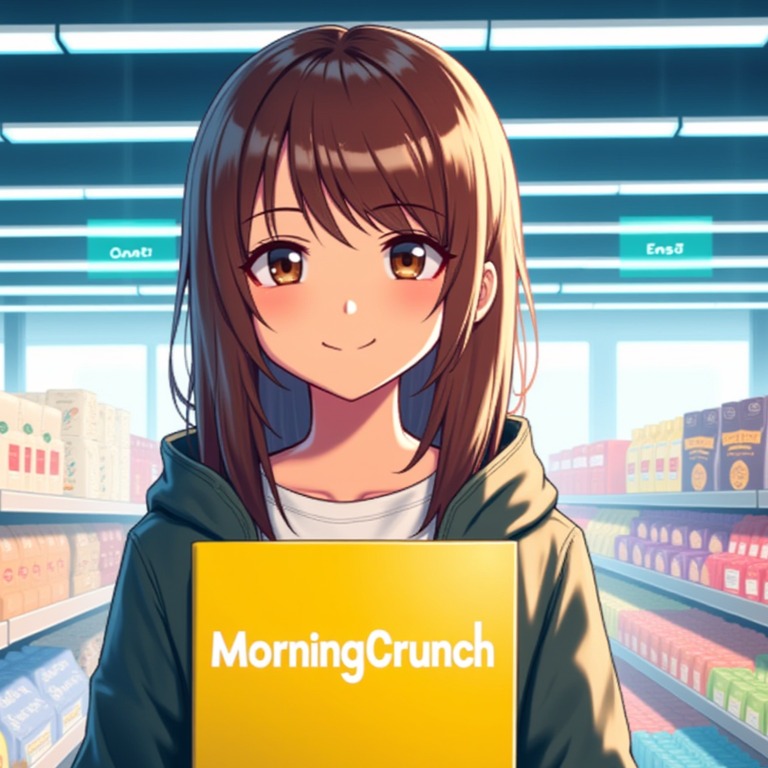} &
\gimg{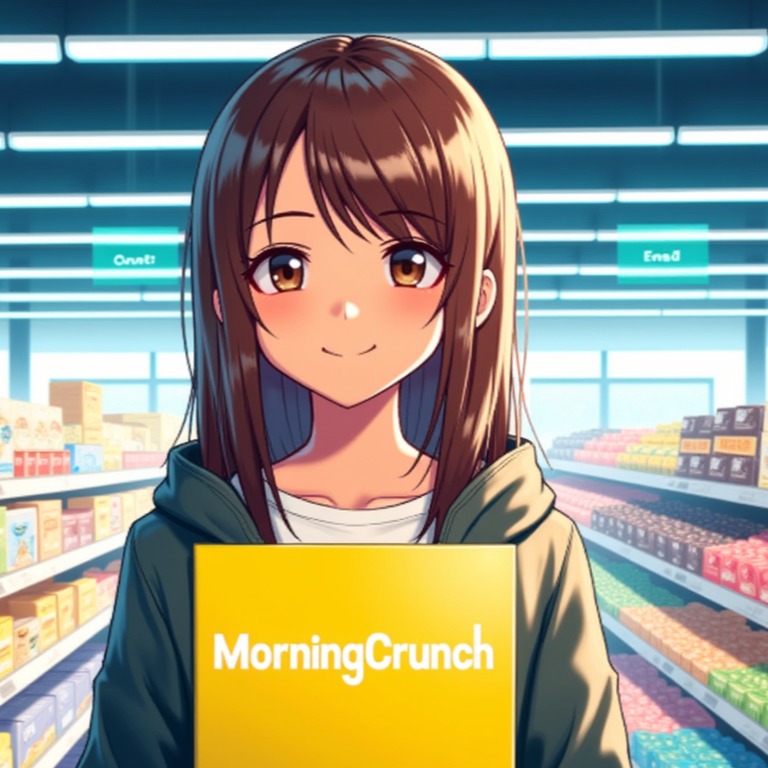} &
\gimg{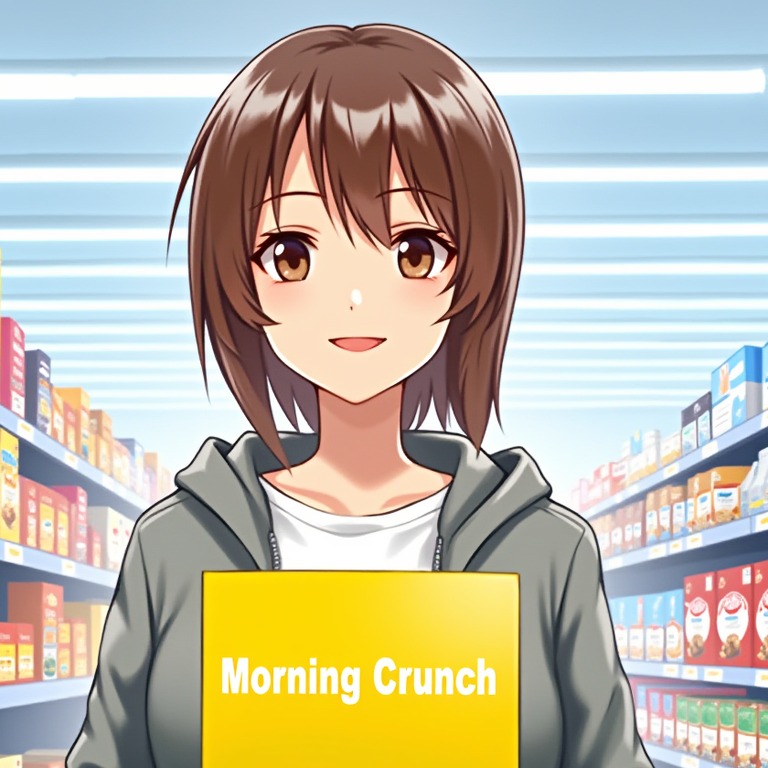} &
\gimg{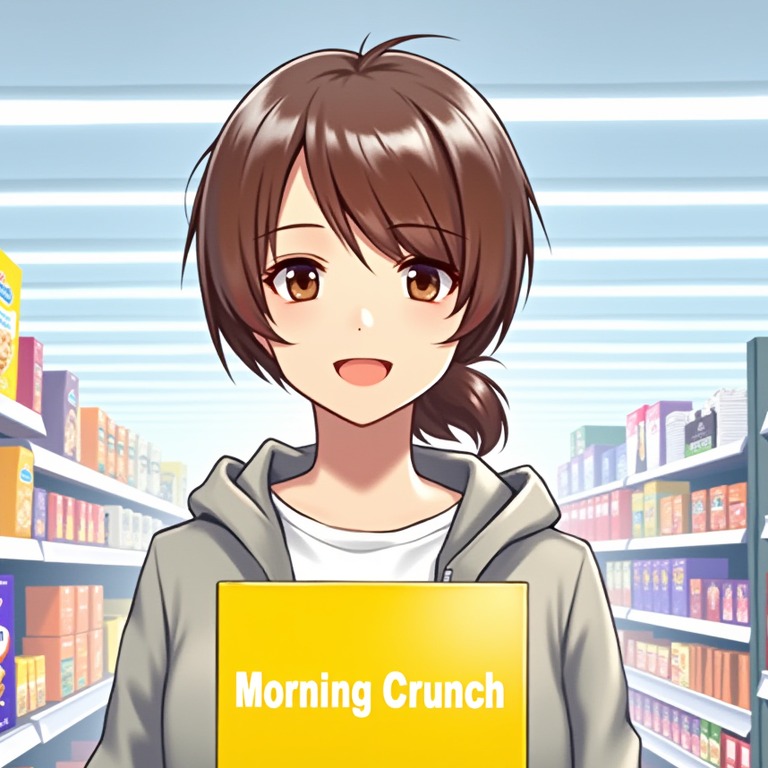} \\[0.3mm]
&
\gimg{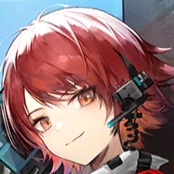} &
\gimg{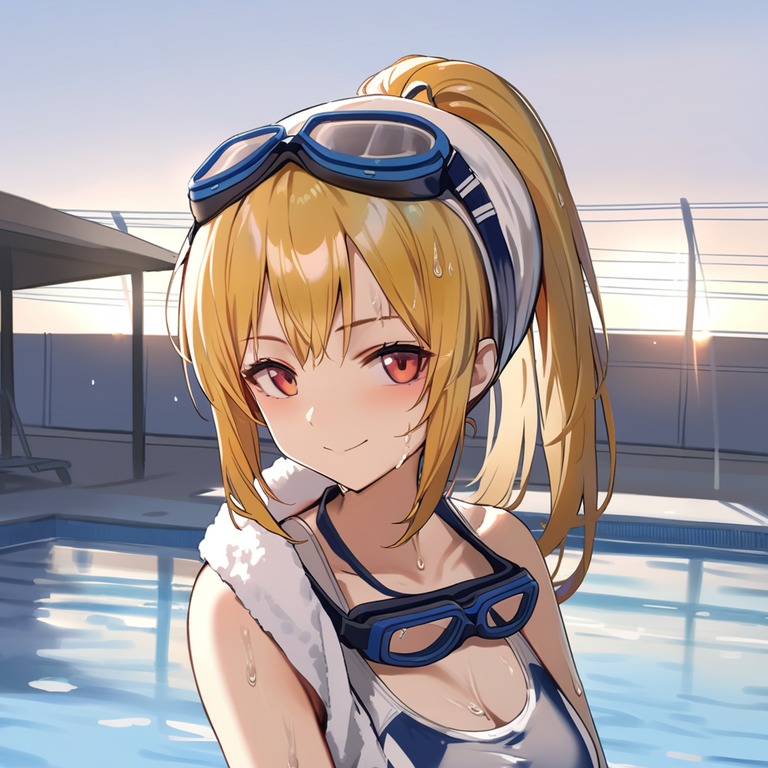} &
\gimg{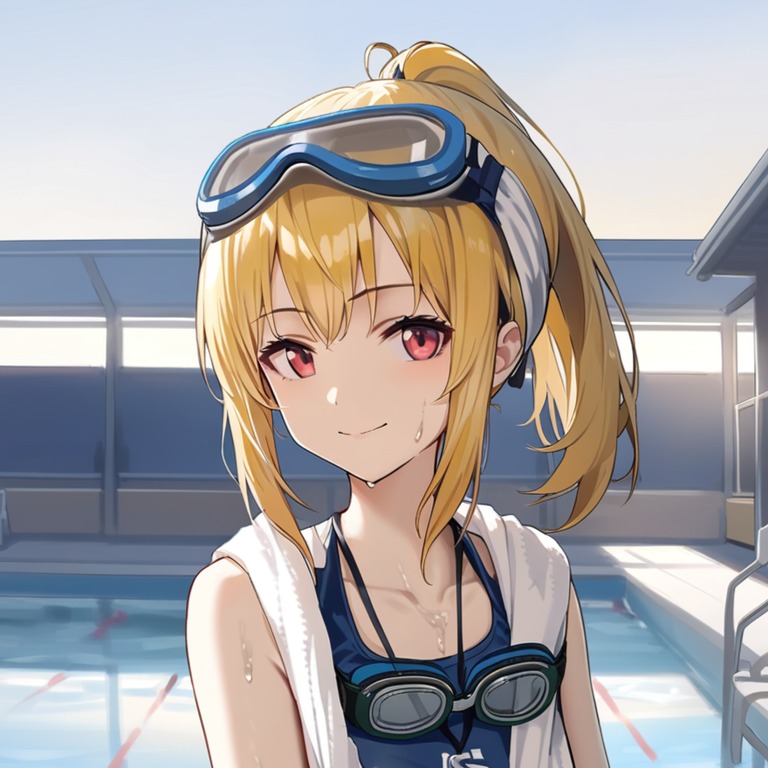} &
\gimg{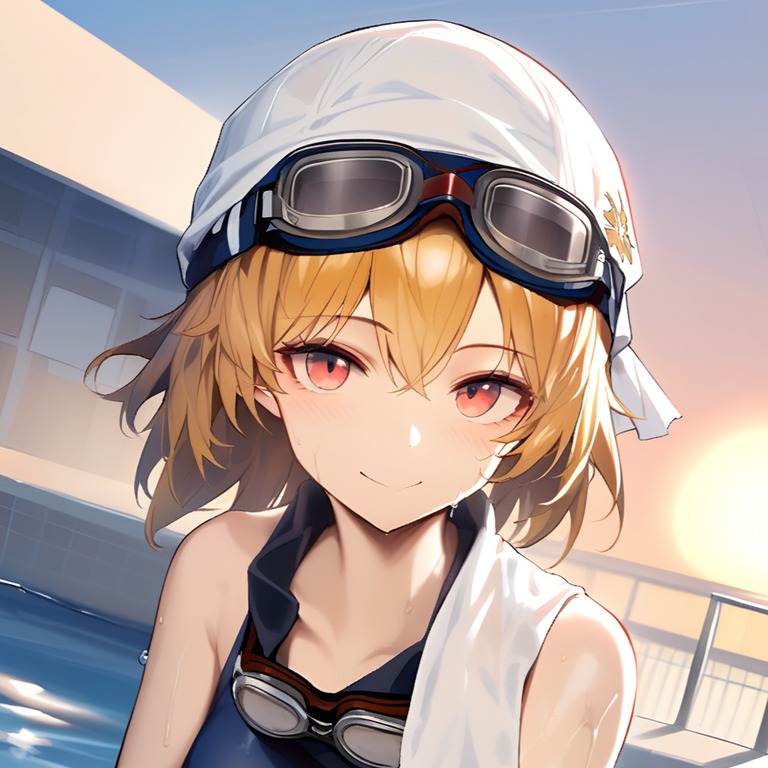} &
\gimg{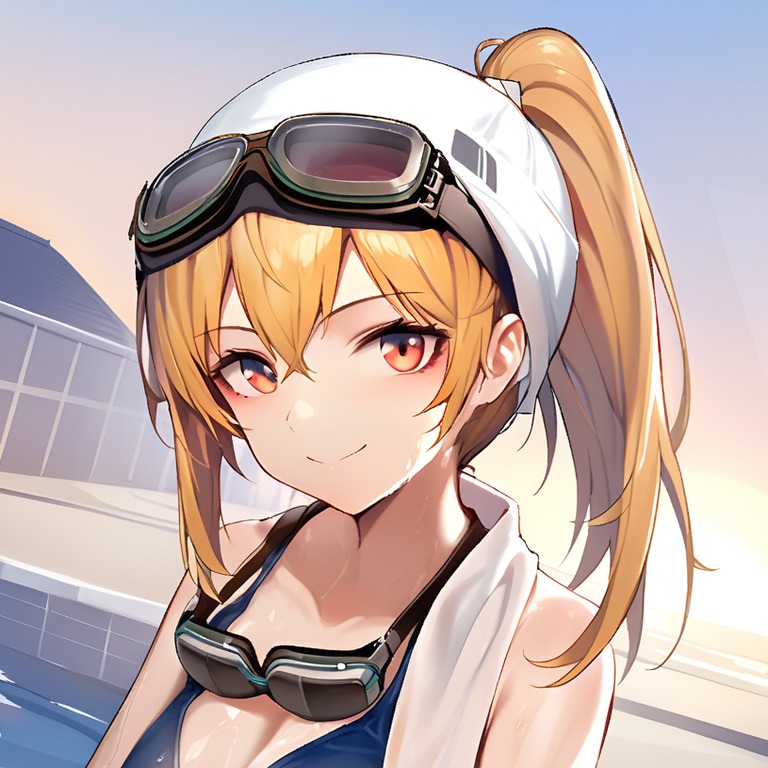} \\[1mm]
\multirow{2}{*}{\smash{\rotatebox[origin=c]{90}{\scriptsize\color{sdxlred}\textbf{SDXL}\;\color{fluxblue}\textbf{FLUX}\;\color{black}$\boldsymbol{\beta}$}}} &
\gimg{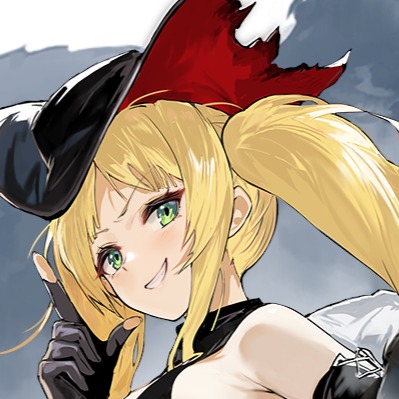} &
\gimg{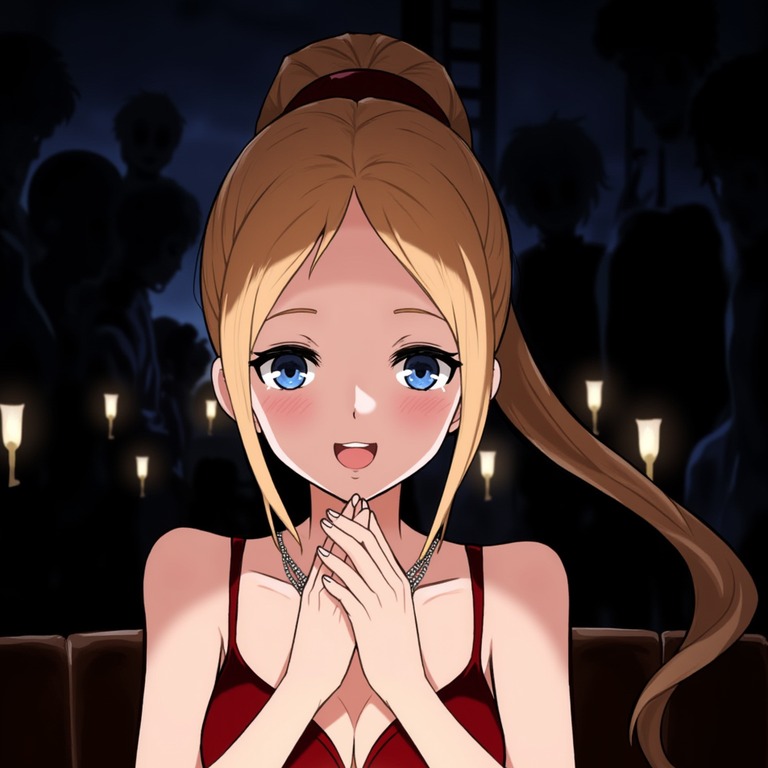} &
\gimg{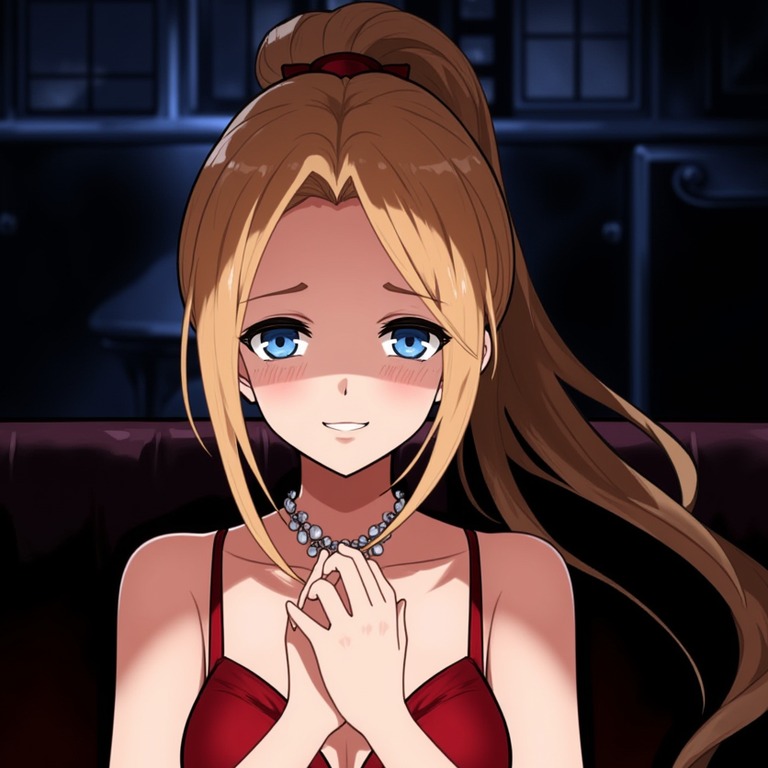} &
\gimg{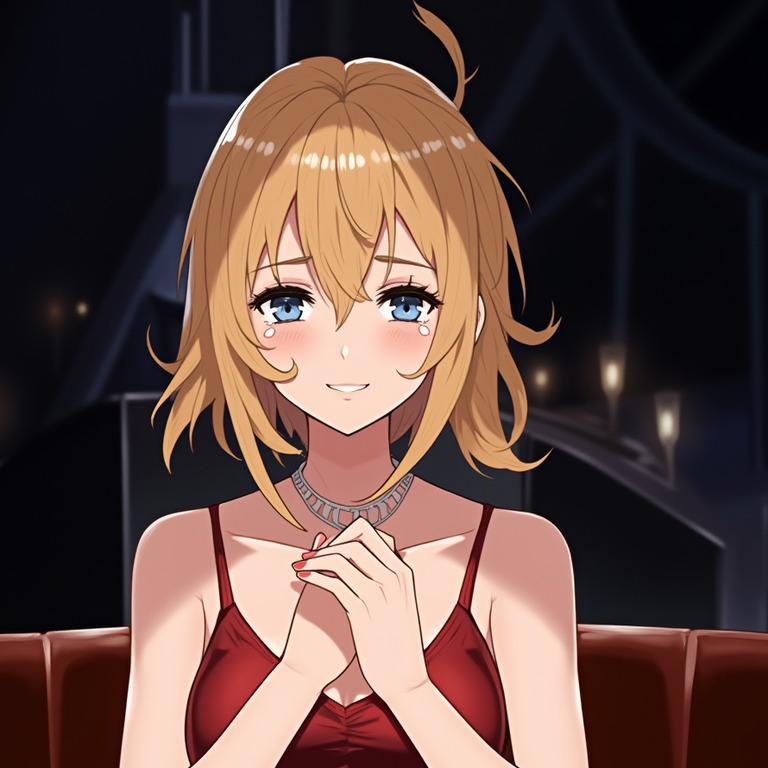} &
\gimg{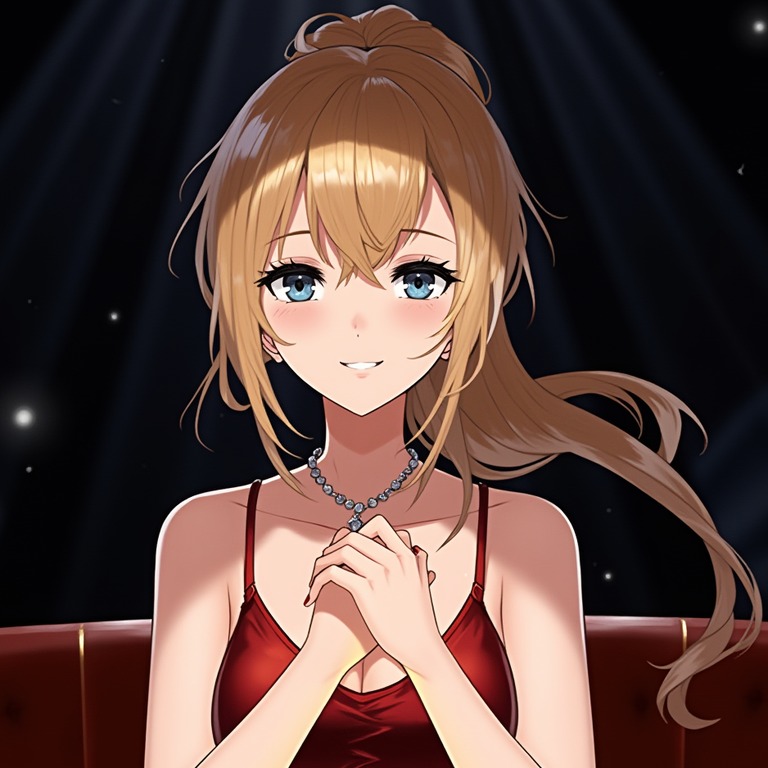} \\[0.3mm]
&
\gimg{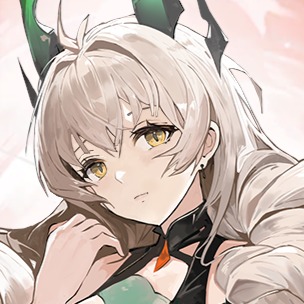} &
\gimg{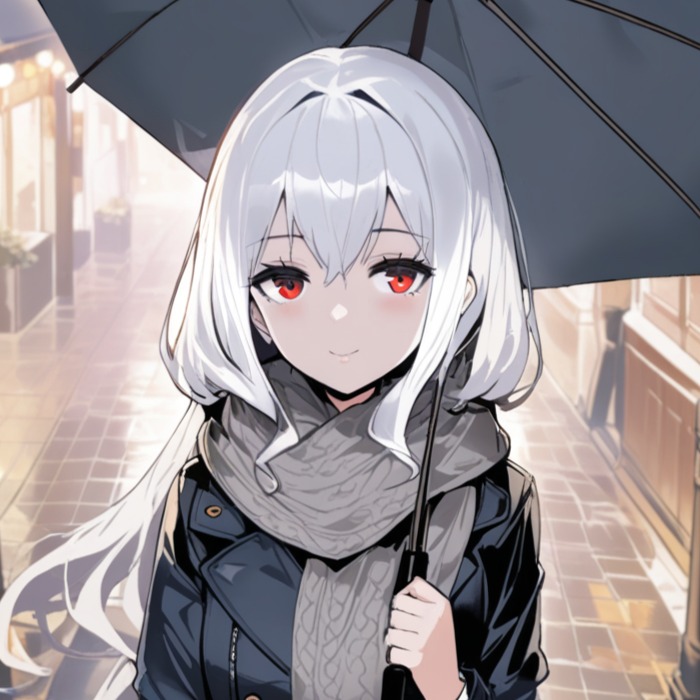} &
\gimg{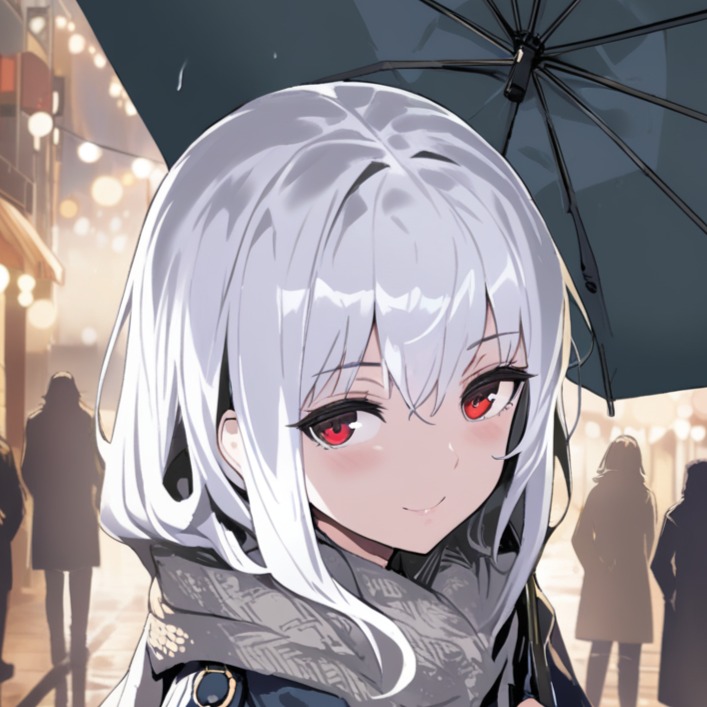} &
\gimg{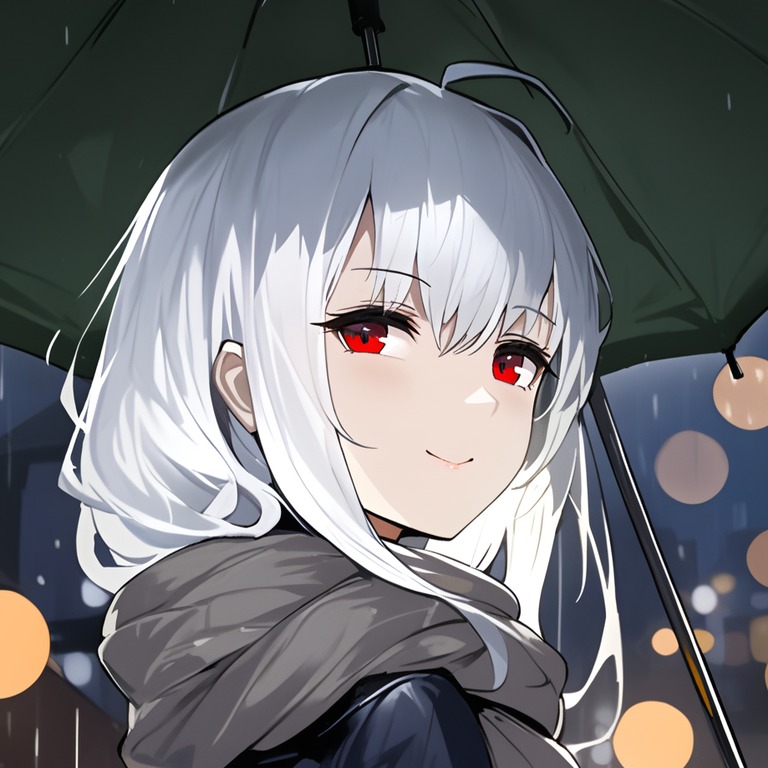} &
\gimg{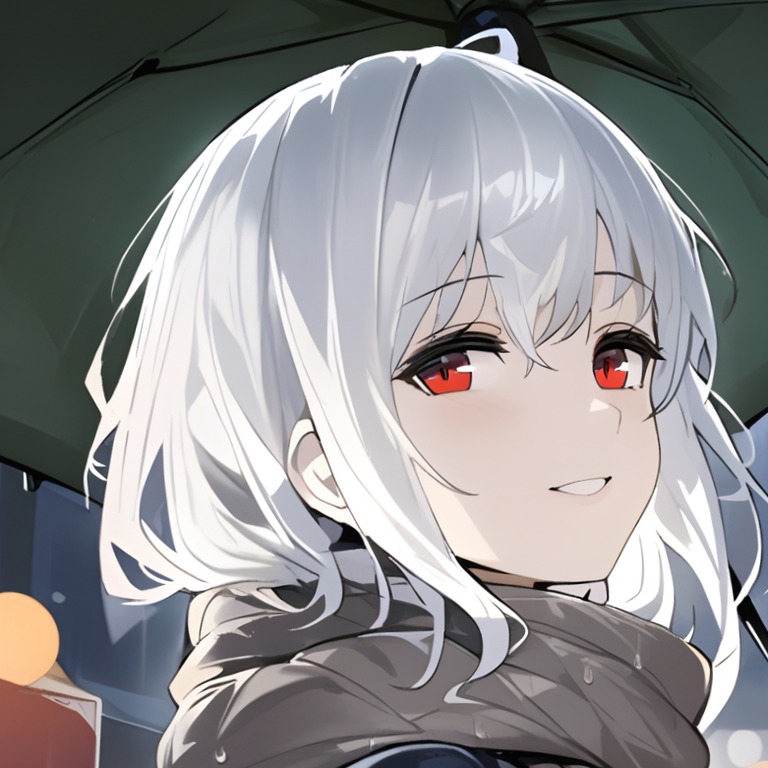} \\[1mm]
\multirow{2}{*}{\smash{\rotatebox[origin=c]{90}{\scriptsize\color{sdxlred}\textbf{SDXL}\;\color{fluxblue}\textbf{FLUX}\;\color{black}$\boldsymbol{\gamma}$}}} &
\gimg{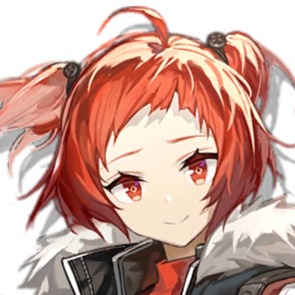} &
\gimg{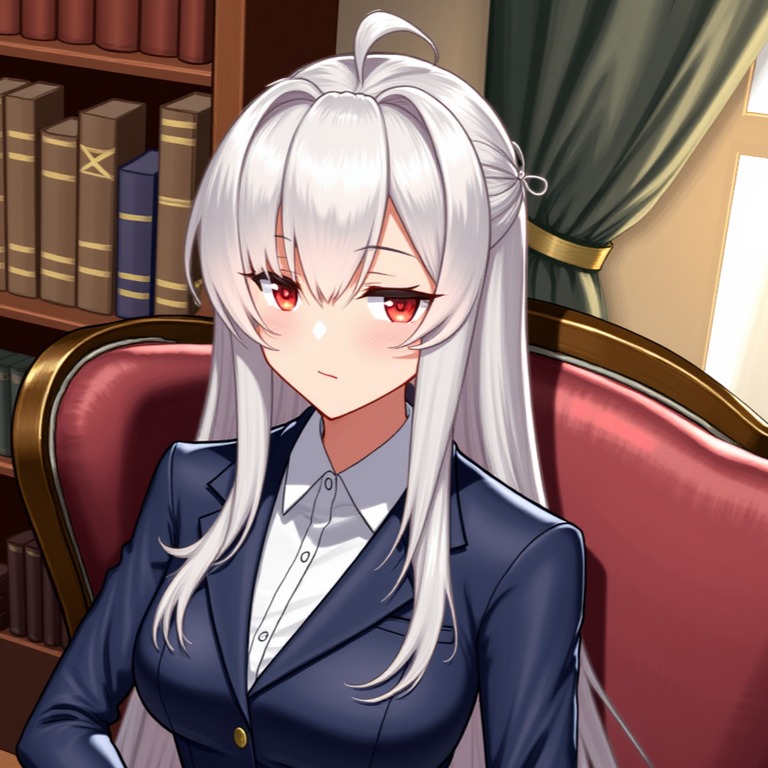} &
\gimg{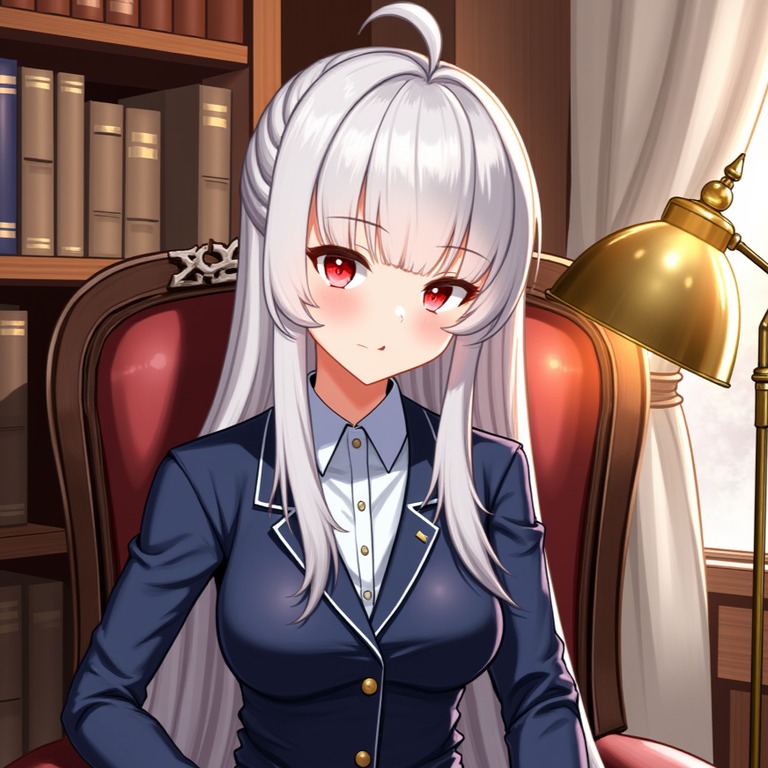} &
\gimg{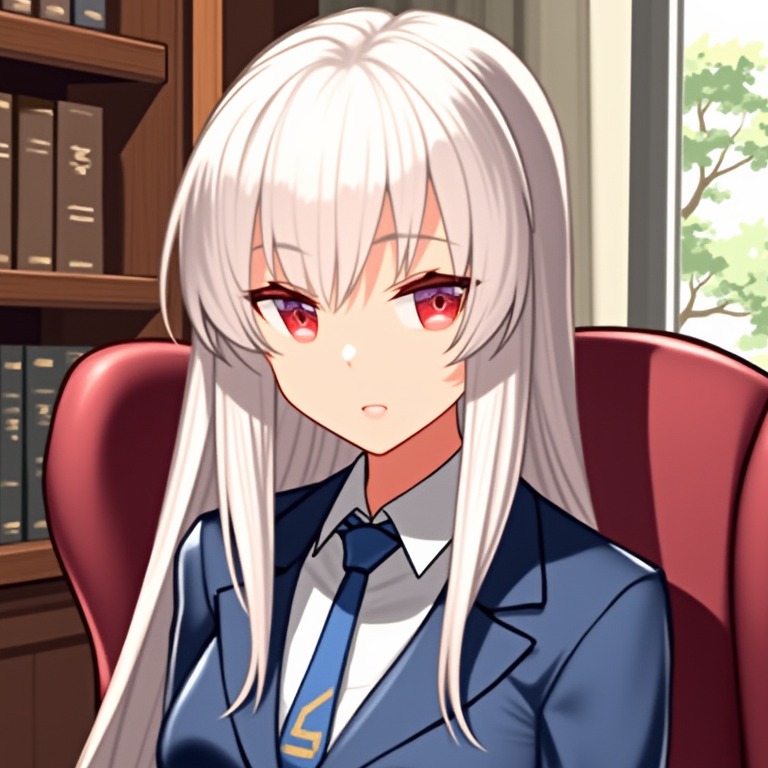} &
\gimg{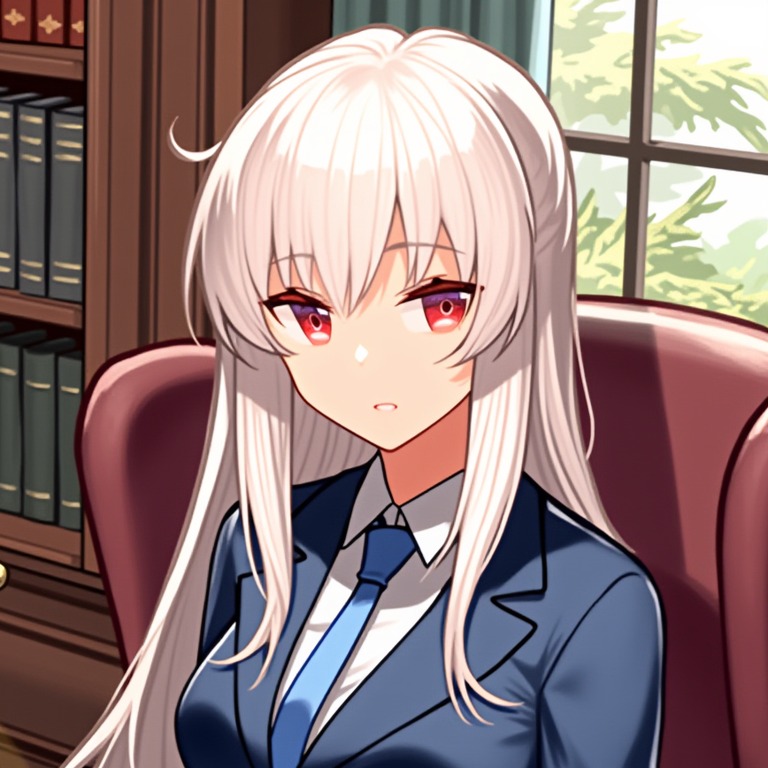} \\[0.3mm]
&
\gimg{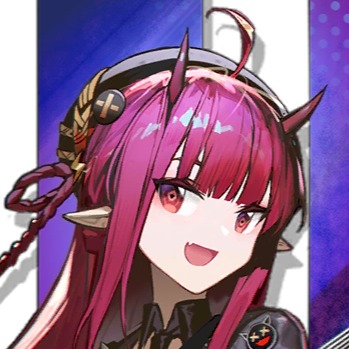} &
\gimg{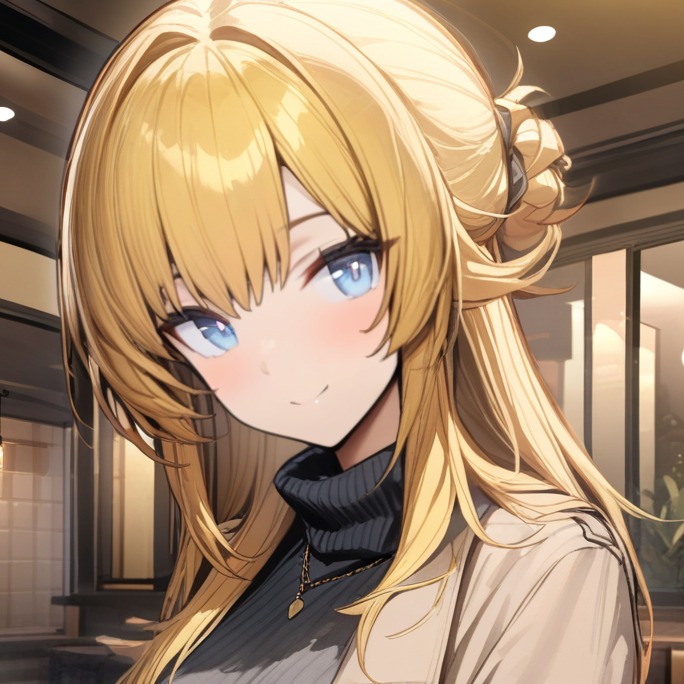} &
\gimg{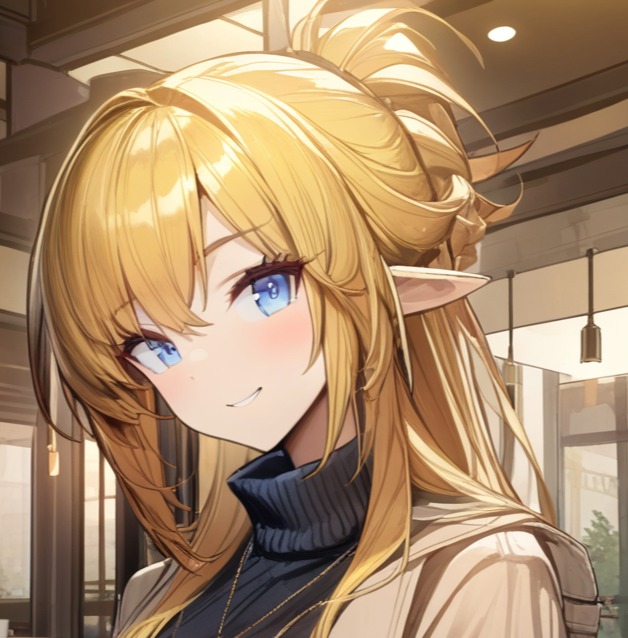} &
\gimg{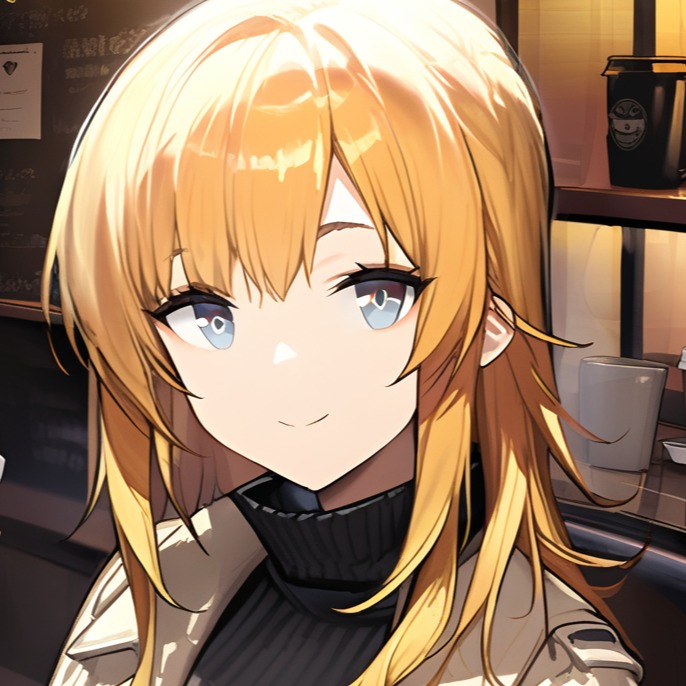} &
\gimg{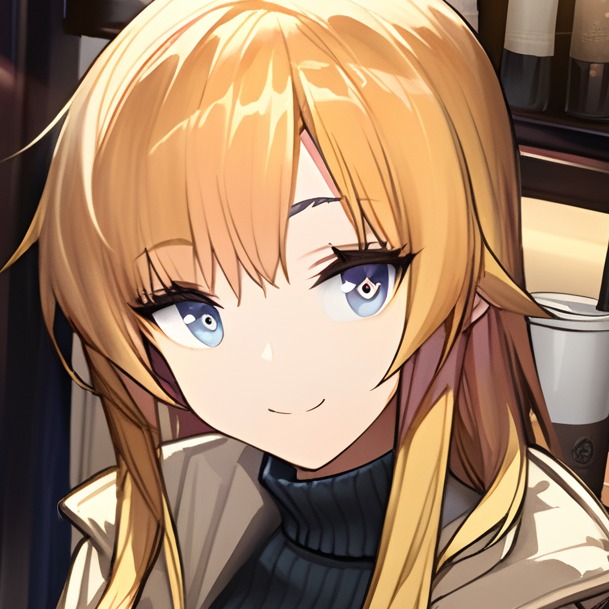} \\
\end{tabular}
\caption{\textbf{Qualitative Comparison (Cross-Architecture).} Domains $\alpha$--$\gamma$, evaluated on both SDXL and FLUX. Reference images (left) define each target domain.}
\label{fig:qualitative-shared}
\end{figure}

We present visual comparisons across six GDA domains (\cref{fig:qualitative-flux-only,fig:qualitative-shared}).


\subsection{Evaluation Protocol}
\label{sec:exp-metrics}

For each dataset, we generate outputs from four variants---Baseline and SGA at $1.0\,N_1$ and $1.5\,N_1$---using 8 domain-specific prompts with $\geq$5 seeds each, yielding $\geq$40 samples per variant.

\subsubsection{Primary Metrics: Perceptual Evaluation.}
Both evaluator types rank the four outputs from lowest to highest quality:
\begin{itemize}
\item \textbf{LLM Judge:} Following the LLM-as-a-Judge paradigm~\cite{zheng2023judging}, which has been extended to visual generation evaluation~\cite{lu2023llmscore,ku2024viescore}, we employ GPT~5.2 with a specialized system prompt evaluating GDA fidelity across three dimensions: geometric and symbolic accuracy, rendering quality, and overall aesthetic coherence.
\item \textbf{Human Evaluation:} Following established protocols for verifiable and reproducible perceptual assessment~\cite{otani2023verifiable,xu2024imagereward}, we conduct blind user studies with 20 participants recruited via the Prolific platform and domain-specific communities.
\end{itemize}
\noindent We report \emph{1st-place rate} and \emph{aggregate rank distribution} as primary metrics.

\subsection{Quantitative Results}
\label{sec:exp-quantitative}

\begin{figure}[!t]
  \centering
  \includegraphics[width=\linewidth]{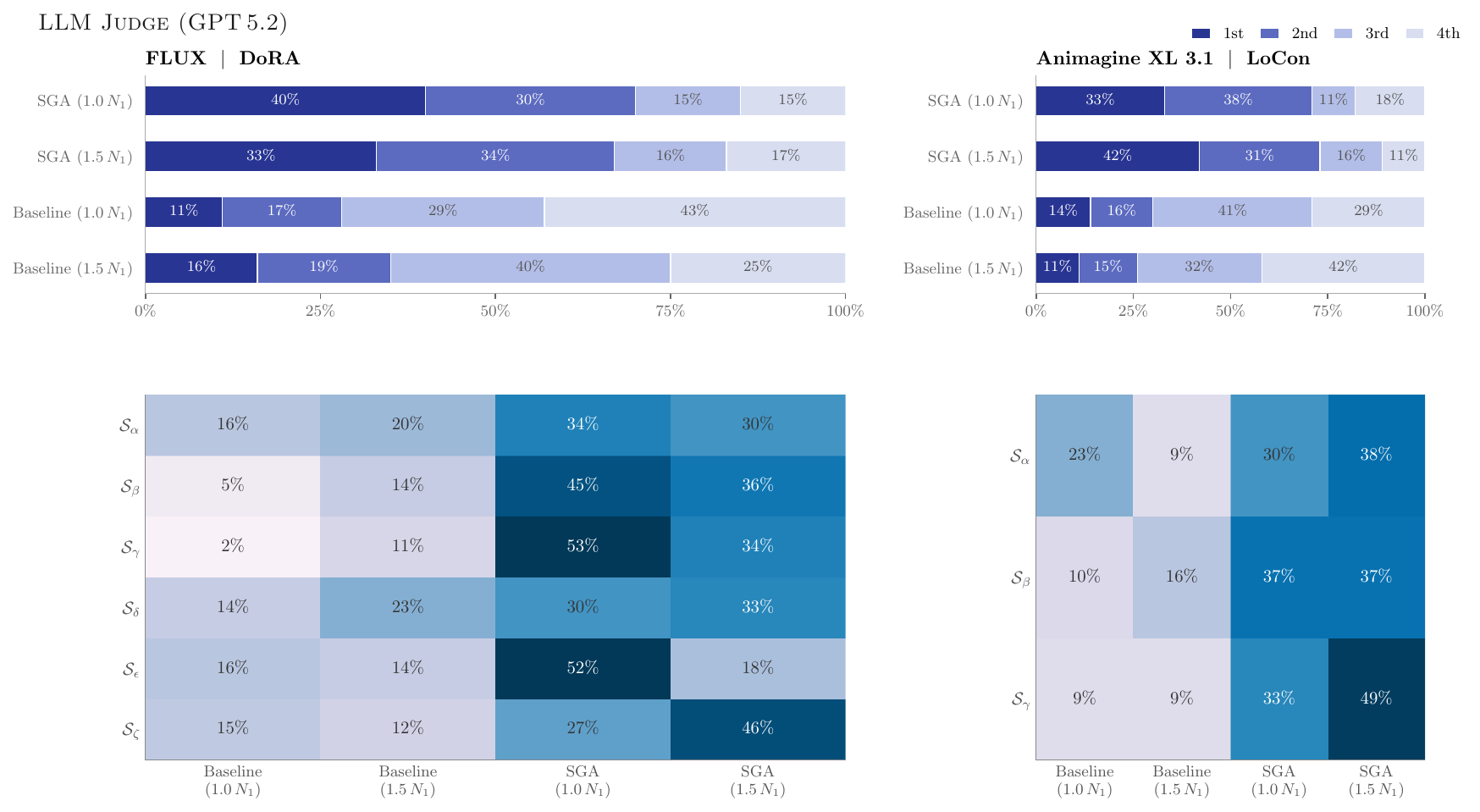}
  \caption{\textbf{Quantitative Evaluation --- LLM Judge (GPT~5.2).} Aggregate rank distribution (left) and per-dataset 1st-place rate heatmap (right). SGA variants achieve favourable rankings across both architectures and all stylistic domains.}
  \label{fig:eval-llm}
\end{figure}

\begin{figure}[!t]
  \centering
  \includegraphics[width=\linewidth]{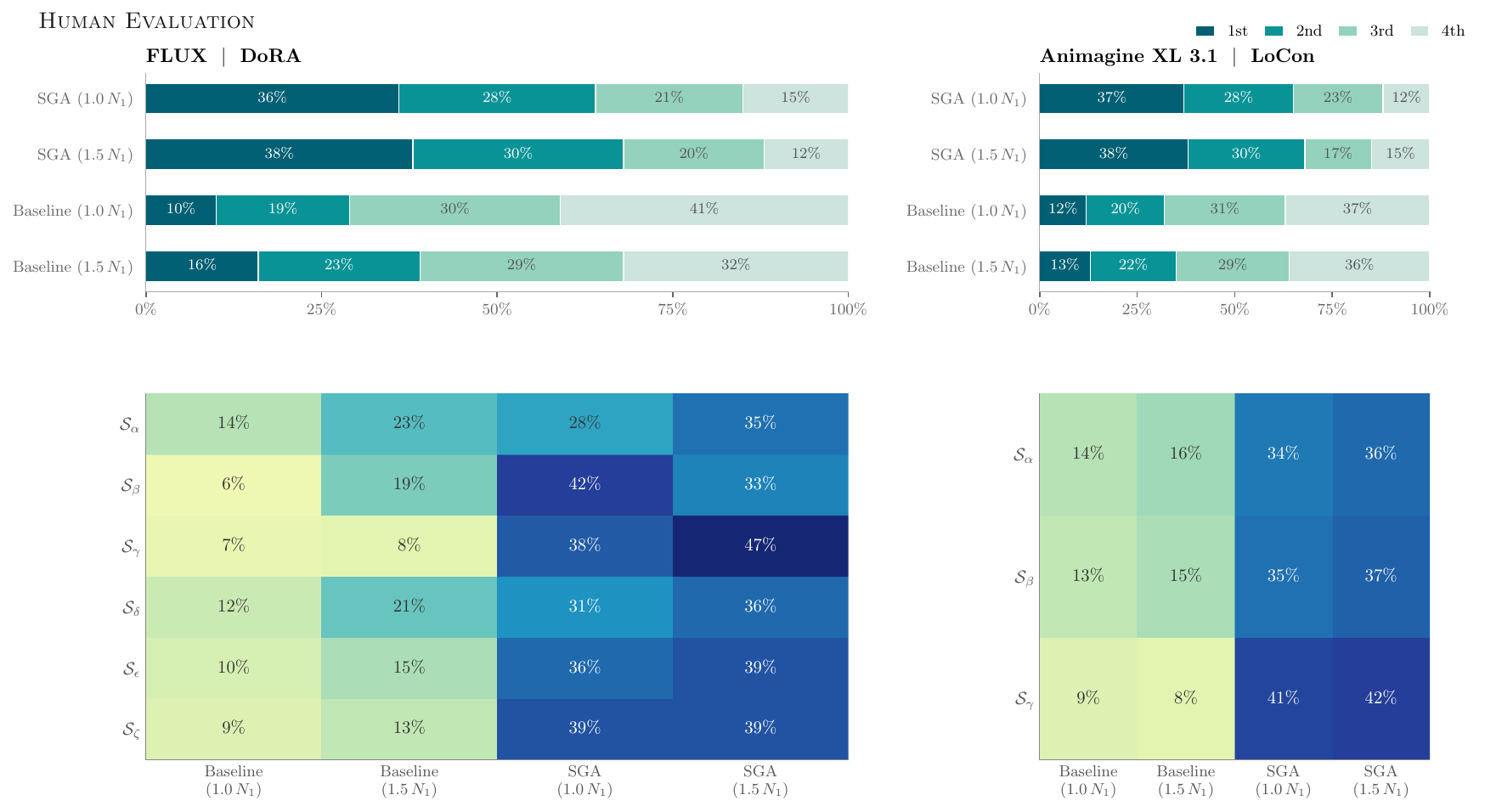}
  \caption{\textbf{Quantitative Evaluation --- Human Evaluation.} Matching layout to \cref{fig:eval-llm}. Human rankings corroborate the automated assessment, with SGA variants accumulating the majority of top rankings.}
  \label{fig:eval-human}
\end{figure}

\noindent As shown in \cref{fig:eval-llm,fig:eval-human}, SGA variants achieve favourable rankings across both evaluator types and architectures. On FLUX, SGA ($1.0\,N_1$) achieves a 1st-place rate of 40\% in the LLM Judge, while both baselines remain below 20\%; human rankings corroborate this trend. The synchronized results across FLUX and Animagine XL 3.1 suggest architecture-independent effectiveness, with per-dataset heatmaps confirming consistency across domains.

\subsubsection{Efficiency.}
Notably, SGA at $1.0\,N_1$ consistently outranks Baseline at $1.5\,N_1$ in both LLM and human evaluations, indicating that SGA achieves superior quality with approximately 33\% less training compute. This suggests that aligning semantic granularity enables more efficient manifold learning.

\subsubsection{Calibration Metrics.}
We additionally report three embedding-based metrics: \textbf{CLIP-I}~\cite{radford2021learning} (image--image similarity, domain fidelity), \textbf{CLIP-T}~\cite{radford2021learning} (text--image similarity, prompt alignment), and \textbf{DINO-I}~\cite{caron2021emerging} (structural correspondence). As shown in \cref{tab:calibration}, SGA improves or maintains all metrics on both frameworks.

\begin{table}[!t]
\centering
\caption{\textbf{Calibration Metrics} (averaged across datasets). Higher is better for all metrics. \textcolor{green!60!black}{$\uparrow$} denotes improvement of SGA over the Baseline.}
\label{tab:calibration}
\setlength{\tabcolsep}{4pt}
\begin{tabular}{@{}l ccc ccc@{}}
\toprule
& \multicolumn{3}{c}{\textbf{FLUX} (6 datasets)} & \multicolumn{3}{c}{\textbf{SDXL} (3 datasets)} \\
\cmidrule(lr){2-4} \cmidrule(lr){5-7}
Metric & Baseline & SGA & & Baseline & SGA & \\
\midrule
CLIP-I & 0.686 & \textbf{0.701} & \textcolor{green!60!black}{$\uparrow$} & 0.715 & \textbf{0.717} & \textcolor{green!60!black}{$\uparrow$} \\
CLIP-T & 0.248 & \textbf{0.253} & \textcolor{green!60!black}{$\uparrow$} & 0.296 & \textbf{0.298} & \textcolor{green!60!black}{$\uparrow$} \\
DINO-I & 0.425 & \textbf{0.427} & \textcolor{green!60!black}{$\uparrow$} & 0.492 & \textbf{0.539} & \textcolor{green!60!black}{$\uparrow$} \\
\bottomrule
\end{tabular}
\end{table}

\subsection{Ablation Study: Stability Analysis}
\label{sec:exp-ablation}

To validate the necessity of each stabilization component, we conducted an ablation study on a representative subset that poses challenging structural constraints. We performed a three-way comparison---\emph{Full SGA}, \emph{Full SGA w/o Tuple-wise Optimization}, and \emph{Full SGA w/o Scale-Adaptive Modulation}---and report the 1st-place rate (\%) under both LLM Judge and Human Evaluation (\cref{tab:ablation}).

\begin{table}[H]
\centering
\caption{\textbf{Ablation Study:} 1st-place rate (\%) across evaluator types. Higher is better.}
\label{tab:ablation}
\setlength{\tabcolsep}{4pt}
\begin{tabular}{@{}l rr rr@{}}
\toprule
& \multicolumn{2}{c}{\textbf{FLUX}} & \multicolumn{2}{c}{\textbf{SDXL}} \\
\cmidrule(lr){2-3} \cmidrule(lr){4-5}
Variant & Human & LLM & Human & LLM \\
\midrule
Full SGA & \textbf{54} & \textbf{55} & \textbf{48} & \textbf{61} \\
w/o Tuple-wise Opt. & 27 & 25 & 16 & 14 \\
w/o Scale-Adaptive Mod. & 19 & 20 & 36 & 25 \\
\bottomrule
\end{tabular}
\end{table}

\noindent Both ablated variants yield substantially lower 1st-place rates, confirming that each component independently contributes to stability. The relative importance differs across architectures: removing Scale-Adaptive Modulation is more detrimental on FLUX, while Tuple-wise Optimization matters more on SDXL---consistent with distinct noise-schedule characteristics. Ablations in Suppl.~G further suggest that Scale-Adaptive Modulation becomes even more critical under FLUX with LoRA, which we attribute to LoRA's coupled magnitude-direction updates amplifying sensitivity to frequency misalignment compared to DoRA's decoupled formulation~\cite{liu2024dora}.

\subsubsection{Architectural Inductive Bias.}
We attribute this asymmetry to the distinct inductive biases of each backbone~\cite{dosovitskiy2021image}. The U-Net's local receptive field provides adequate textural supervision but hinders cross-scale coordination, making Tuple-wise co-sampling critical. Conversely, the DiT's global attention~\cite{peebles2023scalable} lacks an intrinsic frequency prior, rendering Scale-Adaptive Modulation essential.

\section{Conclusion}
\label{sec:conclusion}

We showed that the MSE objective in Flow Matching implicitly optimizes a quadratic interaction field governed by a dynamic NTK, and proposed Semantic Granularity Alignment (SGA) to engineer interventions in this vector residual space. Evaluations across DiT and U-Net architectures confirm that aligning data geometry with optimization structure improves both convergence efficiency and generative quality, underscoring the importance of Data-Training Synergy beyond architectural scaling.

\section{Limitations and Future Work}
\label{sec:limitations}

Two constraints remain: (1)~H-SD depends on upstream detectors, though this is mitigated by high-accuracy YOLO models readily available in practice; (2)~our evaluation is confined to T2I synthesis---extending to Video Generation and Multi-modal Data Mixing remains future work.

\noindent\textbf{Broader Impact.}\;
By lowering the computational barrier for high-fidelity adaptation, SGA may also lower the threshold for misuse; we advocate for provenance tracking and watermarking.


\bibliographystyle{splncs04}
\bibliography{main}


\appendix

\section{Derivation of Granularity-Specific Vector Fields}
\label{appendix:a}

In this section, we provide the rigorous derivation of Theorem 1 presented in the main text. We prove that under the Conditional Flow Matching (CFM) framework, the optimal marginal vector field $u_t(x)$ decomposes into a probability-weighted superposition of sub-fields generated by distinct data granularities.

\subsection{The Marginal Probability as a Gaussian Mixture}

We begin by defining the marginal probability density $p_t(x)$ at time $t$. By definition, the marginal distribution is the expectation of the conditional distribution over the data distribution:
\begin{equation}
p_t(x) = \int p_t(x|x_1) q(x_1) dx_1
\label{eq:marginal}
\end{equation}

In the context of fine-tuning, we operate on a discrete dataset $\mathcal{D} = \{x^{(1)},$ $\dots, x^{(N)}\}$.
The true data distribution $q(x_1)$ is approximated by the empirical distribution~$\hat{q}(x_1)$:
\begin{equation}
q(x_1) \approx \hat{q}(x_1) = \frac{1}{N} \sum_{k=1}^{N} \delta(x_1 - x^{(k)})
\end{equation}

Substituting this empirical distribution into the integral, and utilizing the linearity of integration to interchange the integral and summation operators, we obtain:
\begin{equation}
p_t(x) = \frac{1}{N} \sum_{k=1}^{N} \left[ \int p_t(x|x_1) \cdot \delta(x_1 - x^{(k)}) dx_1 \right]
\end{equation}

Applying the Sifting Property of the Dirac delta function, which states that $\int f(z) \delta(z - a) dz = f(a)$, the integral simplifies to evaluating the conditional density at the specific sample $x^{(k)}$. Crucially, within the CFM framework, the conditional probability path $p_t(x|x_1)$ is \emph{chosen} to be Gaussian by construction---this is a design choice of the flow, not an assumption about the data distribution. Any arbitrary data distribution $q(x_1)$, regardless of its complexity or modality, is compatible with Gaussian conditional paths. Consequently, the marginal distribution $p_t(x)$ takes the form of a Gaussian Mixture Model (GMM):
\begin{equation}
p_t(x) = \frac{1}{N} \sum_{k=1}^{N} \mathcal{N}(x | \mu_t(x^{(k)}), \sigma_t(x^{(k)})^2 I)
\label{eq:gmm}
\end{equation}

\subsection{The Marginal Vector Field as a Nadaraya-Watson Estimator}

We now derive the expression for the marginal vector field $u_t(x)$. The target vector field is defined as the ratio of the aggregate momentum density to the probability density:
\begin{equation}
u_t(x) = \frac{\int u_t(x|x_1) p_t(x|x_1) q(x_1) dx_1}{p_t(x)}
\end{equation}

Applying the same Dirac sifting logic to the numerator, we arrive at the closed-form expression for $u_t(x)$:
\begin{equation}
u_t(x) = \frac{\sum_{i=1}^{N} u_t(x|x^{(i)}) \cdot p_t(x|x^{(i)})}{\sum_{k=1}^{N} p_t(x|x^{(k)})}
\label{eq:nw}
\end{equation}

This form is mathematically equivalent to the Nadaraya-Watson kernel regression estimator. It implies that the optimal vector field at any point $x$ is a weighted average of the conditional vector fields provided by each sample.

\subsection{Decomposition by Data Granularity}

\noindent\textbf{Generality of the partition.}
The decomposition derived in the preceding sections holds for \emph{any} partition of the training set into disjoint groups---the only mathematical requirement is that the index sets $\{\mathcal{I}_\xi\}$ form a valid partition of $\{1,\dots,N\}$.
Semantic segmentation (as adopted in our H-SD pipeline) is one convenient instantiation, but it is by no means the only one.
Any criterion that assigns each sample to exactly one group yields a valid partition, for example:
\emph{(i)}~frequency-band clustering in the latent space,
\emph{(ii)}~resolution or aspect-ratio bucketing,
\emph{(iii)}~caption-level topic grouping,
or \emph{(iv)}~even a random split.
The algebraic identities (\cref{eq:final-decomposition,eq:quadratic-final}) are invariant to the choice of grouping criterion; what changes across different partitions is only the \emph{practical effectiveness} of the resulting sub-fields in reducing cross-scale interference.

Formally, let $\mathcal{T}$ be an arbitrary finite label set (with $|\mathcal{T}|\geq 2$) and let $\{\mathcal{I}_\xi\}_{\xi\in\mathcal{T}}$ be a partition of the dataset indices, \ie, $\mathcal{I}_\xi \cap \mathcal{I}_\eta = \emptyset$ for $\xi\neq\eta$ and $\bigcup_{\xi \in \mathcal{T}} \mathcal{I}_\xi = \{1, \dots, N\}$.
In our experiments we set $\mathcal{T}=\{a,b,c\}$ corresponding to Macro, Meso, and Micro granularities obtained via hierarchical semantic decomposition, but the derivation below is stated for a general $\mathcal{T}$.

We define the Aggregate Local Density $Z_\xi(x)$ and the Group Vector Field $u_t^\xi(x)$ for each group $\xi \in \mathcal{T}$ as:
\begin{equation}
Z_\xi(x) = \sum_{k \in \mathcal{I}_\xi} p_t(x|x^{(k)}), \quad u_t^\xi(x) = \frac{1}{Z_\xi(x)} \sum_{k \in \mathcal{I}_\xi} p_t(x|x^{(k)}) u_t(x|x^{(k)})
\end{equation}

\noindent Note that since the Gaussian conditional density satisfies $p_t(x|x^{(k)}) > 0$ for all $x \in \mathbb{R}^d$ and $t \in (0,1)$, the aggregate density $Z_\xi(x)$ is strictly positive for any non-empty partition, rendering the divisions in the equations above and below universally well-defined without singularities.

We further define the global unnormalized density $Z_{\text{total}}(x) = \sum_{\eta \in \mathcal{T}} Z_\eta(x) = \sum_{k=1}^{N} p_t(x|x^{(k)})$.
Substituting these back into the Nadaraya-Watson estimator, we obtain the final decomposed form:
\begin{equation}
u_t(x) = \sum_{\xi \in \mathcal{T}} \frac{Z_\xi(x)}{Z_{\text{total}}(x)} u_t^\xi(x) = \sum_{\xi \in \mathcal{T}} \alpha_\xi(x) u_t^\xi(x)
\label{eq:final-decomposition}
\end{equation}
where $\alpha_\xi(x) = Z_\xi(x) / Z_{\text{total}}(x)$ represents the local data density (or voting power) of group $\xi$ at position $x$. Note that by definition, $\sum_{\xi} \alpha_\xi(x) = 1$.

\noindent\textit{Remark.}
\cref{eq:final-decomposition} is a purely algebraic identity that follows from the partition structure of $\{\mathcal{I}_\xi\}$ and the linearity of summation; it does \emph{not} depend on how the groups are constructed.
Consequently, the entire downstream analysis---including the Data Interference Matrix (\cref{appendix:b}) and the NTK-based gradient dynamics (\cref{appendix:general-ntk})---applies to any valid partition of the dataset.
Semantic segmentation is the particular instantiation we adopt in practice because it produces groups with well-separated frequency spectra, but the algebraic apparatus itself is general and applies to any valid partition.

\section{Proof of the Data Interference Matrix}
\label{appendix:b}

In this section, we derive Theorem 2, demonstrating that the CFM loss function inherently minimizes a quadratic form governed by a data interference matrix.

\subsection{Loss Decomposition via Partition of Unity}

Locally, at a specific spacetime point $(x, t)$, the squared error between the model output and the marginal target field is defined as:
\begin{equation}
\mathcal{L}_{\text{Kernel}}(x) = \| v_t(x) - u_t(x) \|^2
\end{equation}

\noindent Substituting the decomposed target field $u_t(x) = \sum_{\xi \in \mathcal{T}} \alpha_\xi(x) u_t^\xi(x)$, and crucially, utilizing the Partition of Unity property $\sum_{\xi \in \mathcal{T}} \alpha_\xi(x) = 1$ to rewrite the model output as $v_t(x) = \sum_{\xi \in \mathcal{T}} \alpha_\xi(x) v_t(x)$, we can regroup the error terms:
\begin{equation}
\mathcal{L}_{\text{Kernel}}(x) = \left\| \sum_{\xi \in \mathcal{T}} \alpha_\xi(x) \underbrace{(v_t(x) - u_t^\xi(x))}_{\Delta_\xi(x)} \right\|^2
\end{equation}

Let $\Delta_\xi(x) = v_t(x) - u_t^\xi(x)$ denote the error vector relative to data source $\xi$.

\subsection{Expansion into Quadratic Form}

Expanding the squared norm of the weighted sum $\sum_{\xi} \alpha_\xi \Delta_\xi$, we apply the identity $\|\sum_i v_i\|^2 = \sum_i \sum_j \langle v_i, v_j \rangle$:
\begin{equation}
\mathcal{L}_{\text{Kernel}}(x) = \sum_{\xi \in \mathcal{T}} \sum_{\eta \in \mathcal{T}} \alpha_\xi(x) \alpha_\eta(x) \langle \Delta_\xi(x), \Delta_\eta(x) \rangle
\end{equation}

This expansion reveals two distinct interactions:
\begin{itemize}
\item \textbf{Self-Alignment Terms} (Diagonal, $\xi=\eta$): $\alpha_\xi^2 \|\Delta_\xi\|^2$, representing the independent fitting error for each data source.
\item \textbf{Cross-Interference Terms} (Off-Diagonal, $\xi \neq \eta$): $\alpha_\xi \alpha_\eta \langle \Delta_\xi, \Delta_\eta \rangle$, representing the residual correlation between data sources.
\end{itemize}

\subsection{The Data Interference Matrix $\mathbf{\Omega}$}

To compactly represent this structure, we introduce the following definitions. Let $\mathcal{T}$ be the label set of data granularities with $|\mathcal{T}| = N_g$.

\noindent\textbf{Step 1: Entry-wise definition.}\;
We define the symmetric positive semi-definite (PSD) \emph{Data Interference Matrix} $\mathbf{\Omega}(x) \in \mathbb{R}^{N_g \times N_g}$ with entries:
\begin{equation}
\Omega_{\xi \eta}(x) = \langle \Delta_\xi(x), \Delta_\eta(x) \rangle, \quad \xi, \eta \in \mathcal{T}
\end{equation}

\noindent Mathematically, $\mathbf{\Omega}(x)$ is the Gram matrix of the residual vectors $\{\Delta_\xi(x)\}_{\xi \in \mathcal{T}}$, which guarantees positive semi-definiteness by construction. This ensures $\boldsymbol{\alpha}^\top \mathbf{\Omega} \, \boldsymbol{\alpha} \geq 0$ for all $\boldsymbol{\alpha}$, consistent with the physical interpretation that the squared Euclidean distance is non-negative.

\noindent\textbf{Step 2: Weighting vector.}\;
We define the density weighting vector $\boldsymbol{\alpha}(x) \in \mathbb{R}^{N_g}$ whose $\xi$-th component is $\alpha_\xi(x)$, \ie,
\begin{equation}
\boldsymbol{\alpha}(x) = \big[\alpha_{\xi_1}(x),\, \alpha_{\xi_2}(x),\, \dots,\, \alpha_{\xi_{N_g}}(x)\big]^\top, \quad \text{with } \sum_{\xi \in \mathcal{T}} \alpha_\xi(x) = 1
\end{equation}

\noindent\textbf{Step 3: Quadratic form.}\;
Substituting these definitions, the local loss function is strictly reformulated as a quadratic form:
\begin{equation}
\mathcal{L}_{\text{Kernel}}(x) = \boldsymbol{\alpha}(x)^\top \mathbf{\Omega}(x) \, \boldsymbol{\alpha}(x)
\label{eq:quadratic-final}
\end{equation}

\noindent This formulation holds for any finite partition $\mathcal{T}$ of arbitrary cardinality $N_g \geq 2$. The optimization landscape is mathematically governed by $\mathbf{\Omega}(x)$: diagonal entries $\Omega_{\xi\xi}$ capture the self-alignment error within each granularity, while off-diagonal entries $\Omega_{\xi\eta}$ ($\xi \neq \eta$) quantify the constructive or destructive interference between data sources. In our experiments, we instantiate $\mathcal{T} = \{\text{Macro}, \text{Meso}, \text{Micro}\}$ with $N_g = 3$.

\section{Derivation of the General Flow Matching Objective}
\label{appendix:c}

In this section, we connect the data-source interference analysis from \cref{appendix:b} back to the general Flow Matching (FM) framework. We derive the tractable training objective by transitioning from the marginal vector field to the conditional vector field, utilizing a Gaussian reparameterization path.

\subsection{Equivalence to the Global FM Objective}

Recall from Theorem 2 (\cref{appendix:b}) that the local loss at any spacetime point $(x, t)$ is governed by the quadratic form of the data density weights $\boldsymbol{\alpha}(x)$ and the interference matrix $\mathbf{\Omega}(x)$:
\begin{equation}
\mathcal{L}_{\text{local}}(x) = \boldsymbol{\alpha}(x)^\top \mathbf{\Omega}(x) \boldsymbol{\alpha}(x)
\end{equation}

By reversing the decomposition in \cref{appendix:b}, this quadratic form is strictly mathematically equivalent to the squared Euclidean distance between the model output $v_\theta(x, t)$ and the true marginal vector field $u_t(x)$:
\begin{equation}
\boldsymbol{\alpha}(x)^\top \mathbf{\Omega}(x) \boldsymbol{\alpha}(x) \equiv \| v_\theta(x, t) - u_t(x) \|^2
\end{equation}

Consequently, the global Flow Matching objective is defined as the expectation of this local error over time $t$ and the marginal data distribution $p_t(x)$:
\begin{equation}
\mathcal{L}_{FM}(\theta) = \mathbb{E}_{t \sim \pi(t)} \left[ \mathbb{E}_{x \sim p_t(x)} \left[ \| v_\theta(x, t) - u_t(x) \|^2 \right] \right]
\label{eq:fm-global}
\end{equation}

\noindent\textit{Remark:} The marginal vector field $u_t(x)$ in \cref{eq:fm-global} is computationally intractable, as it requires integrating over the entire data distribution $q(x_1)$ to evaluate the marginal density $p_t(x)$ and its derivative.

\subsection{Tractability via Conditional Flow Matching}

To eliminate the intractable $u_t(x)$, we invoke the core theorem of Conditional Flow Matching~\cite{lipman2023flow,tong2024improving}, which states that fitting the marginal vector field is equivalent to fitting the conditional vector fields $u_t(x|x_1)$ under the expectation over data.

By swapping the order of integration (demarginalization), we obtain the Conditional Flow Matching (CFM) objective:
\begin{equation}
\mathcal{L}_{CFM}(\theta) = \mathbb{E}_{t \sim \pi(t)} \left[ \mathbb{E}_{x_1 \sim q(x_1)} \mathbb{E}_{x \sim p_t(x|x_1)} \left[ \| v_\theta(x, t) - u_t(x|x_1) \|^2 \right] \right]
\label{eq:cfm-appendix}
\end{equation}

Here, the target field shifts from the aggregate marginal field $u_t(x)$ to the tractable single-sample conditional field $u_t(x|x_1)$.

\subsection{Reparameterization for Gaussian Paths}

To evaluate the inner expectation efficiently, we assume the conditional probability path $p_t(x|x_1)$ follows a general Gaussian distribution:
\begin{equation}
p_t(x|x_1) = \mathcal{N}(x \,|\, \mu_t(x_1), \sigma_t(x_1)^2 \mathbf{I})
\end{equation}

We apply the reparameterization trick using standard normal noise $\epsilon \sim \mathcal{N}(0, \mathbf{I})$:

\noindent\textbf{Sample Location} ($x$):
\begin{equation}
x = \Psi_t(x_1) = \mu_t(x_1) + \sigma_t(x_1) \cdot \epsilon
\label{eq:reparam-sample}
\end{equation}

\noindent\textbf{Compute Target Flow} ($u_t$):
The target vector field is defined as the time derivative of the flow map $\Psi_t(x_1)$:
\begin{equation}
u_t(x|x_1) = \frac{d}{dt} \Psi_t(x_1) = \dot{\mu}_t(x_1) + \dot{\sigma}_t(x_1) \cdot \epsilon
\label{eq:reparam-target}
\end{equation}
where $\dot{\mu}_t$ and $\dot{\sigma}_t$ denote the partial derivatives with respect to time $t$.

\subsection{The Empirical Batch Estimator}

Finally, we discretize the expectations into a batch-computable loss function. Let $\mathcal{B} = \{x_1^{(1)}, \dots, x_1^{(B)}\}$ be a batch of data samples, $t \sim \pi(t)$ be sampled times, and $\epsilon \sim \mathcal{N}(0, \mathbf{I})$ be sampled noise.

The general CFM loss function is given by:
\begin{equation}
\mathcal{L}_{\text{General}}(\theta) \!=\! \frac{1}{B} \!\sum_{i=1}^{B}
\left\| v_\theta \!\Big( \underbrace{\mu_{t_i}\!(x_1^{(i)}) \!+\! \sigma_{t_i}\!(x_1^{(i)}) \!\cdot\! \epsilon_i}_{\text{Input } x_{t_i}},\, t_i \Big)
\!-\! \underbrace{\Big( \dot{\mu}_{t_i}\!(x_1^{(i)}) \!+\! \dot{\sigma}_{t_i}\!(x_1^{(i)}) \!\cdot\! \epsilon_i \Big)}_{\text{Target } u_{t_i}} \right\|^2
\label{eq:general-cfm}
\end{equation}

\section{Implementation Details of Scale-Adaptive Modulation}
\label{appendix:d}

In this section, we provide the detailed mathematical formulations and hyperparameter settings for the Scale-Adaptive Modulation strategies introduced in the main text. We outline the specific mechanisms applied to DiT-based and U-Net-based architectures.

\subsection{Conditional Logit-Normal Sampling (For DiT Architectures)}

For MM-DiT backbones (specifically FLUX.1), the training process relies on a Logit-Normal distribution for time-step sampling~\cite{esser2024scaling}. To align the optimization dynamics with the frequency characteristics of our data slices, we introduce a Granularity-Dependent Shift to the sampling process.

\subsubsection{The Sampling Chain.}
Let $z \sim \mathcal{N}(0, 1)$ be a standard normal variable. The standard Logit-Normal sampling transforms $z$ into a time step $t \in (0, 1)$ via a sigmoid function. We modify the mean of this distribution based on the data granularity $S$.

We define the Granularity Bias Function $\Delta(S)$ as:
\begin{equation}
\Delta(S) =
\begin{cases}
+0.5, & \text{if } S = \text{Macro} \\
\phantom{+}0.0, & \text{if } S = \text{Meso} \\
-0.5, & \text{if } S = \text{Micro}
\end{cases}
\label{eq:granularity-bias}
\end{equation}

The intermediate time variable $t_{\text{logit}}$ is computed as:
\begin{equation}
t_{\text{logit}} = \sigma(s \cdot z + \Delta(S)) = \frac{1}{1 + \exp(-(s \cdot z + \Delta(S)))}
\label{eq:logit-time}
\end{equation}
where $s$ is the scale factor (set to $1.0$ in our experiments).

\subsubsection{Resolution-Dependent Flow Shift.}
Following standard practice in FLUX, we further adjust the time step to account for image resolution, obtaining the final training time step~$t$ via a resolution-aware mapping:
\begin{equation}
t = \frac{t_{\text{logit}}}{t_{\text{logit}} + (1 - t_{\text{logit}}) \cdot \exp(-\mu_{\text{res}})}
\label{eq:resolution-shift}
\end{equation}
where $\mu_{\text{res}}$ is a linear function of the total pixel count $H \times W$.

\subsubsection{Effect on Optimization.}
By setting $\Delta(\text{Macro}) > 0$, the probability mass of $t$ shifts towards $1$ (high noise). This biases the model to prioritize low-frequency geometric reconstruction during the early phases of generation. Conversely, $\Delta(\text{Micro}) < 0$ shifts the mass towards $0$ (low noise), encouraging high-frequency details to be refined under sustained gradient supervision in the late denoising stages.

\subsection{Granularity-Adaptive Min-SNR Weighting (For U-Net Architectures)}

For U-Net architectures (\eg, SDXL), we adopt the Min-SNR weighting strategy~\cite{hang2023efficient} to mitigate task conflict across noise levels. We extend this by introducing a Granularity-Adaptive Gamma Threshold.

\subsubsection{Signal-to-Noise Ratio (SNR).}
We define the SNR at time step $t$ as $\text{SNR}(t) = \frac{\bar{\alpha}_t}{1 - \bar{\alpha}_t}$, where $\bar{\alpha}_t$ follows the standard noise schedule~\cite{karras2022elucidating}.

\subsubsection{Granularity-Specific Clamping.}
The standard Min-SNR strategy clamps the loss weight at a fixed threshold $\gamma$ (typically $\gamma=5.0$). We replace this global constant with a granularity-specific function $\Gamma(S)$, formulated to balance structural and textural learning:
\begin{equation}
\Gamma(S) =
\begin{cases}
7.0, & \text{if } S = \text{Micro} \\
5.0, & \text{if } S = \text{Meso} \\
4.0, & \text{if } S = \text{Macro}
\end{cases}
\label{eq:gamma-mapping}
\end{equation}

\subsubsection{The Reweighting Function.}
For the standard $\epsilon$-prediction objective, the final weighting term $\omega(t, S)$ applied to the loss is defined as:
\begin{equation}
\omega(t, S) = \frac{\min(\text{SNR}(t),\, \Gamma(S))}{\text{SNR}(t)} = \min\!\left(1,\, \frac{\Gamma(S)}{\text{SNR}(t)}\right)
\label{eq:reweighting}
\end{equation}

\subsubsection{Effect on Optimization.}
\textbf{Micro-Boost} ($\Gamma=7.0$): By increasing the clamping threshold for Micro slices, the weighting function maintains a unit weight ($1.0$) deeper into the high-SNR regime. This mitigates gradient decay during the late-stage phase of fine-detail reconstruction.

\noindent\textbf{Macro-Relax} ($\Gamma=4.0$): By lowering the threshold for Macro slices, we accelerate weight decay as SNR increases. This regularizes the model, reducing its tendency to overfit to high-frequency noise when learning global structural concepts.

\section{General Framework: Bridging Output-Space Residuals and Parameter-Space Gradient Dynamics}
\label{appendix:general-ntk}

The preceding sections establish that the optimization loss decomposes into a quadratic form governed by the Data Interference Matrix $\mathbf{\Omega}$, whose entries $\Omega_{ij} = \Delta_i^\top \Delta_j$ capture the geometric alignment of residual vectors in the \emph{output space}. A natural and deeper question arises: does this output-space geometry faithfully reflect the actual optimization dynamics in the \emph{parameter space}? In this section, we derive a general, architecture-agnostic framework that formally connects these two spaces through the Neural Tangent Kernel (NTK)~\cite{jacot2018neural}, thereby endowing the data-geometric analysis with concrete dynamical significance.

\subsection{Recall of Core Definitions}

We begin by recalling the core quantities established in the preceding sections, which serve as the starting point for the gradient-space analysis.

From \cref{appendix:a}, the true marginal vector field at any spacetime point $(x, t)$ decomposes as a density-weighted superposition of granularity-specific sub-fields:
\begin{equation}
u_t(x) = \sum_{\xi \in \mathcal{T}} \alpha_\xi(x) \, u_t^\xi(x), \quad \text{with} \quad \sum_{\xi \in \mathcal{T}} \alpha_\xi(x) = 1
\label{eq:general-target-decomp}
\end{equation}
where $\alpha_\xi(x) = Z_\xi(x) / Z_{\text{total}}(x)$ is the \emph{local data density} of source $\xi$---the probability that position $x$ is governed by sub-manifold $\xi$ at time $t$.

From \cref{appendix:b}, the Data Interference Matrix $\mathbf{\Omega}$ is constructed from the per-granularity \emph{vector residuals}:
\begin{equation}
\Delta_\xi(x, t) = v_\theta(x, t) - u_t^\xi(x), \quad \xi \in \mathcal{T}
\label{eq:general-residual}
\end{equation}
with entries $\Omega_{\xi\eta} = \langle \Delta_\xi, \Delta_\eta \rangle = \Delta_\xi^\top \Delta_\eta$. The local Flow Matching loss was shown to be the quadratic form (\cref{eq:quadratic-final}):
\begin{equation}
\boldsymbol{\alpha}(x)^\top \mathbf{\Omega}(x) \, \boldsymbol{\alpha}(x) = \left\| \sum_{\xi \in \mathcal{T}} \alpha_\xi(x) \, \Delta_\xi(x, t) \right\|^2
\label{eq:general-loss-exact}
\end{equation}

\noindent Since for any scalar $c>0$, $\arg\min_\theta f(\theta) = \arg\min_\theta c\,f(\theta)$, we adopt the rescaled form to streamline the subsequent gradient analysis. Hereafter, we redefine:
\begin{equation}
\mathcal{L}_{\text{Kernel}}(x) \triangleq \frac{1}{2}\left\| \sum_{\xi \in \mathcal{T}} \alpha_\xi(x) \, \Delta_\xi(x, t) \right\|^2
\label{eq:general-loss}
\end{equation}

\noindent This rescaling preserves the minimizer and all qualitative properties of the optimization landscape; the factor $\frac{1}{2}$ serves solely to eliminate the constant~$2$ from the chain rule in the gradient derivations that follow.

These quantities---the density weights $\alpha_\xi$, the vector residuals $\Delta_\xi$ composing $\mathbf{\Omega}$, and the quadratic loss structure---are the elements from which we now derive the gradient dynamics in parameter space.

\subsection{Decomposition of the Global Residual via Partition of Unity}

The Partition of Unity property $\sum_{\xi \in \mathcal{T}} \alpha_\xi(x) = 1$ permits an identity expansion of the model output: $v_\theta(x,t) = \left(\sum_{\xi \in \mathcal{T}} \alpha_\xi(x)\right) v_\theta(x,t)$. Combining this with the target decomposition (\cref{eq:general-target-decomp}), the global residual vector $e(x,t) = v_\theta(x,t) - u_t(x)$ decomposes exactly as:
\begin{equation}
e(x,t) = \sum_{\xi \in \mathcal{T}} \alpha_\xi(x) \big( v_\theta(x,t) - u_t^\xi(x) \big) = \sum_{\xi \in \mathcal{T}} \alpha_\xi(x) \, \Delta_\xi(x,t)
\label{eq:global-residual-decomp}
\end{equation}

This identity is the algebraic foundation for connecting the output-space geometry to parameter-space dynamics: the global residual is a density-weighted superposition of per-granularity residuals.

\subsection{From the Global Residual to Parameter Gradients}

Let $J_\theta(x,t) \in \mathbb{R}^{d_{\text{out}} \times P}$ denote the Jacobian matrix of the network output with respect to its parameters:
\begin{equation}
J_\theta(x,t) = \frac{\partial v_\theta(x,t)}{\partial \theta}
\end{equation}

Applying the chain rule to the global loss (\cref{eq:general-loss}), the total parameter gradient is:
\begin{equation}
g_{\text{total}} = \nabla_\theta \mathcal{L}_{\text{Kernel}}(x) = J_\theta(x,t)^\top e(x,t)
\label{eq:general-total-grad-chain}
\end{equation}

Substituting the residual decomposition (\cref{eq:global-residual-decomp}) and invoking linearity of the matrix-vector product:
\begin{equation}
g_{\text{total}} = J_\theta(x,t)^\top \left( \sum_{\xi \in \mathcal{T}} \alpha_\xi(x) \, \Delta_\xi(x,t) \right) = \sum_{\xi \in \mathcal{T}} \alpha_\xi(x) \, J_\theta(x,t)^\top \Delta_\xi(x,t)
\label{eq:general-grad-total}
\end{equation}

This reveals the key structural insight: each output-space residual $\Delta_\xi \in \mathbb{R}^{d_{\text{out}}}$ is \emph{lifted} into the parameter space $\mathbb{R}^P$ via the transposed Jacobian $J_\theta^\top$, which acts as a linear map from the output tangent space to the parameter tangent space. We therefore \emph{define} the gradient component attributable to granularity $\xi$ as:
\begin{equation}
g_\xi \triangleq \alpha_\xi(x) \, J_\theta(x,t)^\top \Delta_\xi(x,t)
\label{eq:general-grad-component}
\end{equation}

\noindent\textit{Remark.} Crucially, $g_\xi$ is \emph{not} defined as the gradient of an isolated per-granularity loss $\frac{1}{2}\alpha_\xi \|\Delta_\xi\|^2$. It is the mathematically exact component of the total gradient $g_{\text{total}} = \sum_\xi g_\xi$ that arises from the density-weighted decomposition of the global residual, preserving the physical meaning of $\alpha_\xi$ as the local field density throughout.

\subsection{Gradient Dynamics and the Emergence of the NTK}

The optimization dynamics are governed by the squared norm of the total gradient, which decomposes via the bilinear structure of the inner product:
\begin{equation}
\| g_{\text{total}} \|^2 = \left\| \sum_{\xi \in \mathcal{T}} g_\xi \right\|^2 = \sum_{\xi \in \mathcal{T}} \sum_{\eta \in \mathcal{T}} \langle g_\xi, g_\eta \rangle
\label{eq:grad-norm-decomp}
\end{equation}

Expanding any pairwise gradient inner product using \cref{eq:general-grad-component}:
\begin{align}
\langle g_\xi, g_\eta \rangle &= \left(\alpha_\xi J_\theta^\top \Delta_\xi\right)^\top \left(\alpha_\eta J_\theta^\top \Delta_\eta\right) \notag \\
&= \alpha_\xi \alpha_\eta \, \Delta_\xi^\top \underbrace{\left(J_\theta \, J_\theta^\top\right)}_{\Theta_\theta(x,t)} \Delta_\eta
\label{eq:general-ntk-expansion}
\end{align}

The matrix $\Theta_\theta(x,t) = J_\theta(x,t) J_\theta(x,t)^\top \in \mathbb{R}^{d_{\text{out}} \times d_{\text{out}}}$ is precisely the local Neural Tangent Kernel (NTK)~\cite{jacot2018neural,chizat2019lazy}. This yields the central result of this section:
\begin{equation}
\boxed{\;\langle g_\xi, g_\eta \rangle = \alpha_\xi \alpha_\eta \, \Delta_\xi^\top \, \Theta_\theta(x,t) \, \Delta_\eta\;}
\label{eq:general-ntk-result}
\end{equation}

\subsection{On the Intractability of the NTK and the Case for Output-Space Intervention}

\cref{eq:general-ntk-result} reveals that the true gradient interference is governed by $\Delta_\xi^\top \Theta_\theta \Delta_\eta$, not the bare residual inner product $\Delta_\xi^\top \Delta_\eta$. In principle, the NTK matrix $\Theta_\theta$ can amplify, attenuate, or rotate the interference pattern relative to $\mathbf{\Omega}$. We do \emph{not} assume that $\Theta_\theta$ is isotropic---for modern generative architectures with billions of parameters, heterogeneous layer types, and attention mechanisms, such an assumption would be unrealistic.

However, direct intervention on $\Theta_\theta$ is equally intractable: the NTK is an implicit function of the entire network architecture, the current parameter state, and the input $(x, t)$. It cannot be analytically computed, stably estimated, or meaningfully controlled during training for models at the scale of FLUX or SDXL.

We therefore adopt a pragmatic strategy grounded in the bilinear structure of \cref{eq:general-ntk-result}: since $\langle g_\xi, g_\eta \rangle = \alpha_\xi \alpha_\eta \, \Delta_\xi^\top \Theta_\theta \, \Delta_\eta$, the gradient inner product is \emph{jointly} determined by the residual vectors $\Delta_\xi, \Delta_\eta$ and the kernel $\Theta_\theta$. While we cannot control $\Theta_\theta$, we \emph{can} control the statistical structure of the residuals through data curation and sampling design:

\begin{enumerate}
\item \textbf{Hierarchical Dataset Construction}: By organizing training data into structurally coherent granularities (Macro, Meso, Micro), we structure the residual vectors $\Delta_\xi$ such that they encode well-separated frequency content, aiming to reduce the magnitude of destructive cross-scale inner products $\Delta_\xi^\top \Theta_\theta \Delta_\eta$ without requiring specific knowledge of the spectral structure of $\Theta_\theta$.

\item \textbf{Tuple-wise Co-sampling}: Mathematically, since the NTK is defined as $\Theta_{\theta}=J_{\theta}J_{\theta}^{\top}$, it is inherently a positive semi-definite (PSD) matrix. However, due to its high dimensionality and structural complexity, a PSD matrix does not possess sign preservation for inner products. That is, even if the output-space residuals are aligned ($\Delta_{\xi}^{\top}\Delta_{\eta}>0$), the actual parameter-space gradient interference $\Delta_{\xi}^{\top}\Theta_{\theta}\Delta_{\eta}$ may still be negative (destructive interference) due to the rotational effects of the NTK. Drawing from widespread empirical observations in Multi-Task Learning (MTL)~\cite{sener2018multi,yu2020gradient,liu2021conflict}, if gradient updates are consecutively dominated by a single task (or granularity), the optimization trajectory suffers from severe oscillation and drift. To mitigate this, we intervene directly in the output residual space: by co-sampling heterogeneous granularities within the same optimization step, we promote synchronous updates. This paradigm encourages the optimizer to navigate the superposition of gradients, aiming to reduce both the amplitude and frequency of oscillations.

\item \textbf{Scale-Adaptive Modulation}: Mathematically, the convergence rate of gradient descent is governed by the condition number of the interaction matrix~\cite{arora2019fine,du2019gradient}, where diagonal dominance ensures stable and accelerated optimization. Furthermore, the NTK exhibits a well-documented Spectral Bias~\cite{rahaman2019spectral,basri2020frequency}, inherently learning low-frequency components faster than high-frequency ones. By conditioning the time-step distribution on granularity, we explicitly align the data frequency with the corresponding eigenspectrum of the NTK. Operating each sub-manifold in a well-suited noise regime is expected to amplify its self-alignment signal ($\Delta_{\xi}^{\top}\Theta_{\theta}\Delta_{\xi}$), encouraging the local gradient interaction matrix toward greater diagonal dominance. This is intended to act as a heuristic preconditioning mechanism, promoting the constructive component of the NTK-projected gradient while mitigating spectral conflation.
\end{enumerate}

In summary, the SGA framework does not require the NTK to be well-conditioned or isotropic. Instead, it promotes optimization stability by shaping the \emph{input} to the bilinear form---the residual vectors $\Delta_\xi$---through principled data and sampling design, so that the effective gradient $g_{\text{total}}$ is more likely to remain within a cone of stable descent even under an unknown and complex $\Theta_\theta$.

\subsection{Broader Implications: A Speculative Connection to Scaling Laws}

The following discussion is speculative but suggestive. While the primary focus of this work is to resolve cross-scale gradient conflicts during generative fine-tuning, the quadratic optimization geometry and NTK framework derived in \cref{appendix:general-ntk} provide a lens through which to interpret the empirical successes of Scaling Laws~\cite{mccandlish2018empirical}. We hypothesize that the macroscopic benefits of scaling data ($N$) and batch size ($B$) may be partly tied to the stabilization of the Data Interference Matrix $\mathbf{\Omega}$ and the well-conditioning of the local vector field.

\subsubsection{Large Batch Sizes Enable Accurate Estimation of Population Interference.}
As established in \cref{eq:general-ntk-result}, the true gradient interference between any two data samples is governed by their NTK-projected residuals: $\langle g_{\xi}, g_{\eta} \rangle = \alpha_{\xi}\alpha_{\eta}\Delta_{\xi}^{\top}\Theta_{\theta}(x,t)\Delta_{\eta}$. In standard training with a small batch size $B$, the empirical estimation of these gradient interactions suffers from high statistical variance. The optimization trajectory can be significantly influenced by stochastic, local sampling noise rather than the true geometric structure of the data manifold. As the dataset scales, permitting a proportionally larger $B$, the empirical batch-level gradient interactions reliably approximate the true population expected interference. This allows the optimizer to descend along a trajectory that more closely reflects the global NTK geometry.

\subsubsection{Heterogeneous Co-occurrence Explicitly Resolves Cross-Term Conflicts.}
Beyond variance reduction, scaling $B$ fundamentally alters the interaction dynamics within the Data Interference Matrix $\mathbf{\Omega}$. With a small $B$, a single optimization step is highly likely to sample homogeneous data. Consequently, the off-diagonal cross-interference terms in $\mathbf{\Omega}$ are frequently absent from the current update step, leading to isolated feature fitting and subsequent gradient conflicts across iterations. Conversely, a massive $B$ greatly increases the likelihood of co-occurrence of a broadly heterogeneous set of samples. This macroscopic diversity forces the optimizer to simultaneously navigate the complete, fully-populated geometry of $\mathbf{\Omega}$. By explicitly confronting and resolving these heterogeneous cross-term conflicts at every step, the model is encouraged to converge toward a geometric equilibrium that tends to generalize better.

\subsubsection{Massive Datasets Well-Condition the Target Vector Field Density.}
From a purely data-centric perspective, scaling the total sample size $N$ directly improves the mathematical properties of the learning objective. As shown in \cref{eq:nw}, the optimal marginal vector field $u_t(x)$ functions as a Nadaraya-Watson kernel regression estimator. When $N$ is small, the empirical distribution is sparse. Therefore, at any inference point $x$, the local density weights $\boldsymbol{\alpha}(x)$ are overly influenced by a few outlier data points, resulting in a rugged, highly non-linear target vector field. As $N$ scales exponentially, the empirical manifold densely and uniformly populates the true data manifold. This macroscopic scaling tends to ``well-condition'' the microscopic local density $\boldsymbol{\alpha}(x)$, promoting a smoother probability assignment. From the perspective of our quadratic form $\mathcal{L}_{\text{local}}(x) = \boldsymbol{\alpha}^{\top}\mathbf{\Omega}\boldsymbol{\alpha}$, a smooth and well-conditioned density field tends to reduce the Lipschitz constant of the target vector field, making the residual field $\Delta(x,t)$ inherently easier to fit.

\subsubsection{Summary: SGA as a Structured Analogy to Scaling Laws.}
In conclusion, under compute- and data-constrained fine-tuning scenarios, the proposed Semantic Granularity Alignment (SGA) framework can be viewed as a structured analogy to certain optimization benefits typically associated with Scaling Laws.

First, the H-SD mechanism explicitly expands the number of valid training samples $N$ utilized for generative modeling. This physical scale expansion is expected to help smooth the estimation of the local target vector field density. Simultaneously, by hierarchically decomposing the data manifold, H-SD disentangles complex, interwoven signals, allowing crucial semantic features to be presented independently across distinct slices, thereby lowering the fitting difficulty of the gradient update field.

Furthermore, the other two core components of SGA---Tuple-wise Optimization and Scale-Adaptive Modulation---are designed to execute targeted geometric interventions under the objective constraint of small-batch updates. This structured co-sampling and frequency band modulation help suppress the gradient oscillations induced by heterogeneous cross-terms and reduce the statistical variance of single-step updates. In doing so, SGA strives to closely approximate the accurate, global population interference estimation that is otherwise typically achieved only with a massive batch size $B$. Ultimately, SGA represents an attempt to partially compensate for macroscopic scale disadvantages through microscopic prior design, and aims to achieve this by aligning the intrinsic geometry of the data with the underlying optimization dynamics.

\section{Full Experimental Setup}
\label{appendix:f}

\subsection{FLUX Training Configuration}

All FLUX experiments share a unified training configuration, summarized in \cref{tab:flux-config}.

\begin{table}[H]
\centering
\caption{\textbf{FLUX training hyperparameters.} All domains use the same configuration.}
\label{tab:flux-config}
\begin{tabular}{@{}ll@{}}
\toprule
\textbf{Hyperparameter} & \textbf{Value} \\
\midrule
Backbone & FLUX.1-dev~\cite{flux2024} \\
Adaptation Method & DoRA~\cite{liu2024dora} \\
LoRA Rank & 32 \\
LoRA Alpha & 16 \\
Batch Size & 8 \\
Learning Rate & $1 \times 10^{-4}$ \\
Training Resolution & $1280 \times 1280$ \\
Aspect-Ratio Bucketing (ARB) & Enabled (all domains) \\
Optimizer & AdamW~\cite{loshchilov2019decoupled} \\
Hardware & NVIDIA RTX PRO 6000 Blackwell (96\,GB GDDR7) \\
\bottomrule
\end{tabular}
\end{table}

\subsection{Per-Domain Training Budget}

As stated in the main text, training budgets are reported as multiples of a reference GPU-time: $N_1$ denotes the shorter checkpoint and $N_2 = 1.5\,N_1$ the longer one. \cref{tab:gpu-time} lists the estimated wall-clock GPU hours for each domain on a single RTX PRO 6000.

\begin{table}[H]
\centering
\caption{\textbf{Estimated GPU hours per domain (FLUX).} $N_1$ and $N_2$ denote the two evaluation checkpoints. Actual models used for evaluation are the closest saved checkpoints to the listed times.}
\label{tab:gpu-time}
\begin{tabular}{@{}lcc@{}}
\toprule
\textbf{Domain} & $\boldsymbol{N_1}$ \textbf{(hours)} & $\boldsymbol{N_2}$ \textbf{(hours)} \\
\midrule
$\alpha$ & 14 & 21 \\
$\beta$  & 12 & 18 \\
$\gamma$ & 14 & 21 \\
$\delta$ & 9  & 13.5 \\
$\varepsilon$ & 8  & 12 \\
$\zeta$  & 14 & 21 \\
\bottomrule
\end{tabular}
\end{table}

\noindent\textit{Note:} The GPU hours listed above are estimated values. In practice, we select the saved checkpoint closest to each target time point for evaluation, rather than training to the exact duration. The deviation between the estimated time and the actual checkpoint used does not exceed 30 minutes in any case.

\subsection{Animagine XL 3.1 (SDXL U-Net) Training Configuration}

For the U-Net architecture experiments, we use Animagine XL~3.1~\cite{podell2024sdxl} as the base model. The full configuration is listed in \cref{tab:sdxl-config}.

\begin{table}[H]
\centering
\caption{\textbf{Animagine XL 3.1 (SDXL) training hyperparameters.}}
\label{tab:sdxl-config}
\begin{tabular}{@{}ll@{}}
\toprule
\textbf{Hyperparameter} & \textbf{Value} \\
\midrule
Backbone & Animagine XL 3.1~\cite{podell2024sdxl} \\
Optimizer & Lion \\
U-Net Learning Rate & $3 \times 10^{-5}$ \\
Text Encoder Learning Rate & $3 \times 10^{-6}$ \\
Batch Size & 2 \\
\midrule
\multicolumn{2}{@{}l}{\textit{Linear layers}} \\
\quad LoRA Rank & 64 \\
\quad LoRA Alpha & 32 \\
\midrule
\multicolumn{2}{@{}l}{\textit{Convolutional layers}} \\
\quad Conv Rank & 16 \\
\quad Conv Alpha & 8 \\
\midrule
Caption Shuffle & Enabled \\
Offset Noise & Enabled (default strength) \\
Hardware & NVIDIA RTX PRO 6000 Blackwell (96\,GB GDDR7) \\
\bottomrule
\end{tabular}
\end{table}

\subsection{Per-Domain Training Budget (SDXL)}

The same checkpoint selection protocol applies to the SDXL experiments. \cref{tab:gpu-time-sdxl} lists the estimated GPU hours per domain.

\begin{table}[H]
\centering
\caption{\textbf{Estimated GPU hours per domain (Animagine XL 3.1).} The same checkpoint selection protocol as FLUX applies: actual models used are the closest saved checkpoints to the listed times, with deviation not exceeding 30 minutes.}
\label{tab:gpu-time-sdxl}
\begin{tabular}{@{}lcc@{}}
\toprule
\textbf{Domain} & $\boldsymbol{N_1}$ \textbf{(hours)} & $\boldsymbol{N_2}$ \textbf{(hours)} \\
\midrule
$\alpha$ & 2 & 3 \\
$\beta$  & 2 & 3 \\
$\gamma$ & 1.5 & 2.5 \\
\bottomrule
\end{tabular}
\end{table}

\subsection{H-SD Data Processing Overhead}

The Hierarchical Semantic Decomposition (H-SD) pipeline introduces a one-time preprocessing step per dataset.
Leveraging the mature GPU-accelerated ecosystem---including YOLO-based detectors for semantic parsing and ESRGAN algorithms for super-resolution---the end-to-end processing time is approximately \textbf{15--30~minutes} per dataset on the same hardware.
This overhead is negligible compared to the training durations reported in \cref{tab:gpu-time,tab:gpu-time-sdxl} and is amortized across all subsequent training runs.

\section{Extended Ablation Study: Qualitative Analysis}
\label{appendix:g}

This section complements the quantitative ablation results reported in the main text (Table~2) with detailed qualitative comparisons. For each architecture, we visualize the outputs of the three ablation variants---\emph{Full SGA}, \emph{w/o Tuple-wise Optimization}, and \emph{w/o Scale-Adaptive Modulation}---under identical prompts and seeds.

\subsection{Reference Images}

\begin{figure}[H]
\centering
\includegraphics[height=3.8cm]{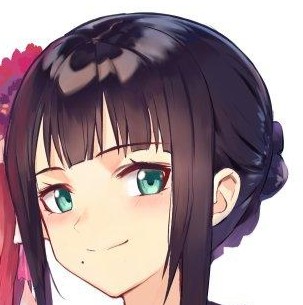}\hspace{6pt}
\includegraphics[height=3.8cm]{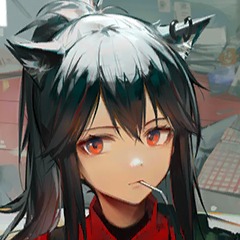}\hspace{6pt}
\includegraphics[height=3.8cm]{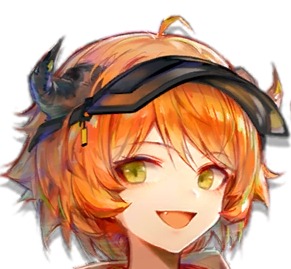}\hspace{6pt}
\includegraphics[height=3.8cm]{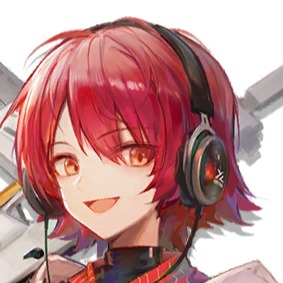}
\caption{\textbf{Reference images} defining the target domain for all ablation experiments.}
\label{fig:ablation-ref}
\end{figure}

\cref{fig:ablation-ref} presents representative reference images that define the target domain for all ablation experiments. These images share consistent stylistic attributes that the fine-tuned model is expected to reproduce.

\subsection{FLUX Ablation}

\begin{figure}[H]
\centering
\begin{tabular}{@{}c@{\hspace{2pt}}c@{\hspace{2pt}}c@{}}
\includegraphics[width=0.32\linewidth]{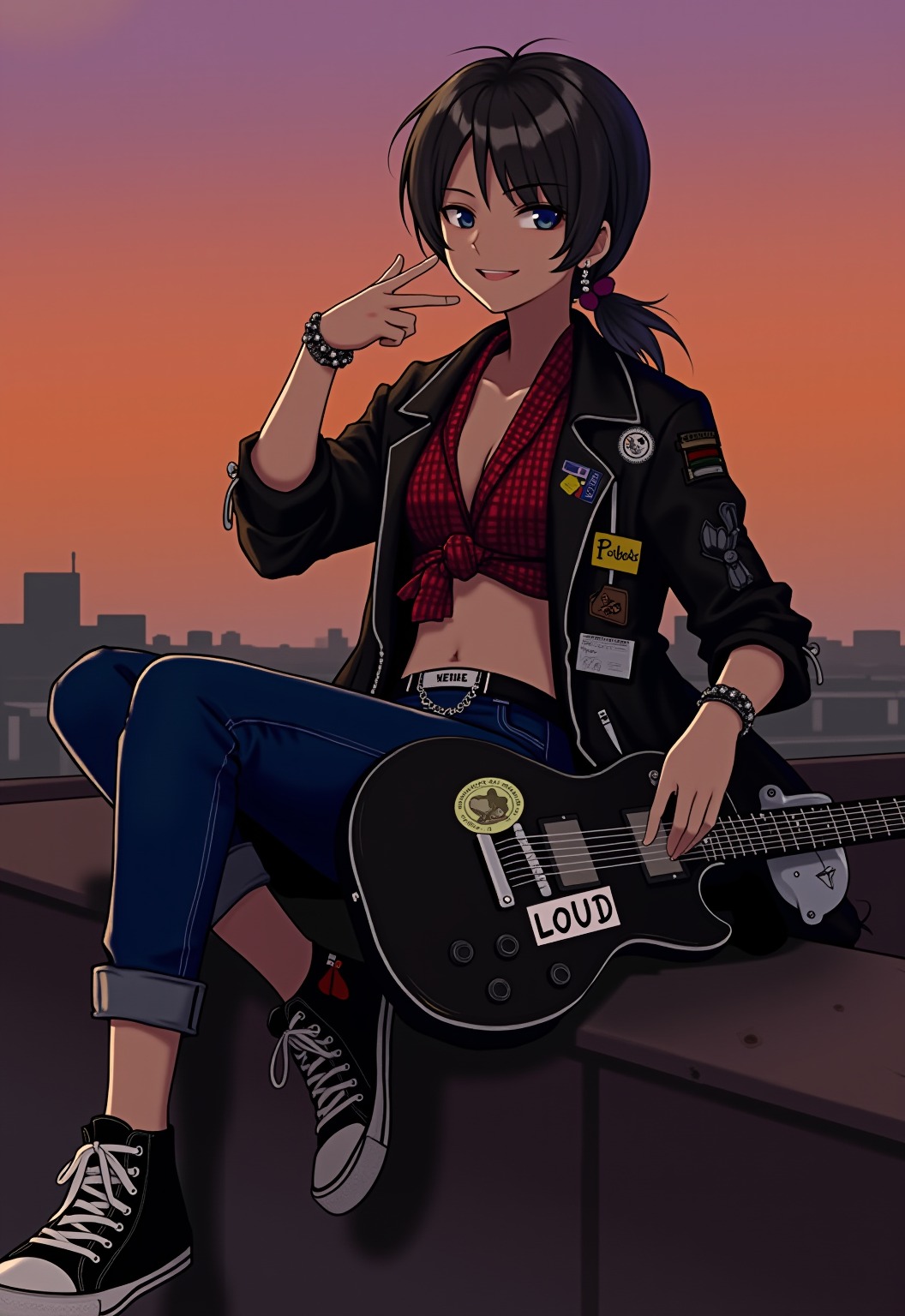} &
\includegraphics[width=0.32\linewidth]{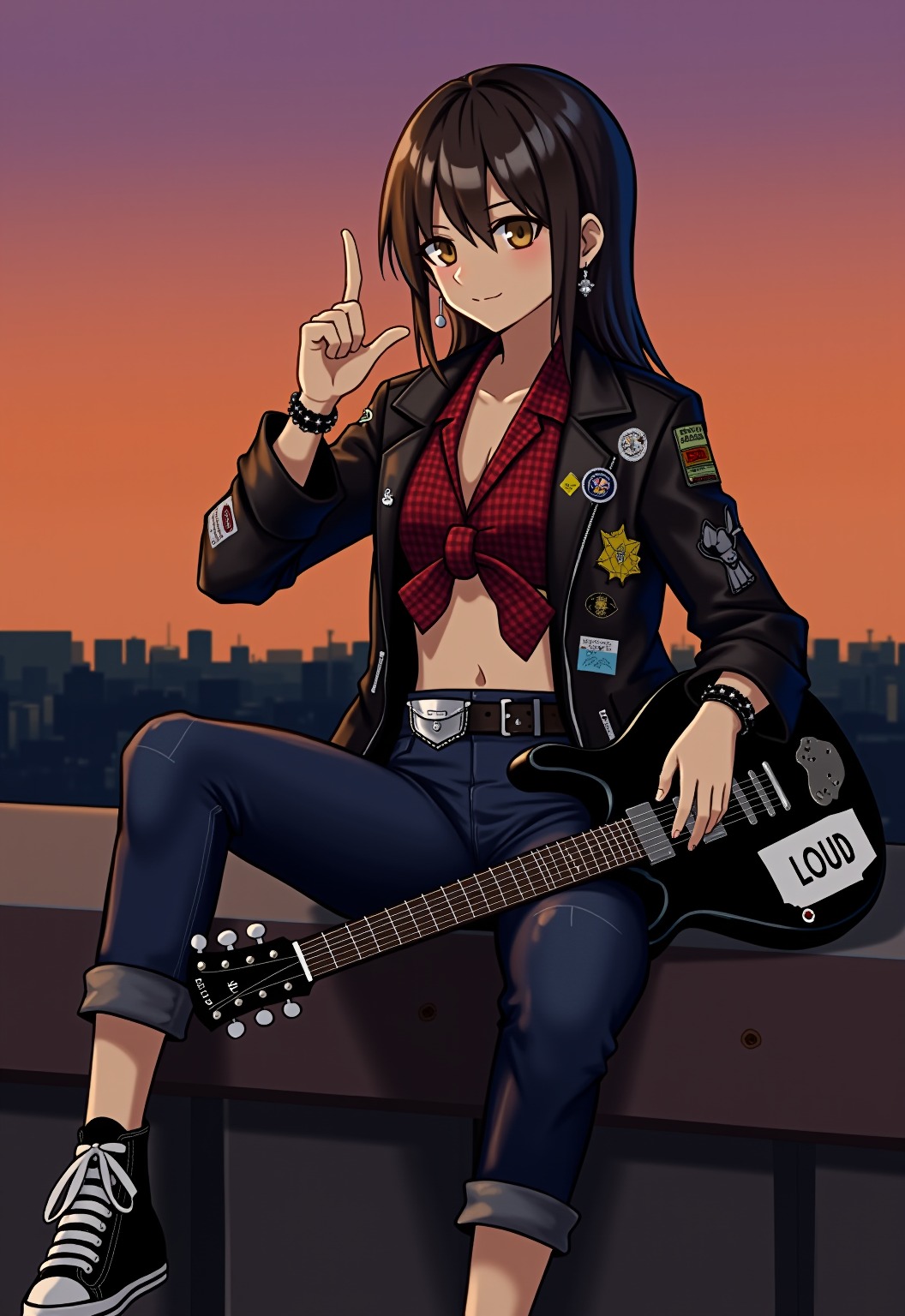} &
\includegraphics[width=0.32\linewidth]{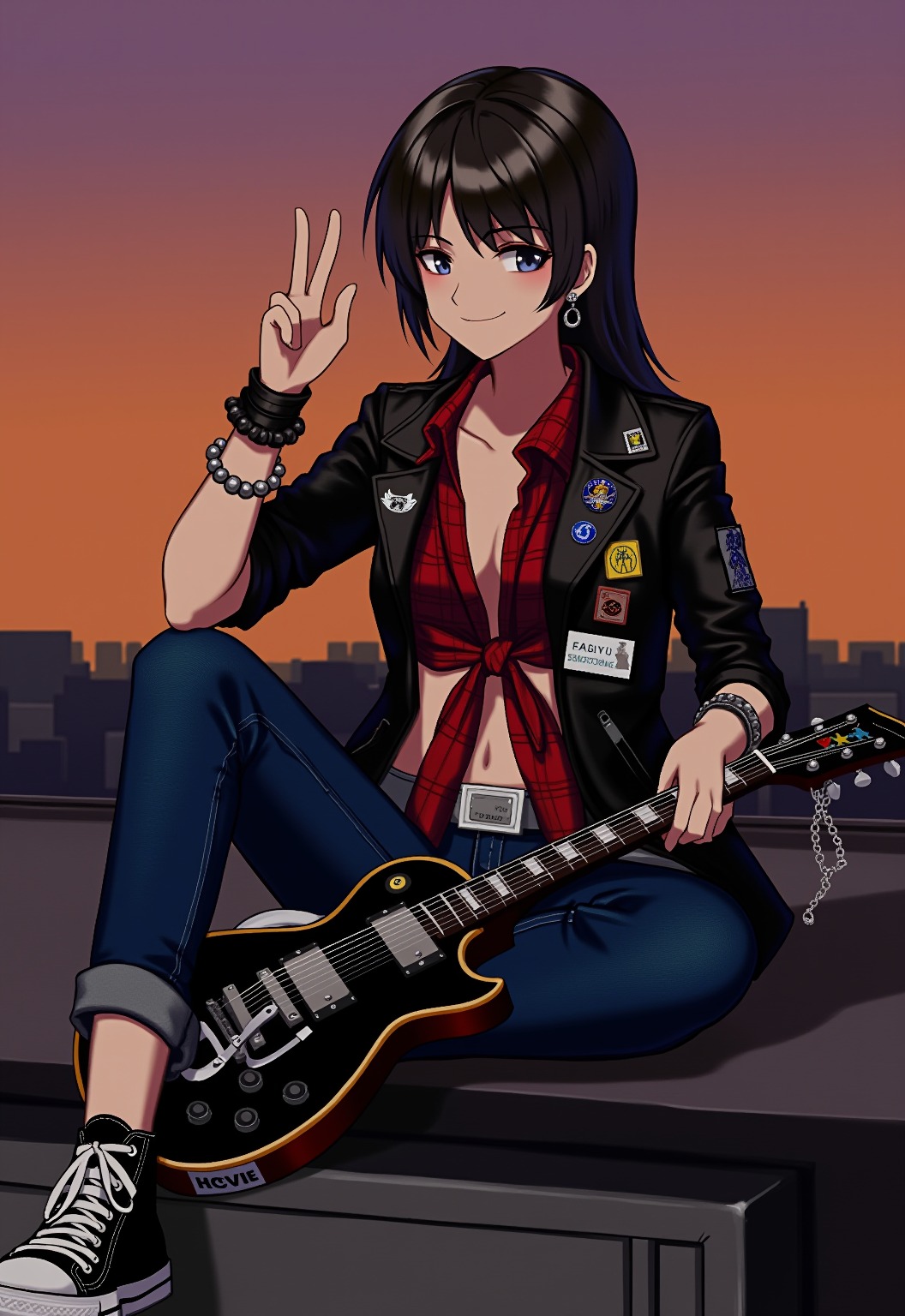} \\
{\small Full SGA} & {\small w/o Tuple-wise Opt.} & {\small w/o Scale-Adapt. Mod.} \\
\end{tabular}
\caption{\textbf{FLUX ablation (Group 1).} Full SGA maintains domain fidelity; removing Scale-Adaptive Modulation introduces subtle but visible stylistic drift.}
\label{fig:ablation-flux-1}
\end{figure}

\begin{figure}[H]
\centering
\begin{tabular}{@{}c@{\hspace{2pt}}c@{\hspace{2pt}}c@{}}
\includegraphics[width=0.32\linewidth]{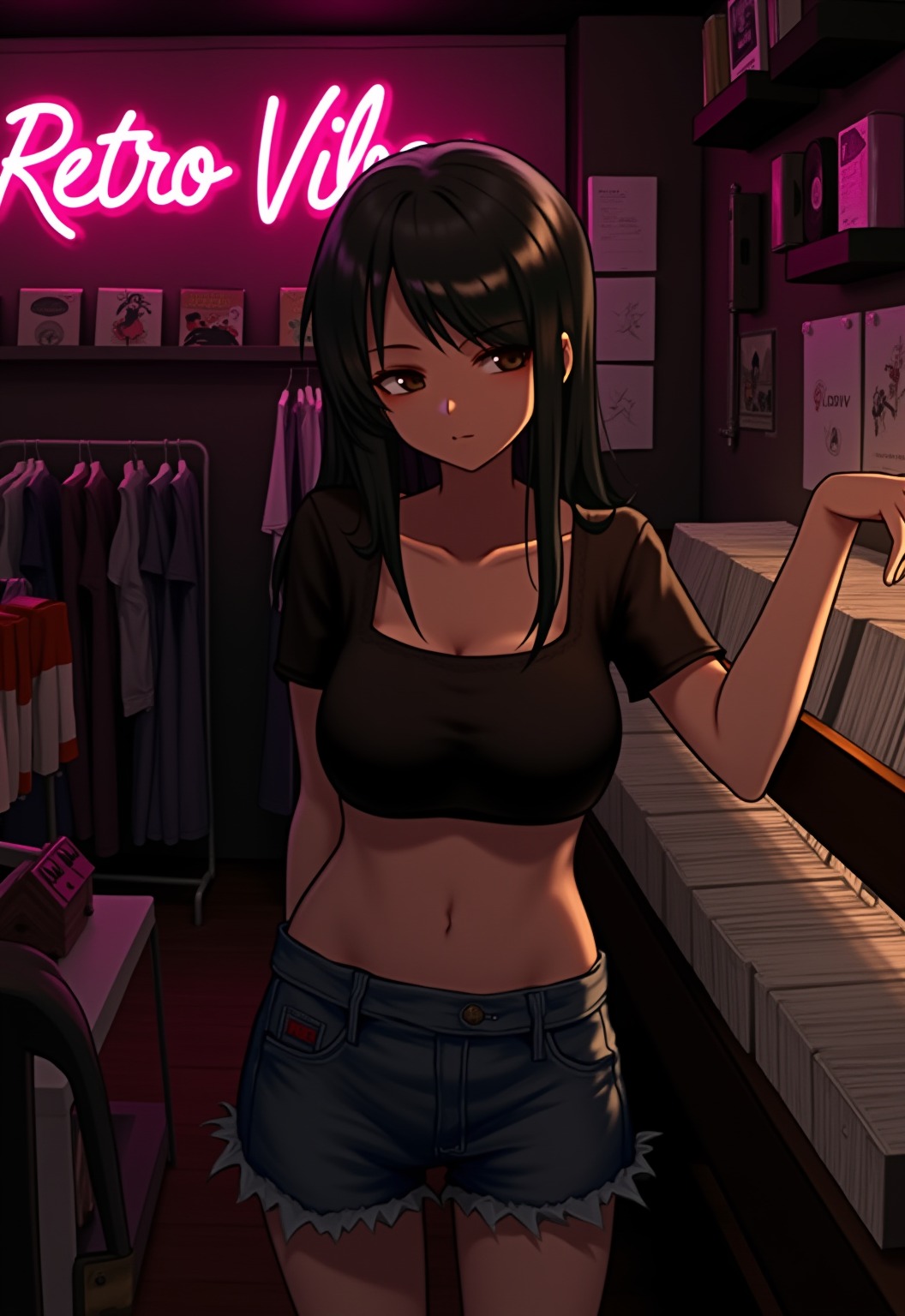} &
\includegraphics[width=0.32\linewidth]{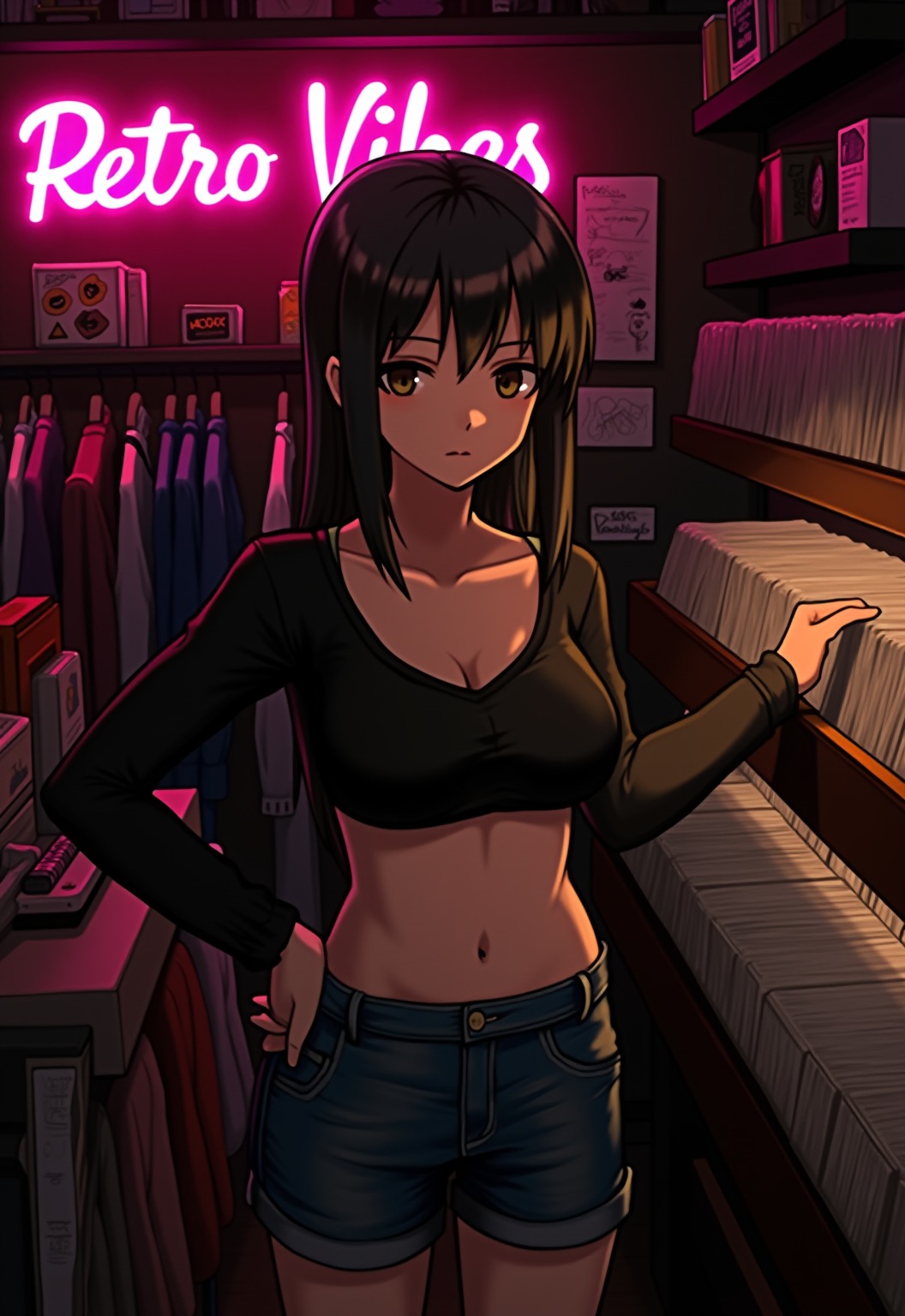} &
\includegraphics[width=0.32\linewidth]{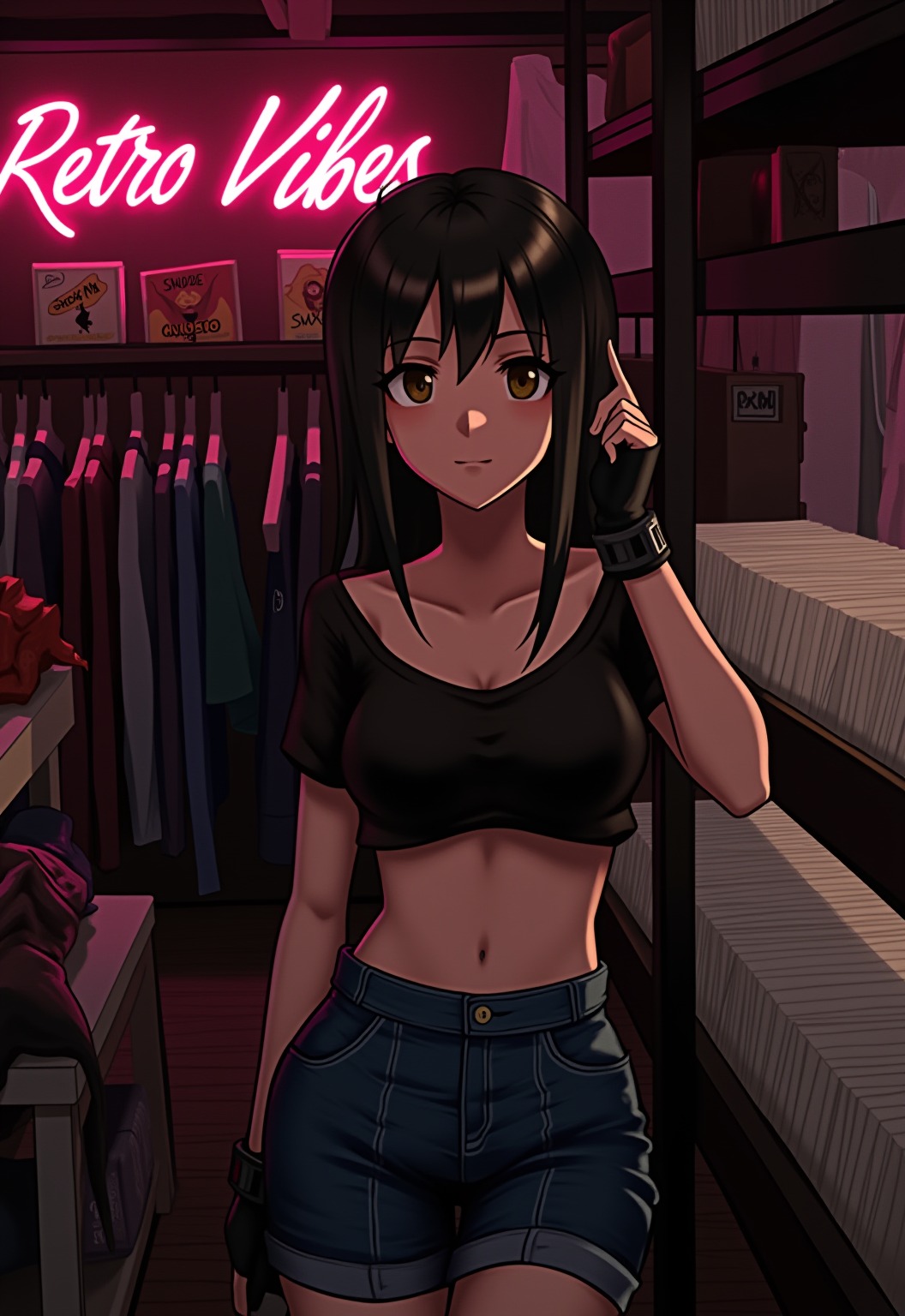} \\
{\small Full SGA} & {\small w/o Tuple-wise Opt.} & {\small w/o Scale-Adapt. Mod.} \\
\end{tabular}
\caption{\textbf{FLUX ablation (Group 2).} The w/o Scale-Adaptive Modulation variant exhibits increased OOD tendencies in fine-grained detail rendering.}
\label{fig:ablation-flux-2}
\end{figure}

\begin{figure}[H]
\centering
\begin{tabular}{@{}c@{\hspace{2pt}}c@{\hspace{2pt}}c@{}}
\includegraphics[width=0.32\linewidth]{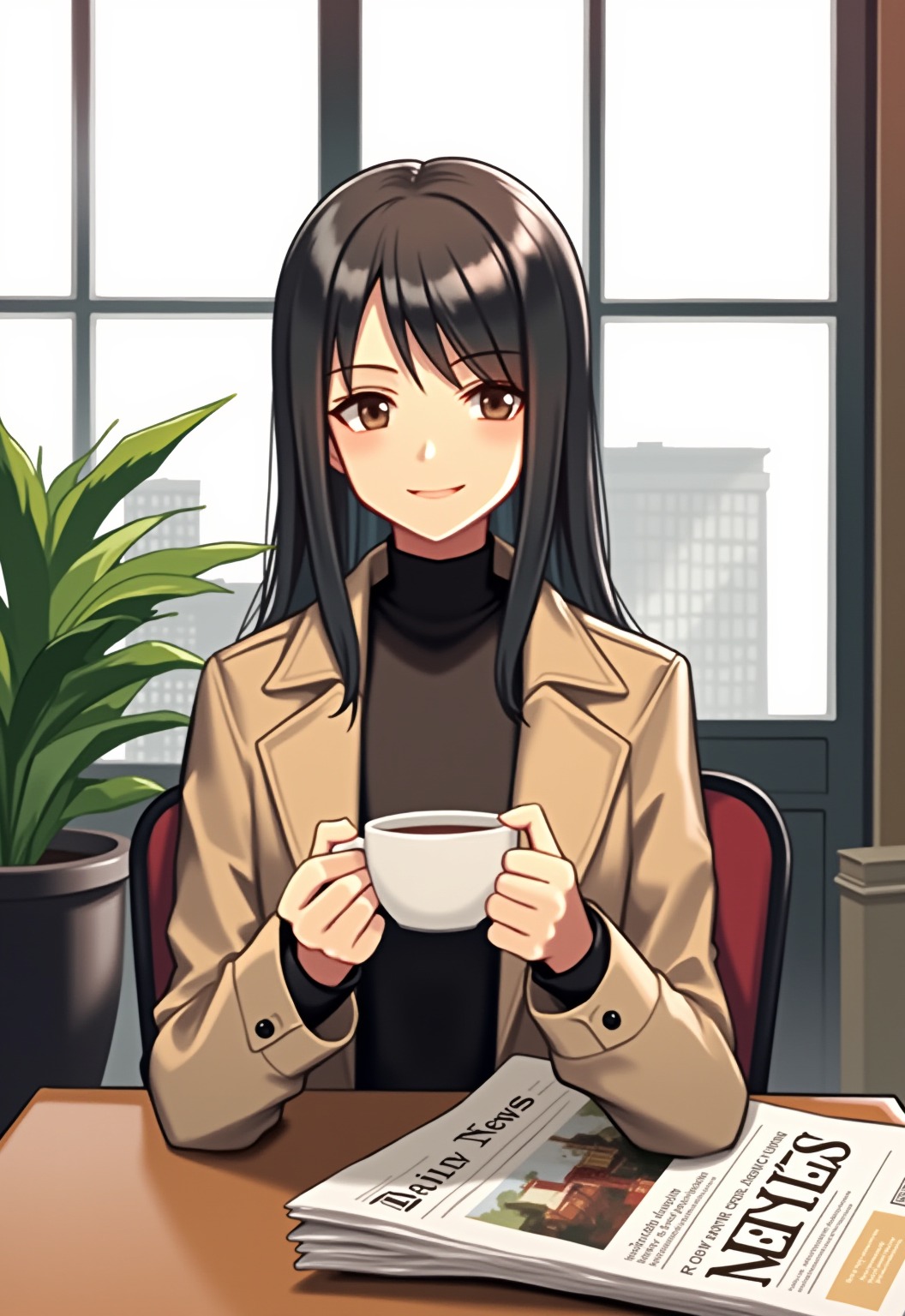} &
\includegraphics[width=0.32\linewidth]{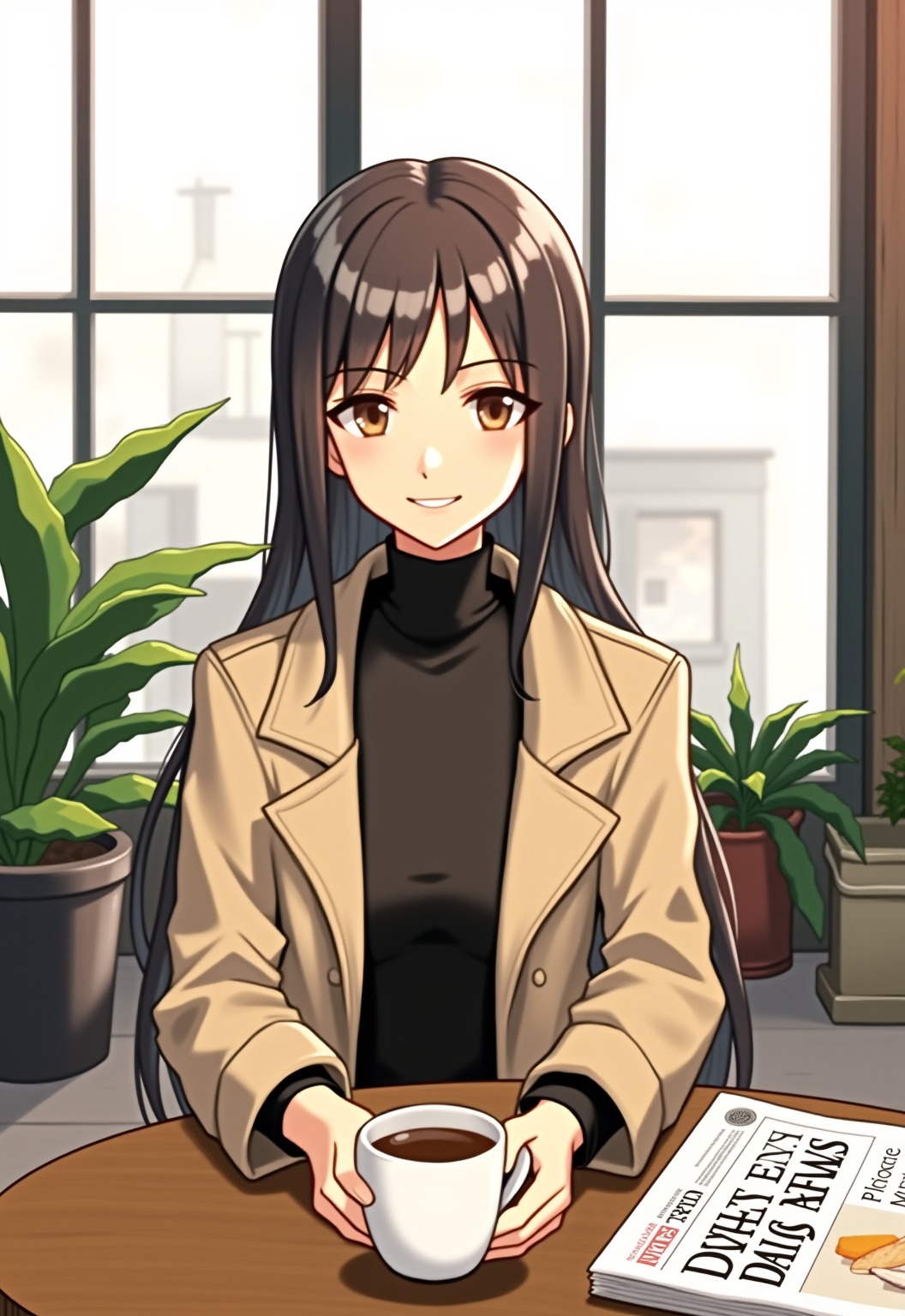} &
\includegraphics[width=0.32\linewidth]{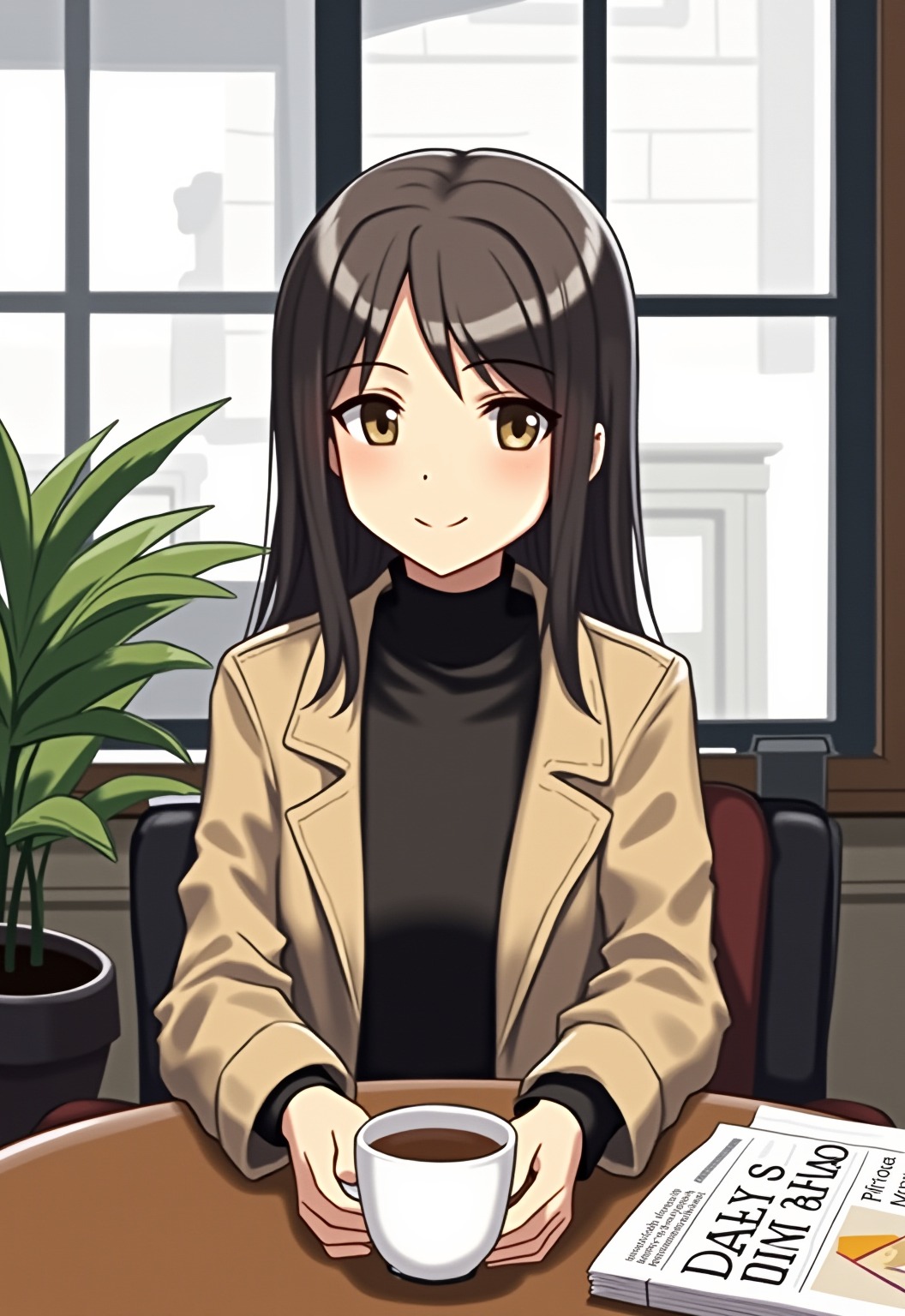} \\
{\small Full SGA} & {\small w/o Tuple-wise Opt.} & {\small w/o Scale-Adapt. Mod.} \\
\end{tabular}
\caption{\textbf{FLUX ablation (Group 3).} All three variants produce structurally coherent outputs---consistent with FLUX's global attention mechanism---but the Full SGA variant best preserves the target domain's micro-level characteristics.}
\label{fig:ablation-flux-3}
\end{figure}

\cref{fig:ablation-flux-1,fig:ablation-flux-2,fig:ablation-flux-3} present three groups of ablation comparisons on the FLUX backbone. A key observation is that when Scale-Adaptive Modulation is removed (\ie, the granularity-dependent time-step bias is disabled), the out-of-distribution (OOD) rate increases noticeably: the generated outputs frequently deviate from the target domain characteristics, producing outputs that fall outside the learned stylistic manifold.

In contrast, the gap between \emph{Full SGA} and \emph{w/o Tuple-wise Optimization} is comparatively smaller on FLUX. We attribute this to two factors:
\begin{enumerate}
\item \textbf{Implicit SGA from ARB bucketing under large batch sizes.}
With a batch size of 8, the Aspect-Ratio Bucketing (ARB) mechanism naturally groups images of heterogeneous aspect ratios into each optimization step.
This diverse spatial composition acts as an implicit form of cross-granularity co-sampling, partially substituting the role of explicit Tuple-wise Optimization.
\item \textbf{Global attention robustness.}
The MM-DiT architecture~\cite{peebles2023scalable} employs global self-attention across all spatial tokens, granting inherently stronger structural coherence than local-receptive-field architectures and allowing the model to maintain compositional consistency without explicit cross-scale synchronization.
\end{enumerate}

\subsection{FLUX LoRA Ablation: Effect of Scale-Adaptive Modulation}

\begin{figure}[H]
\centering
\begin{tabular}{@{}c@{\hspace{4pt}}c@{}}
\includegraphics[height=5.5cm]{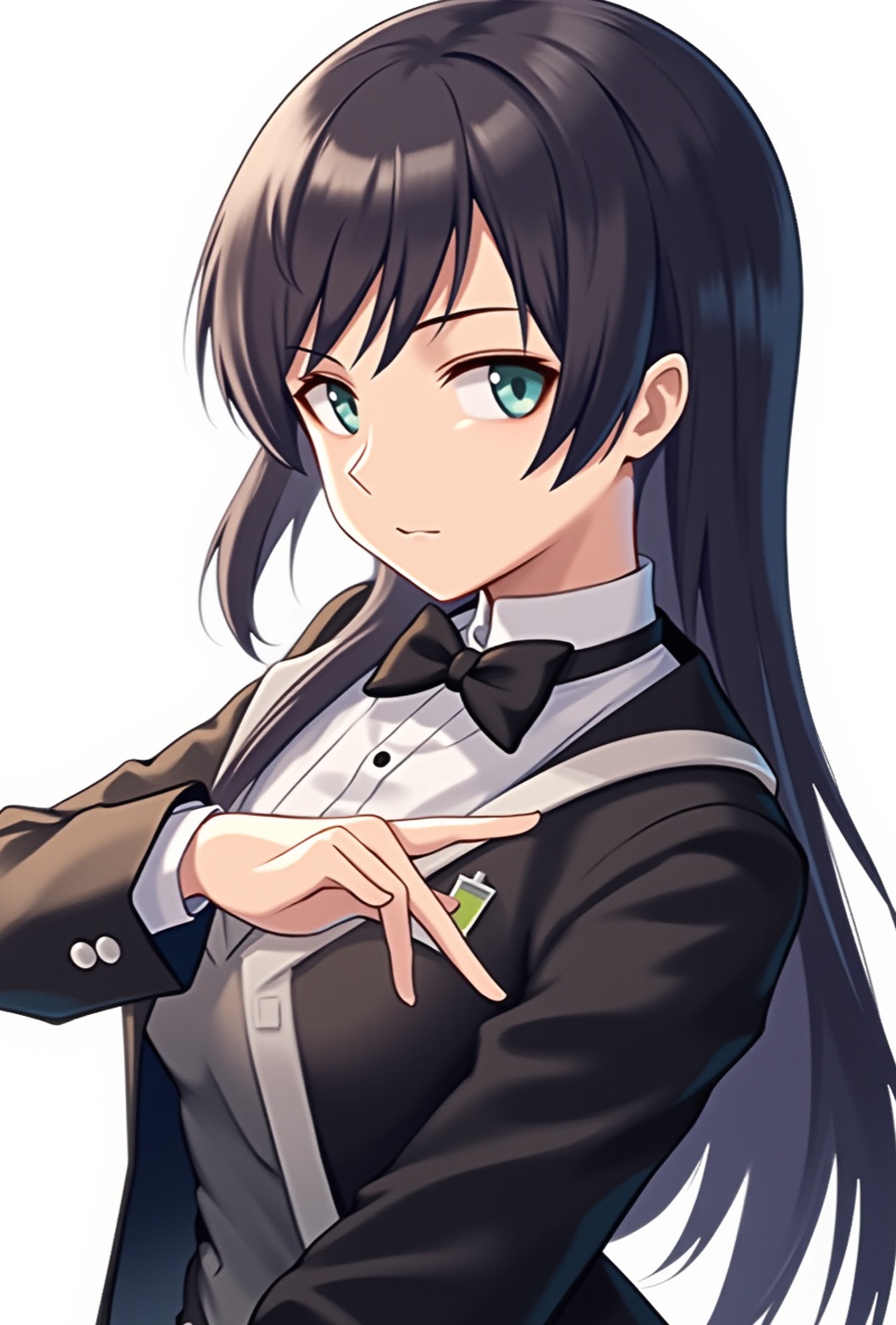} &
\includegraphics[height=5.5cm]{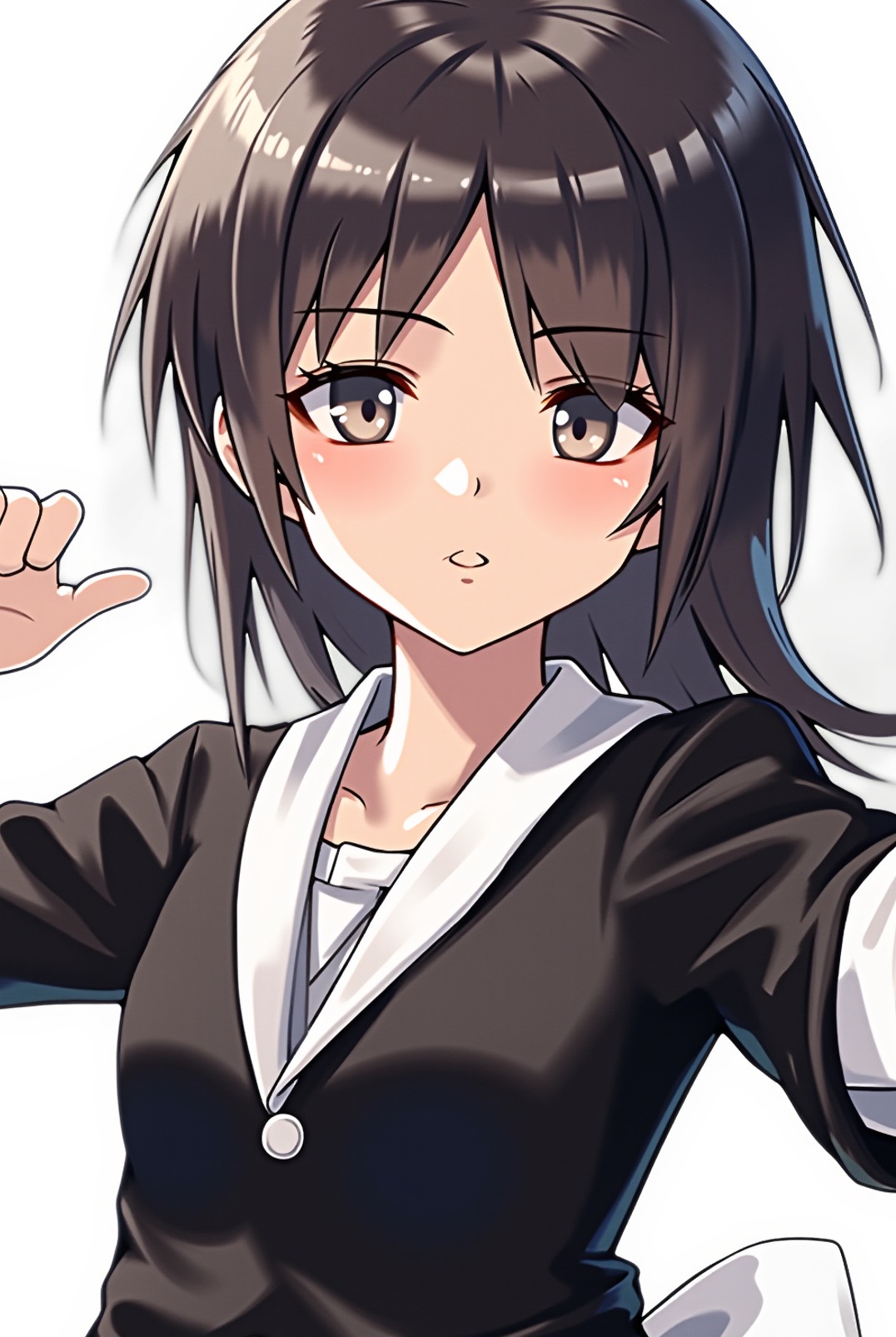} \\
{\small LoRA + Full SGA} & {\small LoRA + w/o Scale-Adapt. Mod.} \\
\end{tabular}
\caption{\textbf{LoRA Ablation (Group 1).} LoRA without Modulation produces visible OOD drift.}
\label{fig:ablation-lora-1}
\end{figure}

\begin{figure}[H]
\centering
\begin{tabular}{@{}c@{\hspace{4pt}}c@{}}
\includegraphics[height=5.5cm]{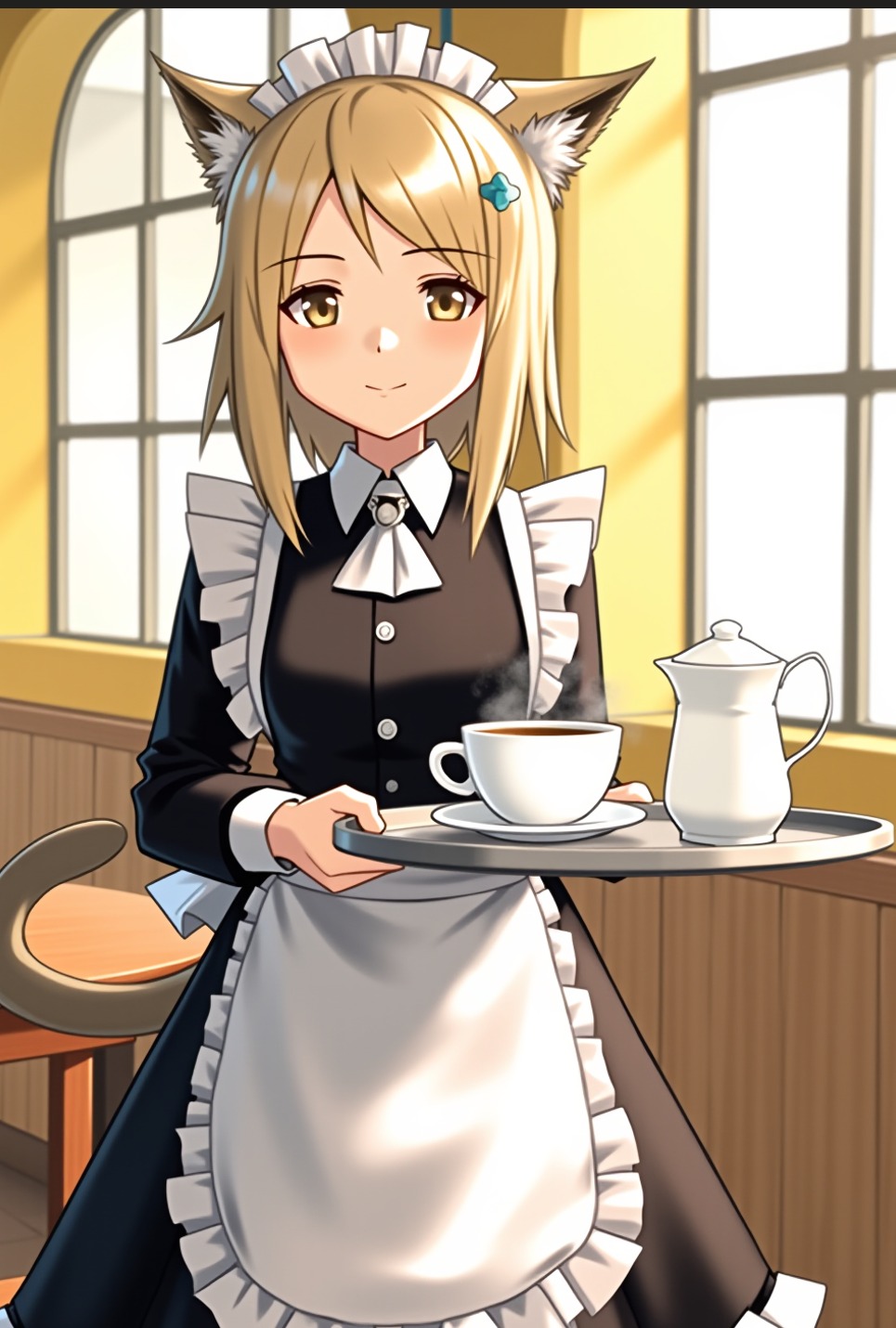} &
\includegraphics[height=5.5cm]{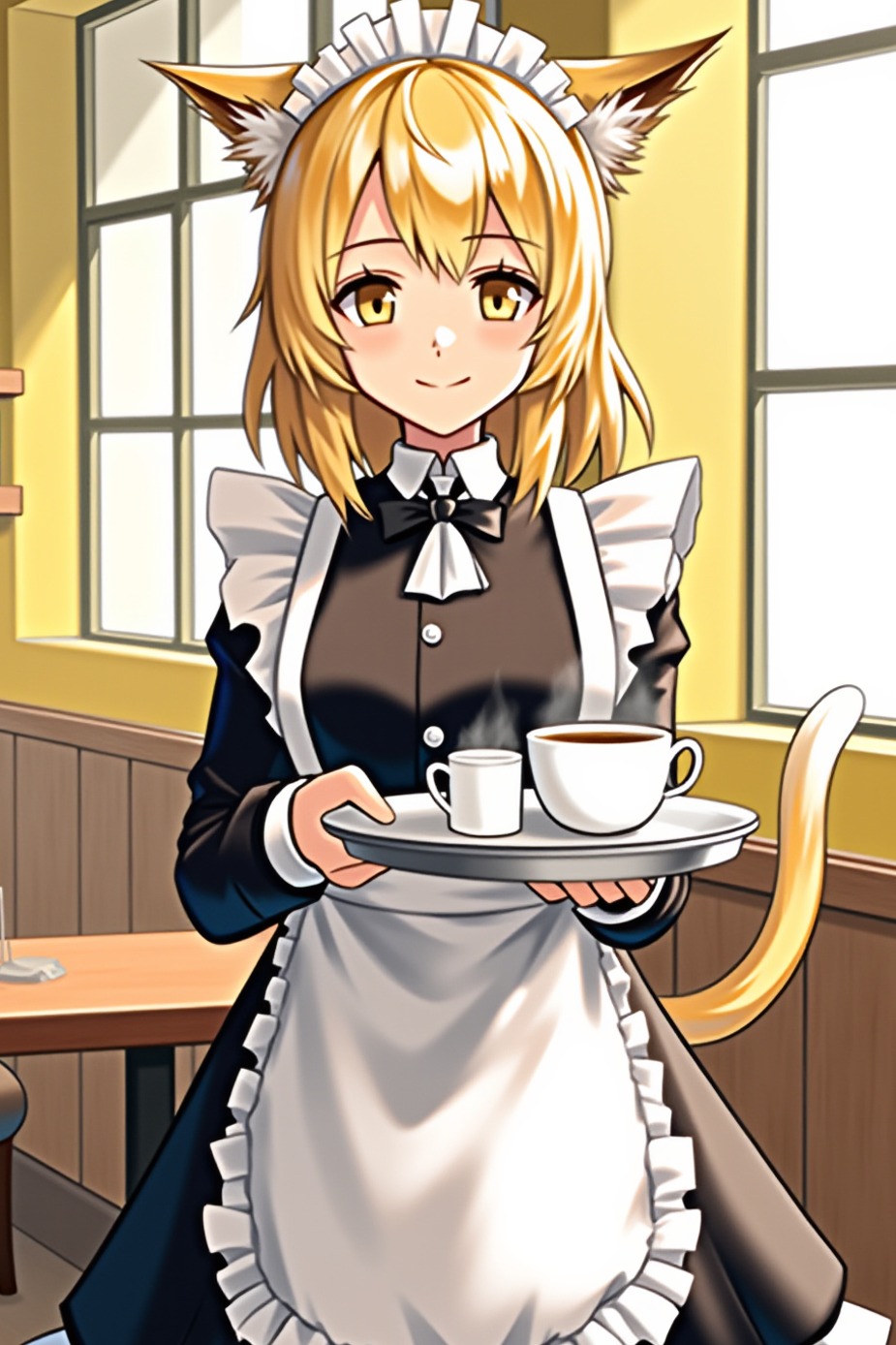} \\
{\small LoRA + Full SGA} & {\small LoRA + w/o Scale-Adapt. Mod.} \\
\end{tabular}
\caption{\textbf{LoRA Ablation (Group 2).} LoRA exhibits more pronounced domain deviation due to coupled updates.}
\label{fig:ablation-lora-2}
\end{figure}

\begin{figure}[H]
\centering
\begin{tabular}{@{}c@{\hspace{4pt}}c@{}}
\includegraphics[height=4.5cm]{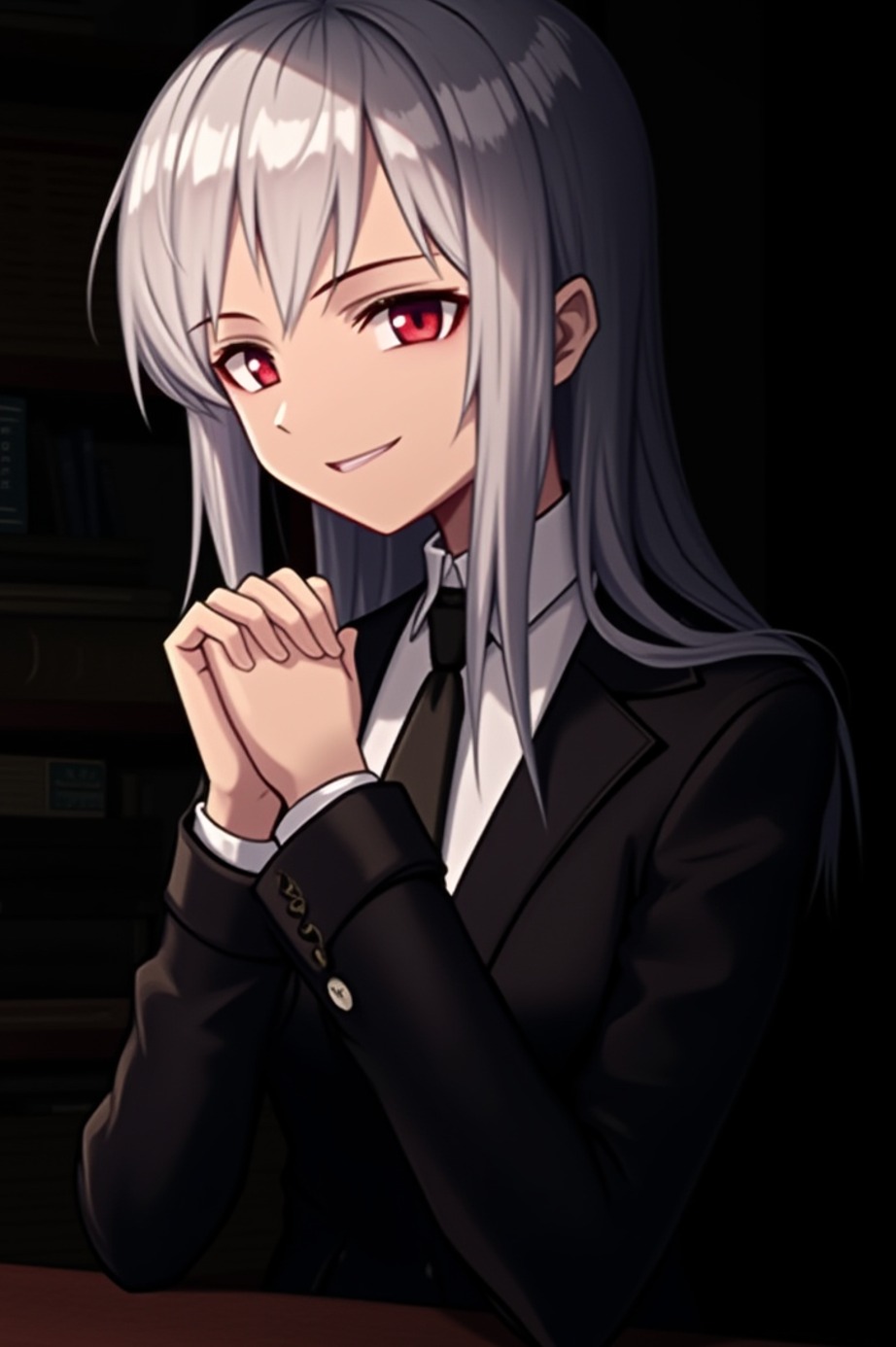} &
\includegraphics[height=4.5cm]{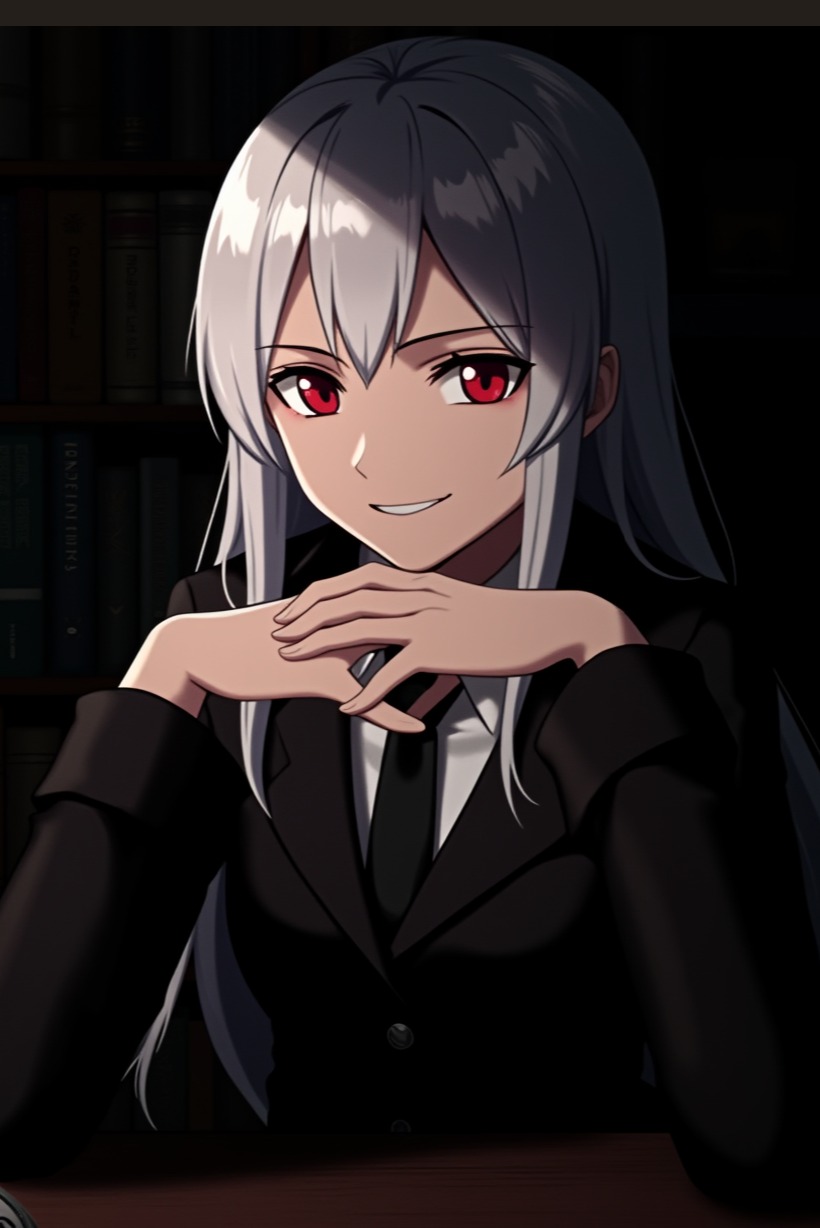} \\
{\scriptsize LoRA + Full SGA} & {\scriptsize LoRA + w/o Mod.} \\[4pt]
\includegraphics[height=4.5cm]{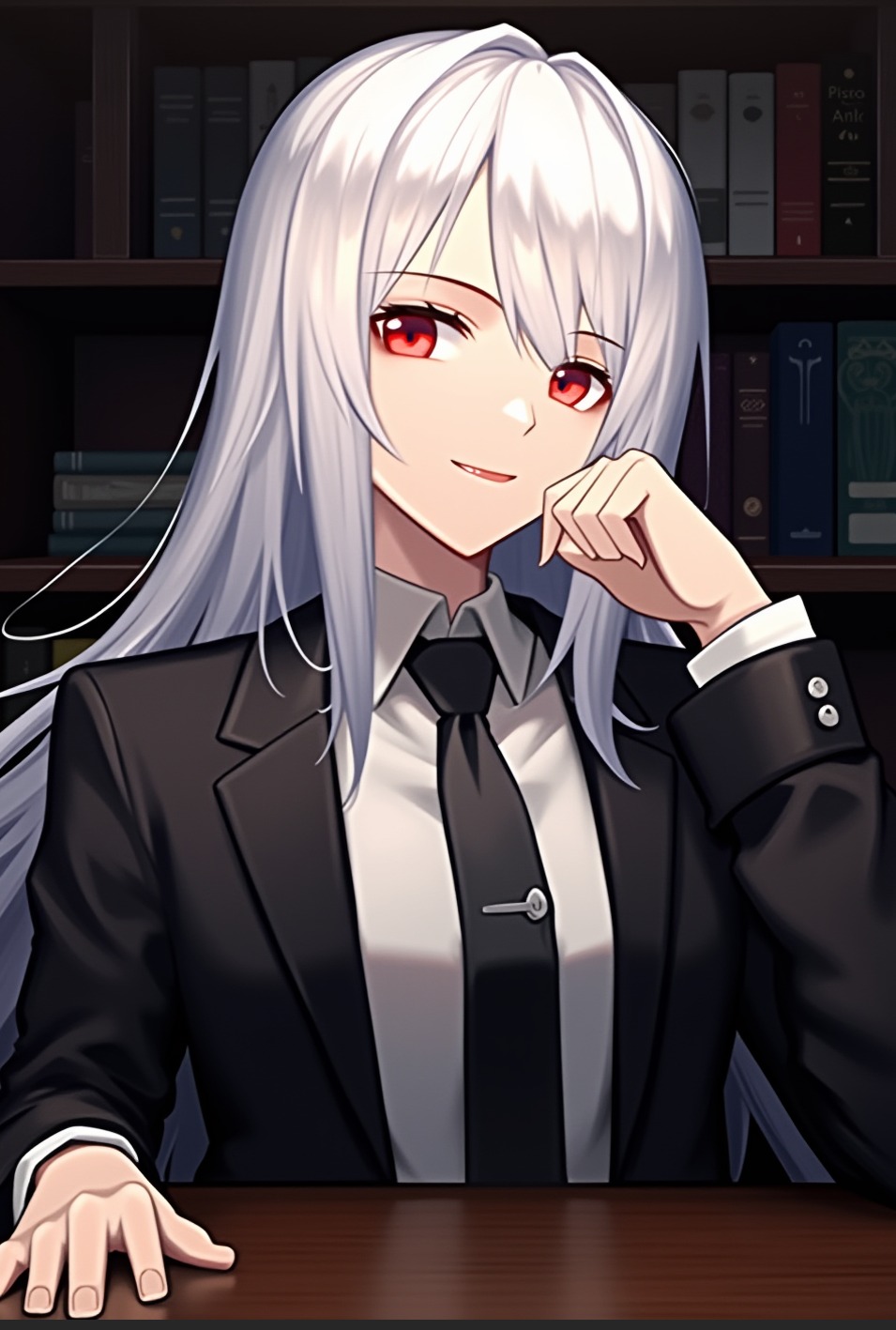} &
\includegraphics[height=4.5cm]{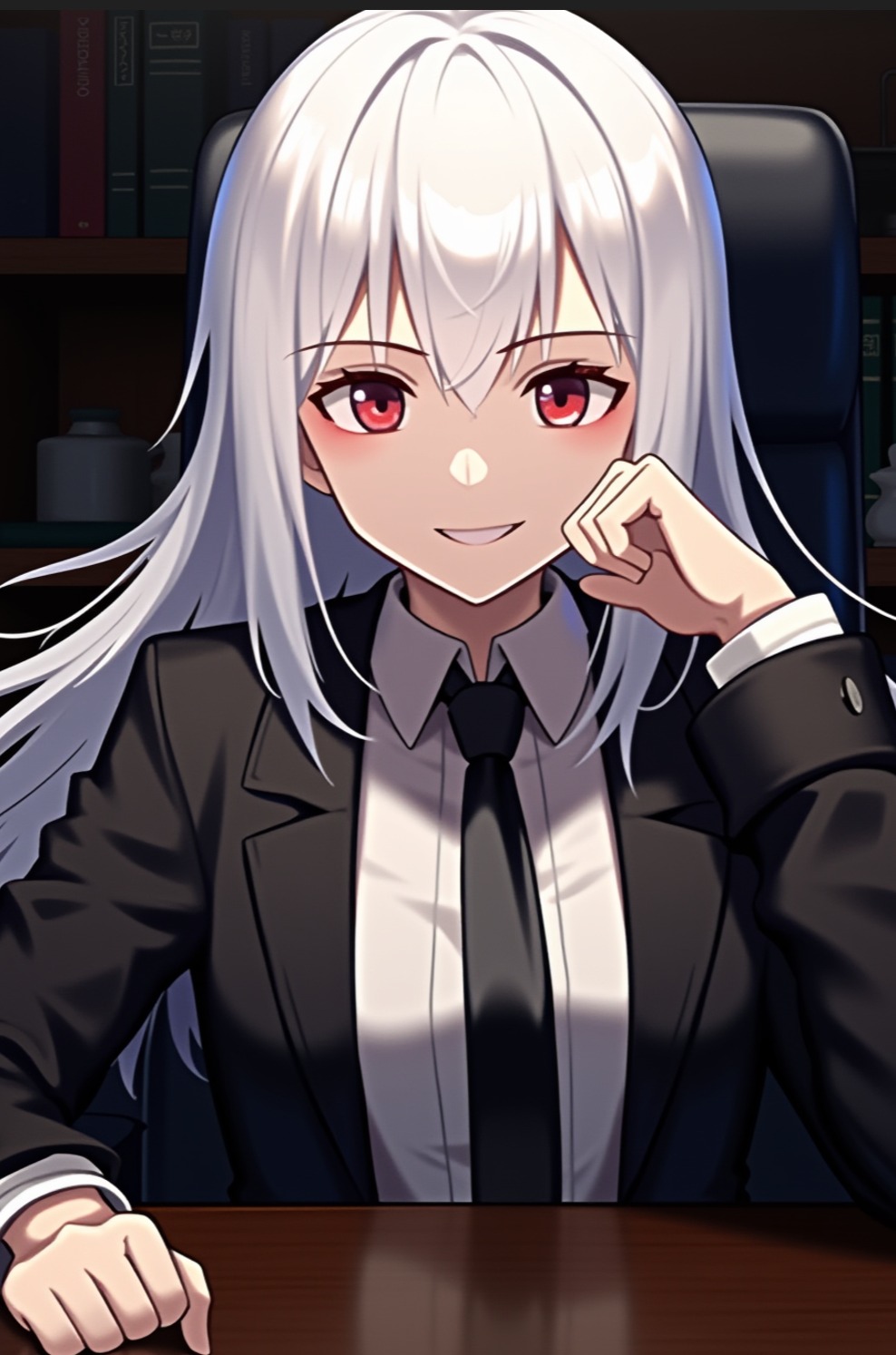} \\
{\scriptsize LoRA + Full SGA} & {\scriptsize LoRA + w/o Mod.} \\
\end{tabular}
\caption{\textbf{LoRA Ablation (Group 3).} LoRA without Scale-Adaptive Modulation (right column) shows the most pronounced OOD deviation.}
\label{fig:ablation-lora-3}
\end{figure}

We conduct an analogous ablation using standard LoRA~\cite{hu2022lora} as the adaptation method (\cref{fig:ablation-lora-1,fig:ablation-lora-2,fig:ablation-lora-3}). Compared to the DoRA-based results in \cref{fig:ablation-flux-1,fig:ablation-flux-2,fig:ablation-flux-3}, removing Scale-Adaptive Modulation under LoRA leads to \emph{considerably more pronounced} OOD drift. We attribute this to LoRA's single low-rank subspace, which couples magnitude and direction updates, increasing susceptibility to cross-scale interference. These results suggest that Scale-Adaptive Modulation is especially critical when using standard LoRA.

\subsection{SDXL Component Ablation}

\begin{figure}[H]
\centering
\begin{tabular}{@{}c@{\hspace{3pt}}c@{\hspace{3pt}}c@{}}
\includegraphics[width=0.32\linewidth]{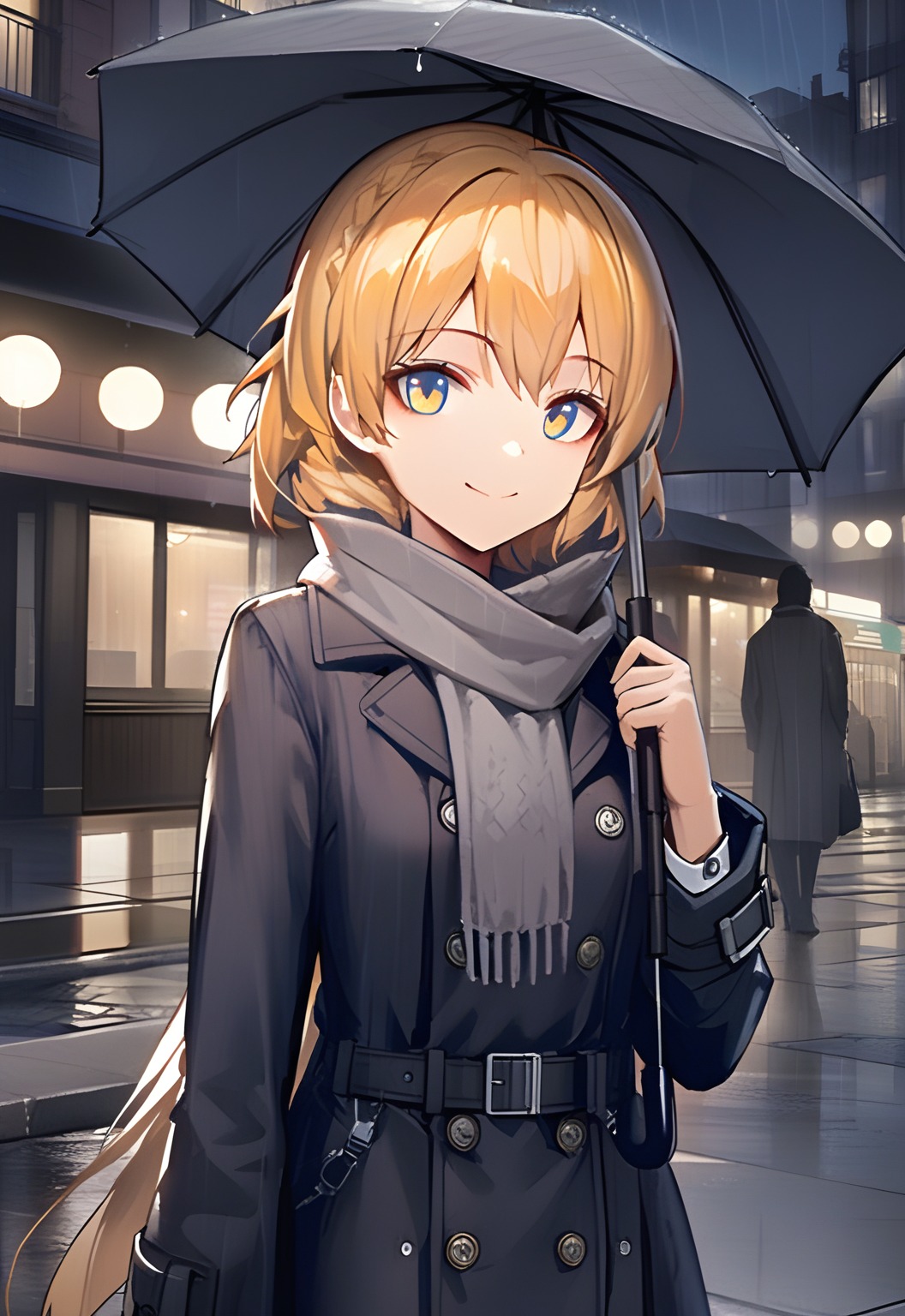} &
\includegraphics[width=0.32\linewidth]{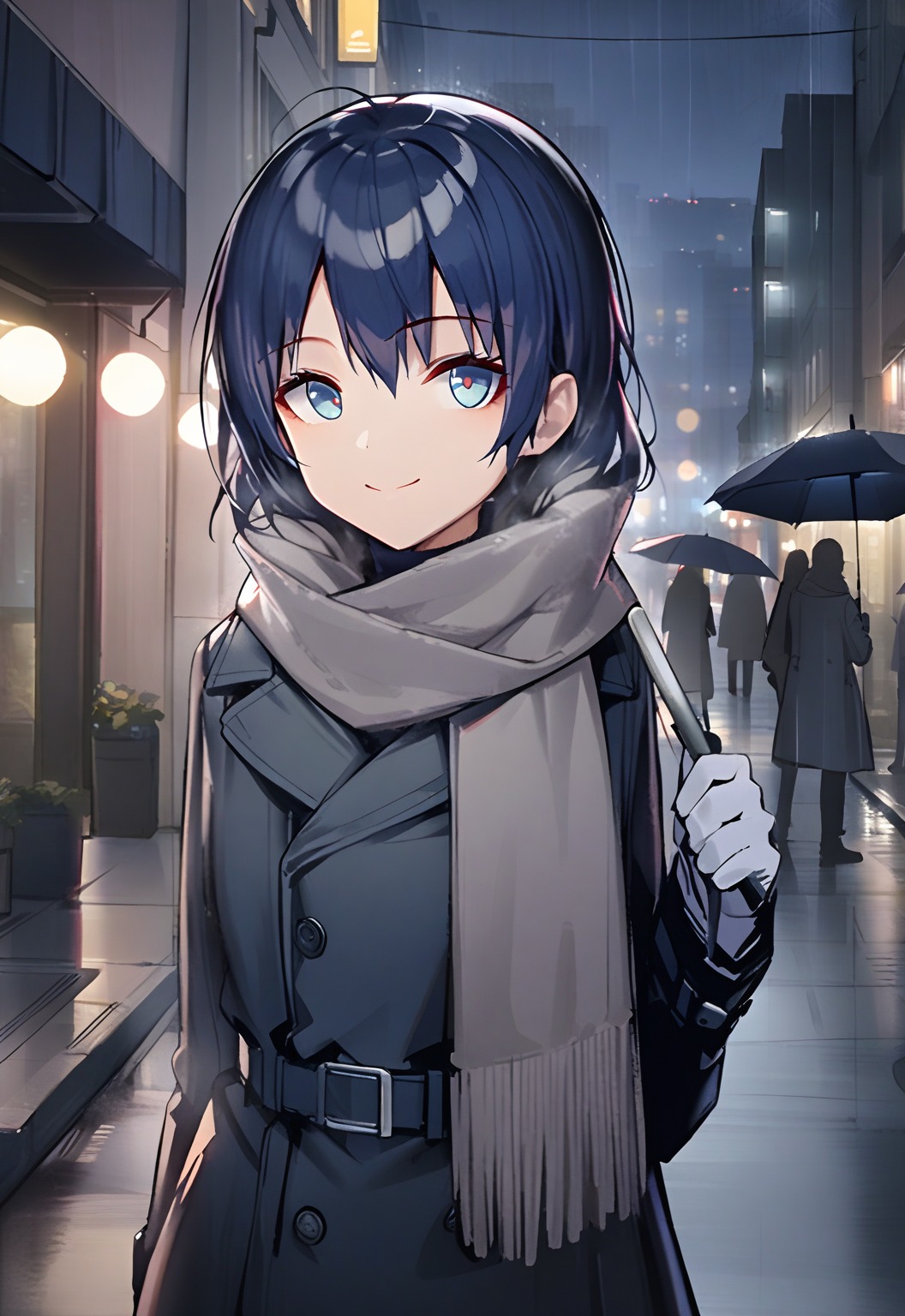} &
\includegraphics[width=0.32\linewidth]{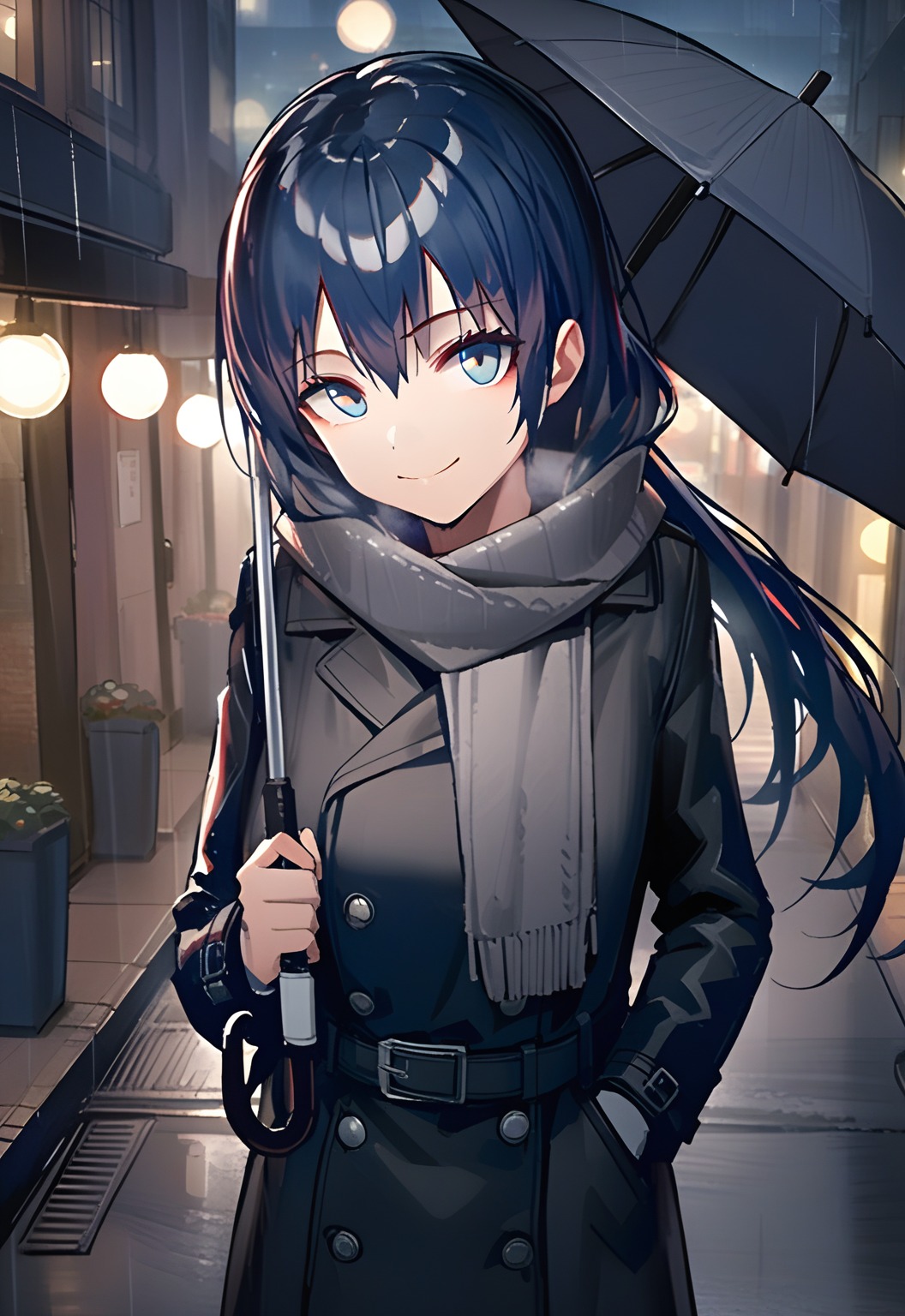} \\
{\small Full SGA} & {\small w/o Tuple-wise Opt.} & {\small w/o Min-SNR} \\
\end{tabular}
\caption{\textbf{SDXL ablation (Group 1).} Removing SGA produces noticeable anatomical errors; removing Min-SNR yields subtler structural degradation.}
\label{fig:ablation-sdxl-1}
\end{figure}

\begin{figure}[H]
\centering
\begin{tabular}{@{}c@{\hspace{3pt}}c@{\hspace{3pt}}c@{}}
\includegraphics[width=0.32\linewidth]{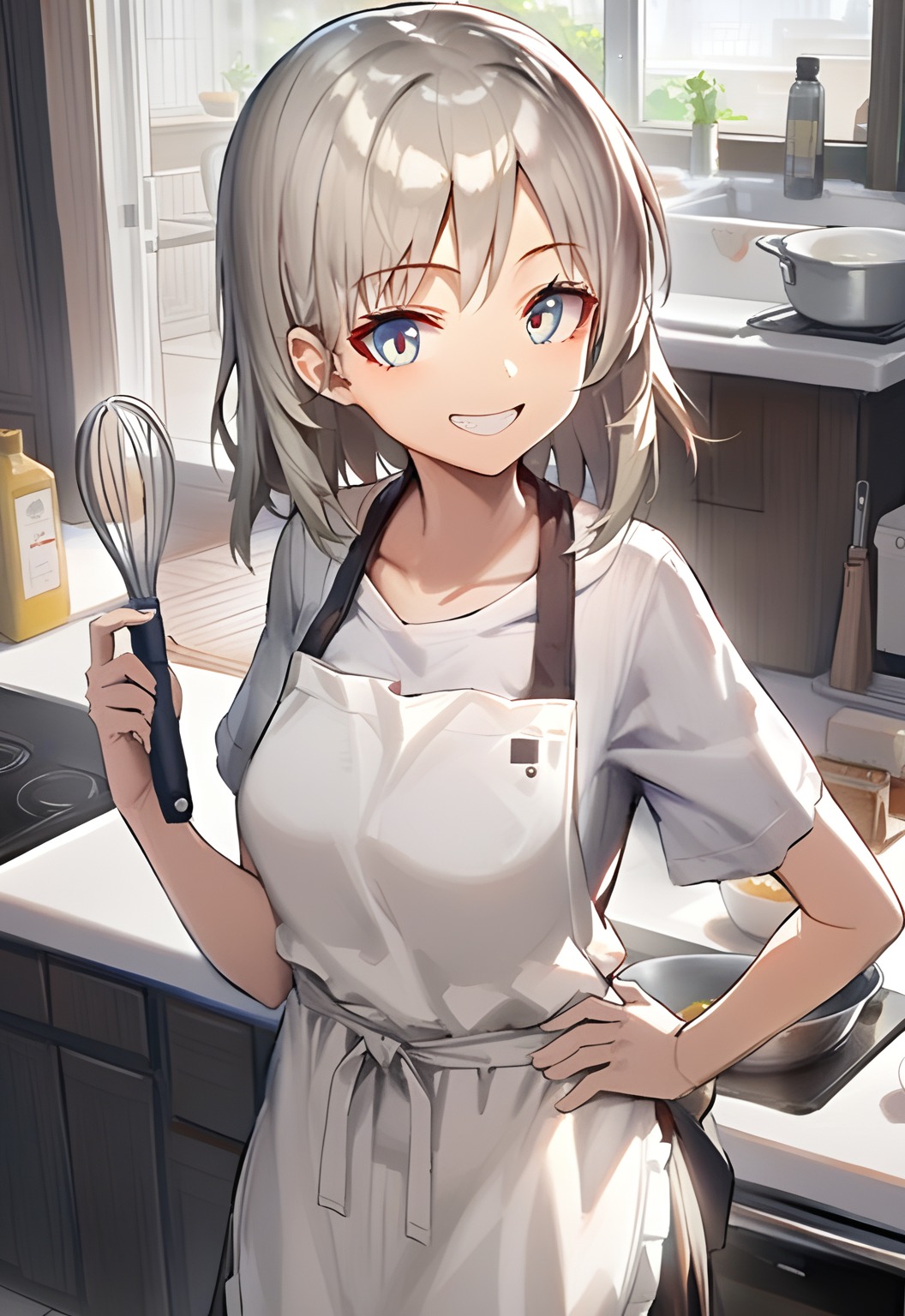} &
\includegraphics[width=0.32\linewidth]{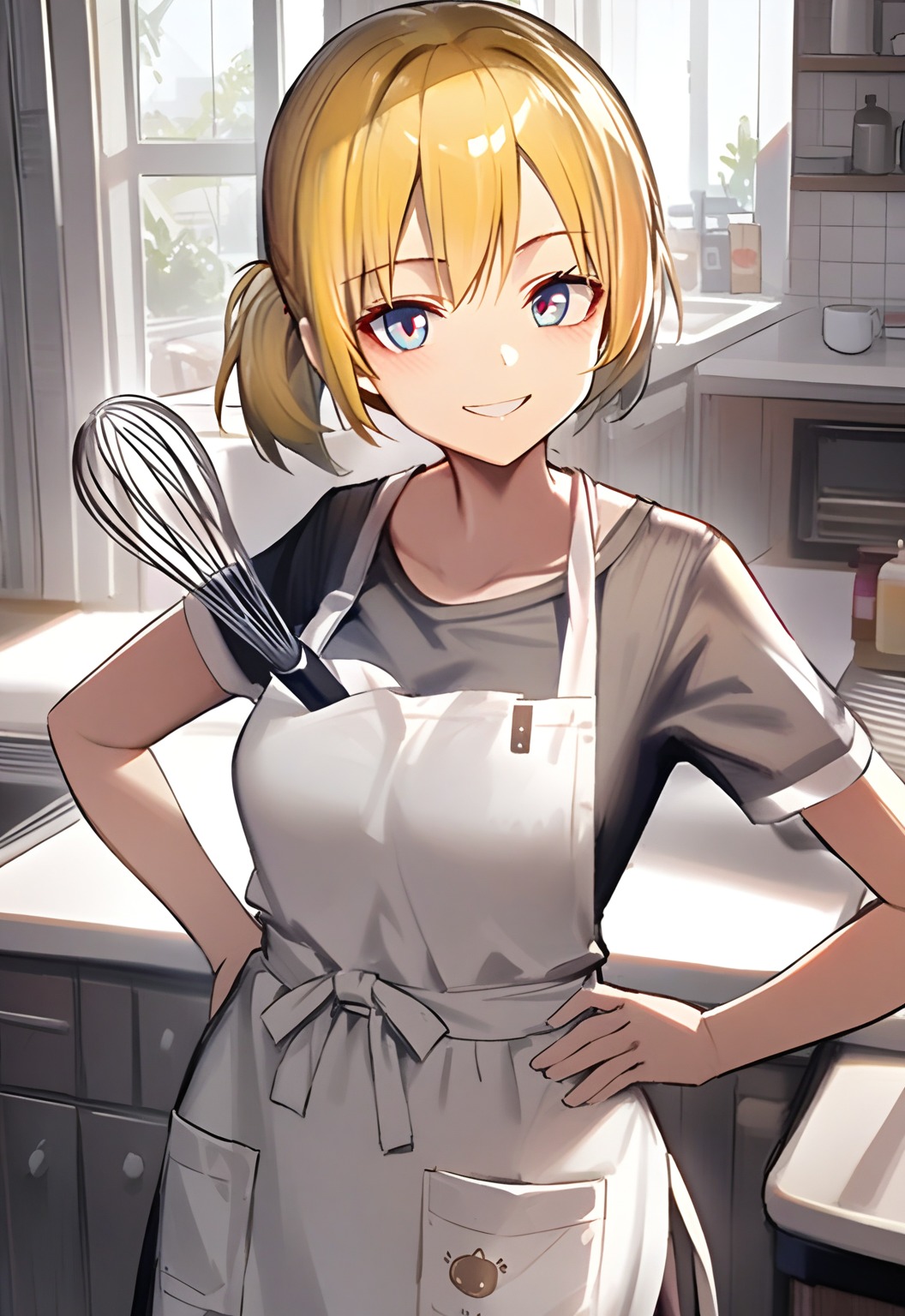} &
\includegraphics[width=0.32\linewidth]{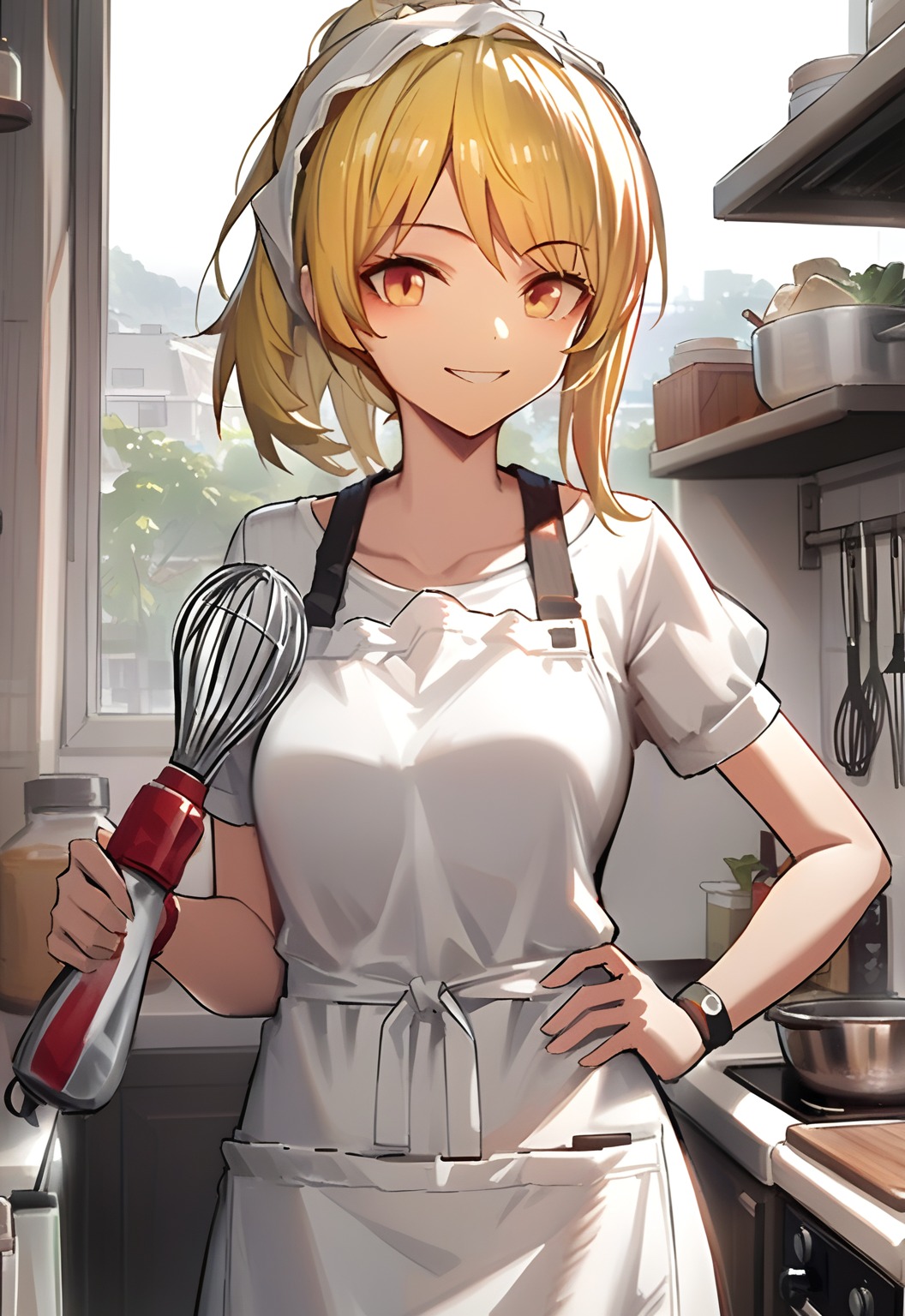} \\
{\small Full SGA} & {\small w/o Tuple-wise Opt.} & {\small w/o Min-SNR} \\
\end{tabular}
\caption{\textbf{SDXL ablation (Group 2).} Full SGA best preserves limb coherence, consistent with CNN-based models requiring explicit structural regularization.}
\label{fig:ablation-sdxl-2}
\end{figure}

\begin{figure}[H]
\centering
\begin{tabular}{@{}c@{\hspace{3pt}}c@{\hspace{3pt}}c@{}}
\includegraphics[width=0.32\linewidth]{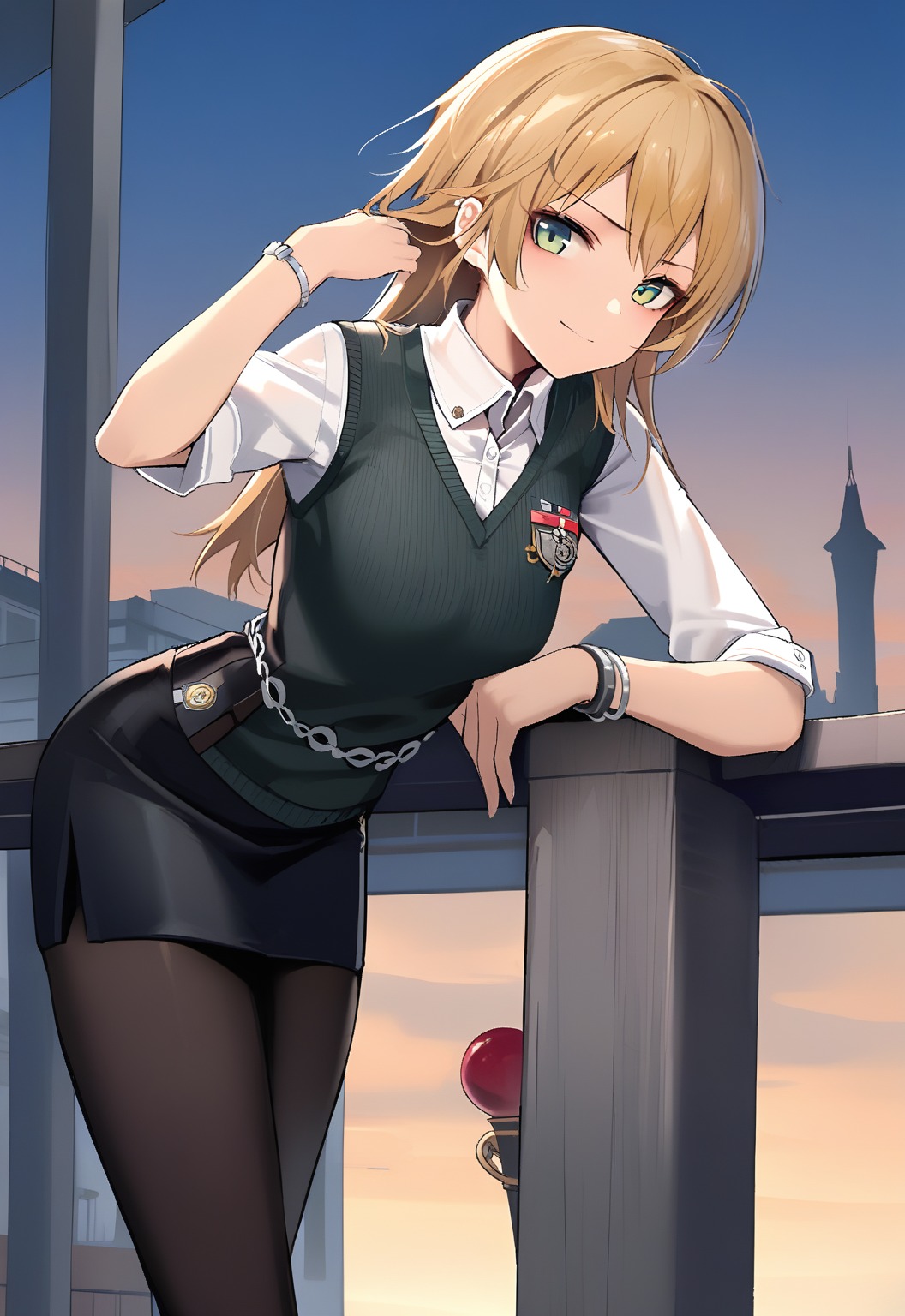} &
\includegraphics[width=0.32\linewidth]{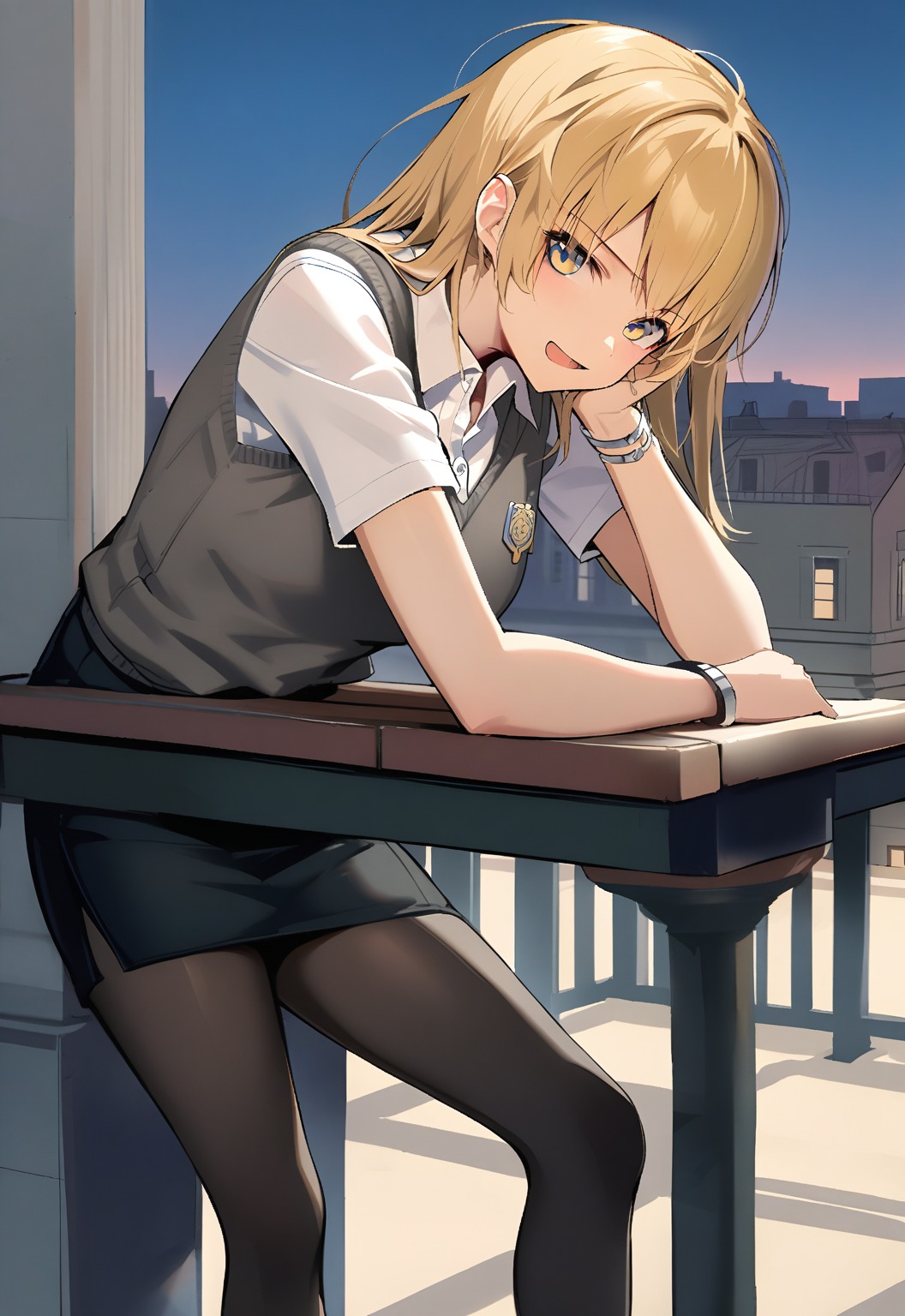} &
\includegraphics[width=0.32\linewidth]{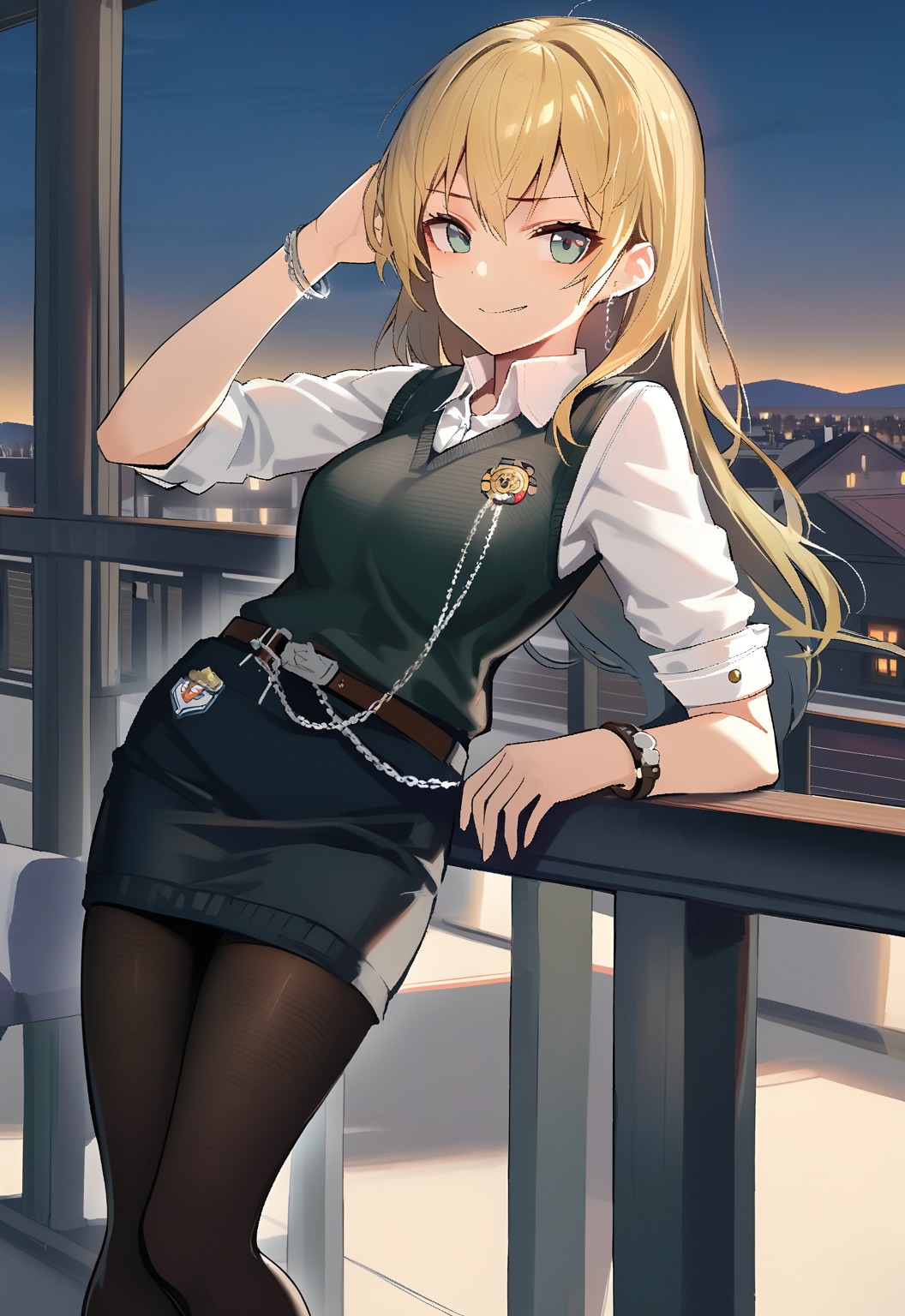} \\
{\small Full SGA} & {\small w/o Tuple-wise Opt.} & {\small w/o Min-SNR} \\
\end{tabular}
\caption{\textbf{SDXL ablation (Group 3).} SGA removal consistently causes the most significant anatomical degradation, supporting its critical role in compensating for CNN's limited global awareness.}
\label{fig:ablation-sdxl-3}
\end{figure}

We conduct analogous ablation studies on SDXL (\cref{fig:ablation-sdxl-1,fig:ablation-sdxl-2,fig:ablation-sdxl-3}). Notably, the primary advantage of Full SGA lies \emph{not} in color fidelity or atmospheric rendering, but in the \emph{anatomical correctness of body and limb structures}. Due to CNN's local receptive field, SDXL exhibits sharper sensitivity to fine-grained details yet faces greater challenges in maintaining global structural coherence. Both Min-SNR weighting and SGA contribute to stabilizing limb anatomy, but SGA appears to yield a \emph{larger overall gain}. We attribute this to two factors: (1)~the small batch size $B{=}2$ on SDXL makes it difficult for ARB buckets to serve as an implicit SGA---explicit tuple-wise coordination is therefore more necessary; and (2)~the intrinsic inductive bias of CNNs favors local detail over global structure, amplifying the need for explicit cross-scale intervention.

\section{Discussion: SGA in the Open-Source Diffusion Model Ecosystem}
\label{appendix:ecosystem}

The open-source diffusion model community has evolved along two principal architectural lineages, each presenting distinct optimization characteristics that contextualize the practical relevance of SGA. \cref{tab:ecosystem} summarizes this landscape.

\begin{table}[H]
\centering
\caption{\textbf{Open-source diffusion model ecosystem.} Representative foundation models and their community-driven derivatives, organized by backbone architecture.}
\label{tab:ecosystem}
\resizebox{\linewidth}{!}{%
\begin{tabular}{@{}lll@{}}
\toprule
\textbf{Architecture} & \textbf{Foundation Models} & \textbf{Community Derivatives} \\
\midrule
\multirow{2}{*}{U-Net}
  & SD\,1.5 / 2.1~\cite{rombach2022high}, & Pony Diffusion V6~\cite{ponyv6}, Animagine XL~\cite{animaginexl}, \\
  & SDXL~\cite{podell2024sdxl} & NoobAI-XL~\cite{noobaixl}, Illustrious XL~\cite{illustrious} \\
\midrule
\multirow{2}{*}{MM-DiT}
  & PixArt-$\alpha$~\cite{chen2023pixart}, SD\,3.5~\cite{esser2024scaling}, & --- \\
  & AuraFlow~\cite{auraflow}, FLUX.1~\cite{flux2024}, FLUX.2~\cite{flux2} & \\
\bottomrule
\end{tabular}%
}
\end{table}

\noindent\textbf{U-Net architecture: plasticity, ecosystem, and strong community priors.}
The U-Net lineage benefits from exceptionally low compute requirements and a mature community ecosystem, making it accessible for diverse fine-tuning scenarios.
Notably, the aesthetically aligned community derivatives listed in \cref{tab:ecosystem} possess strong learned priors acquired through large-scale curated training.
In this context, H-SD signal amplification can help \emph{counteract} these priors by providing structured, multi-granularity supervision that helps prevent the model from collapsing to its pre-existing stylistic attractor.
Furthermore, Tuple-wise Optimization enhances the structural robustness of generated outputs---a critical benefit given the CNN's inherent locality bias, as demonstrated in~\cref{appendix:g}.

\noindent\textbf{MM-DiT architecture: strong priors, hyperparameter sensitivity, and sample efficiency.}
Among MM-DiT models, the FLUX series~\cite{flux2024,flux2} stands as the flagship open-source release.
Its extremely powerful generative prior, combined with the hyperparameter sensitivity introduced by guidance distillation, has been a well-documented challenge in the practitioner community.
Moreover, the substantially larger parameter count of MM-DiT architectures may exacerbate optimization instability under few-sample fine-tuning regimes: with limited data, the empirical estimation of the target vector field density $\boldsymbol{\alpha}(x)$ becomes sparse and noisy (\cref{eq:nw}), leading to rugged loss landscapes and unreliable gradient updates.
In this context, the H-SD decomposition serves a dual purpose: by hierarchically expanding the effective number of training samples $N$ through structured semantic slicing, it empirically helps smooth the local vector field estimation (\cref{appendix:general-ntk}), contributing to a more stable optimization trajectory and faster convergence.
Concurrently, under a conservative hyperparameter configuration, the H-SD signal amplification enables effective suppression of the dominant prior, while the Scale-Adaptive Modulation component stabilizes training gradients by aligning the time-step distribution with the frequency characteristics of each data granularity (\cref{appendix:d}).
This combined mechanism facilitates more reliable domain adaptation without resorting to aggressive learning rates or extended training schedules that risk catastrophic forgetting~\cite{kirkpatrick2017overcoming}.

\section{Pseudocode}
\label{appendix:pseudocode}

This section provides pseudocode for the two core procedures of the SGA framework: \emph{(i)}~Hierarchical Semantic Decomposition (H-SD), the offline data preprocessing pipeline, and \emph{(ii)}~the SGA training loop, integrating Tuple-wise Optimization and Scale-Adaptive Modulation.

\begin{algorithm}[H]
\footnotesize
\caption{Hierarchical Semantic Decomposition (H-SD) Pipeline}
\label{alg:hsd}
\KwInput{Raw image directory $\mathcal{D}$;\; granularity levels $\mathcal{T} = \{\xi_1, \dots, \xi_L\}$ ordered coarse-to-fine;\; per-level config $\mathcal{C} = \{(\xi,\, \mathbf{s}_\xi,\, \tau^{\downarrow}_\xi,\, \tau^{\uparrow}_\xi)\}$}
\KwHyper{Pairwise IoU thresholds $\{\theta_{\xi_i, \xi_j}\}$;\; super-resolution model}
\KwOutput{Granularity-annotated dataset $\mathcal{D}' = \{(x, \xi, r)\}$}

$\mathcal{D}' \leftarrow \emptyset$\;
\ForEach(\Comment*[f]{Streaming: constant memory}){image $x$ from $\texttt{Source}(\mathcal{D})$}{
    $r \leftarrow \texttt{RootName}(x)$\;

    \Comment{--- Stage 1: Coarsest-level partitioning ---}
    \uIf(\Comment*[f]{Pre-split mode}){$\xi_1$\textnormal{-detection enabled}}{
        $\mathcal{R} \leftarrow \texttt{Detect}(x,\, \xi_1)$ \Comment*[r]{Detect coarsest-level regions}
    }
    \lElse(\Comment*[f]{Whole-image mode}){$\mathcal{R} \leftarrow \{x\}$}

    \ForEach{region $R \in \mathcal{R}$}{
        Register $(R,\, \xi_1,\, r)$ into $\mathcal{D}'$ \Comment*[r]{Each region = one coarsest-level sample}

        \Comment{--- Stage 2: Finer-level detection + IoU deduplication ---}
        $\{b_\xi\}_{\xi \in \mathcal{T} \setminus \xi_1} \leftarrow \texttt{DetectSubRegions}(R)$ \Comment*[r]{Detect finer levels within $R$}
        \ForEach{level pair $(\xi_i, \xi_j)$ with $i < j$}{
            \lIf{$\texttt{IoU}(b_{\xi_j},\, b_{\xi_i}) > \theta_{\xi_i, \xi_j}$}{suppress coarser level $\xi_i$}
        }

        \Comment{--- Stage 3: Per-level crop + adaptive resize ---}
        \ForEach{active level $\xi \in \mathcal{T} \setminus \xi_1$}{
            $x_\xi \leftarrow \texttt{Crop}(R,\, b_\xi)$\;
            $s^* \leftarrow \texttt{BestAspectMatch}(x_\xi,\, \mathbf{s}_\xi)$ \Comment*[r]{Select from target sizes}
            \lIf{scale ratio $> \tau^{\uparrow}_\xi$}{\textbf{discard} $x_\xi$}
            \lIf{scale ratio $> \tau^{\downarrow}_\xi$}{$x_\xi \leftarrow \texttt{Downsample}(x_\xi,\, s^*)$}
            \lIf{upscale needed}{$x_\xi \leftarrow \texttt{SuperResolve}(x_\xi)$}
            $x_\xi \leftarrow \texttt{FramingCrop}(x_\xi,\, s^*)$ \Comment*[r]{Centroid-based composition}
            Register $(x_\xi,\, \xi,\, r)$ into $\mathcal{D}'$\;
        }
    }
}
\Return{$\mathcal{D}'$}\;
\end{algorithm}

\begin{algorithm}[H]
\footnotesize
\caption{SGA Training Loop with Dynamic Tuple Accumulation}
\label{alg:sga}
\KwInput{H-SD dataset $\mathcal{D}' = \{(x, \xi, r)\}$ in semantic groups; pre-trained model $v_\theta$; arch.\ type $\mathcal{A} \in \{\texttt{U-Net}, \texttt{MM-DiT}\}$}
\KwHyper{Learning rate $\eta$;\; U-Net $\Gamma\text{-table} = [4.0,\, 5.0,\, 7.0]$;\; DiT $\Delta\text{-table} = [+0.5,\, 0.0,\, -0.5]$}
\KwOutput{Fine-tuned parameters $\theta^*$}

\For{each epoch}{
    \texttt{ShuffleWithinGroups}($\mathcal{D}'$) \Comment*[r]{Randomize order within each semantic group}
    Arrange $\mathcal{D}'$ into batches via ARB, preserving group contiguity\;
    $\mathrm{acc\_count} \leftarrow 0$;\quad $K_{\mathrm{target}} \leftarrow \texttt{null}$\;

    \ForEach{batch $\mathcal{B}$ from data loader}{
        Read group size $K_g$ and granularity $\xi$ from batch metadata\;
        \lIf{$K_{\mathrm{target}} = \texttt{null}$}{$K_{\mathrm{target}} \leftarrow K_g$}

        \Comment{--- Scale-Adaptive Modulation ---}
        \uIf(\Comment*[f]{Conditional Logit-Normal}){$\mathcal{A} = \texttt{MM-DiT}$}{
            $z \sim \mathcal{N}(0, 1)$;\quad
            $t_{\mathrm{logit}} \leftarrow \sigma\!\big(z + \Delta[\xi]\big)$\;
            $\mu_{\mathrm{res}} \leftarrow \texttt{LinInterp}\!\big(\tfrac{H}{2} \cdot \tfrac{W}{2}\big)$;\quad
            $t \leftarrow t_{\mathrm{logit}} \,/\, \big(t_{\mathrm{logit}} + (1 - t_{\mathrm{logit}}) \cdot e^{-\mu_{\mathrm{res}}}\big)$\;
            $\epsilon \sim \mathcal{N}(\mathbf{0}, \mathbf{I})$;\quad
            $x_t \leftarrow \mu_t(x_1) + \sigma_t \cdot \epsilon$\;
            $\mathcal{L} \leftarrow \big\| v_\theta(x_t, t) - u_t(x_t | x_1) \big\|^2$\;
        }
        \Else(\Comment*[f]{Adaptive Min-SNR}){
            Sample $t$ from noise schedule;\quad
            $\epsilon \sim \mathcal{N}(\mathbf{0}, \mathbf{I})$;\quad
            $x_t \leftarrow \sqrt{\bar{\alpha}_t}\, x_1 + \sqrt{1 - \bar{\alpha}_t}\, \epsilon$\;
            $\mathcal{L}_{\mathrm{raw}} \leftarrow \big\| \epsilon_\theta(x_t, t) - \epsilon \big\|^2$;\quad
            $\omega \leftarrow \min\!\big(\mathrm{SNR}(t),\;\Gamma[\xi]\big) / \mathrm{SNR}(t)$\;
            $\mathcal{L} \leftarrow \omega \cdot \mathcal{L}_{\mathrm{raw}}$\;
        }

        \Comment{--- Tuple-wise gradient accumulation ---}
        $\mathrm{acc\_count} \leftarrow \mathrm{acc\_count} + 1$;\quad
        \texttt{Backward}$\!\big(\mathcal{L}\big)$ with gradient sync \textbf{deferred}\;

        \If(\Comment*[f]{Tuple complete}){$\mathrm{acc\_count} \geq K_{\mathrm{target}}$}{
            \texttt{SyncGradients}();\quad \texttt{ClipGradNorm}($\theta$)\;
            $\theta \leftarrow \theta - \eta \cdot \texttt{Optimizer}(\nabla_\theta)$;\quad
            $\nabla_\theta \leftarrow \mathbf{0}$;\quad
            $\mathrm{acc\_count} \leftarrow 0$;\quad $K_{\mathrm{target}} \leftarrow \texttt{null}$\;
        }
    }
}
\end{algorithm}

\subsection*{Notes}

\noindent\textbf{Data representation.}\;
The H-SD pipeline produces $\mathcal{D}' = \{(x_i, \xi_i, r_i)\}$, where $\xi_i \in \{\textsc{Macro}, \textsc{Meso}, \textsc{Micro}\}$ is the semantic granularity and $r_i$ is the \emph{root name} identifying the source image.
Samples sharing the same root name form a \emph{semantic group} with size $K_g \in \{1, 2, 3\}$, determined once during dataset construction.
$\Delta[\xi]$ and $\Gamma[\xi]$ index the architecture-specific modulation tables by granularity (see Suppl.\ D).

\smallskip
\noindent\textbf{ARB orthogonality.}\;
Semantic grouping and Aspect-Ratio Bucketing operate on independent axes (semantics vs.\ resolution) and compose without special-case logic. Samples of the same granularity may reside in different resolution buckets matching each crop's actual aspect ratio.

\smallskip
\noindent\textbf{VRAM equivalence.}\;
Each forward/backward pass processes one batch; accumulation across $K_g$ sub-batches is achieved by deferring the optimizer step, introducing \emph{no additional peak memory} compared to baseline training at the same per-step batch size.

\smallskip
\noindent\textbf{Intra-group shuffling.}\;
At each epoch boundary, sample ordering within each group is randomly permuted, preventing the model from learning a fixed granularity presentation order while preserving cross-scale co-occurrence.

\smallskip
\noindent\textbf{Modular pipeline automation.}\;
The H-SD pipeline is modular and self-contained.
The detection backend (\texttt{Detect}/\texttt{DetectSubRegions}) supports both specialized YOLO-based detectors and, for domain-specific tasks, multimodal vision-language detectors (\eg, Grounding DINO), selectable via a single configuration flag.
Each granularity level admits multiple target resolution templates $\mathbf{s}_\xi$ (following the ARB bucketing logic of NovelAI~\cite{novelai2022bucketing}), enabling aspect-ratio-aware output that directly interfaces with downstream training pipelines.
Built-in resolution presets allow rapid configuration, while per-dataset overrides through hierarchical config inheritance support deployment across heterogeneous data sources with minimal manual effort.

\smallskip
\noindent\textbf{Mature ecosystem integration.}\;
The preprocessing pipeline leverages GPU-accelerated open-source components---object detectors for semantic parsing, ESRGAN models for super-resolution---within a streaming, constant-memory architecture.
As reported in Suppl.~F, end-to-end H-SD processing takes 15--30~minutes per dataset, an overhead negligible relative to training.

\end{document}